%% file: arxiv.tex
\documentclass[11pt]{article}

\usepackage[eandd,preprint]{neurips_2026}

\setcitestyle{numbers,square,sort&compress,comma}

\usepackage[utf8]{inputenc}
\usepackage[T1]{fontenc}
\usepackage{hyperref}
\usepackage{url}
\usepackage{booktabs}
\usepackage{tabularx}
\usepackage{amsfonts}
\usepackage{amsmath}
\usepackage{nicefrac}
\usepackage{microtype}
\usepackage{xcolor}
\usepackage{graphicx}
\usepackage{float}
\usepackage{enumitem}
\usepackage{tikz}
\usetikzlibrary{arrows.meta, positioning, shapes.geometric, fit, backgrounds}
\usepackage{pdfpages}
\usepackage{subcaption}
\usepackage{pifont}

\newcommand{\pcfbench}{\textsc{PCFBench}}
\newcommand{\ecoinvent}{\texttt{ecoinvent}}
\newcommand{\kgco}{kgCO$_2$e}
\newcommand{\cmark}{\textcolor{green!55!black}{\ding{51}}}
\newcommand{\xmark}{\textcolor{red!75!black}{\ding{55}}}

\newif\ifanonrelease\anonreleasefalse
\newcommand{\datahost}{Hugging Face}
\newcommand{\datalink}{\url{https://huggingface.co/datasets/Watershed-Climate/PCFBench}}
\newcommand{\codelink}{\url{https://github.com/watershed-climate/pcfbench}}

\title{PCFBench: A Diagnostic Benchmark for Product Carbon Footprint Estimation}

\author{%
  \normalfont
  Krishna Rao\thanks{Equal contribution.}%
  \hspace{0.5em}\thanks{Corresponding author:
    \texttt{krishna@watershedclimate.com}} \quad
  Andrew Dumit\footnotemark[1] \quad
  Shaena Ulissi \quad
  Jacob Feintzeig \quad
  P.~James Joyce \\
  Daniel Frank \quad
  Steven Watson \quad
  Jonathan Glidden \quad
  Gizem Ilayda Dinc \\
  Travis M. Kwee \\
  Watershed Technology, Inc. \\[0.4em]
  \small Dataset: \datalink \\
  \small Code: \codelink
}

\begin{document}

\maketitle

\begin{abstract}

AI systems are being deployed on high-stakes, 
domain-specific workflows that demand correctness not just in the final output, 
but at every intermediate step.
One such workflow is estimating a product carbon footprint (PCF),
the greenhouse-gas emissions attributable to a physical product.
AI agents are increasingly being used to generate PCFs, but existing 
evaluations score either total emissions (hiding error
sources and cancelling mistakes) or sub-tasks in isolation (missing
compositional interactions). We introduce \pcfbench{}, the first 
benchmark to carve PCF modeling into
independently-evaluable tasks that require decomposition, retrieval, ontology
matching, and numerical extraction. It comprises 614
expert-labelled items across six tasks.
Together they probe reasoning under under-specification,
conflicting context, and numerical constraints.
Across eight frontier LLMs from four
providers, no single model dominates.  Although the strongest models
estimate total product emissions within $2\times$ of declared totals
on 77\% of products, this rate drops to 37--58\% when
the PCF is generated step by step,
with only 45--75\% obeying mass conservation.
These failures undermine the transparency practitioners need to
compare products and drive decarbonization.
We release the dataset and evaluation harness to support targeted progress.
\end{abstract}

\input{sections/introduction}
\input{sections/related_work}
\input{sections/methodology}

\input{sections/datasets}
\input{sections/results}

\newpage
\bibliographystyle{plainnat}
\bibliography{references}

\newpage
\appendix
\input{sections/appendix}

\end{document}

%% file: sections/introduction.tex
\section{Introduction}
\label{sec:introduction}

\begin{figure}[t]
\centering
\includegraphics[width=\textwidth]{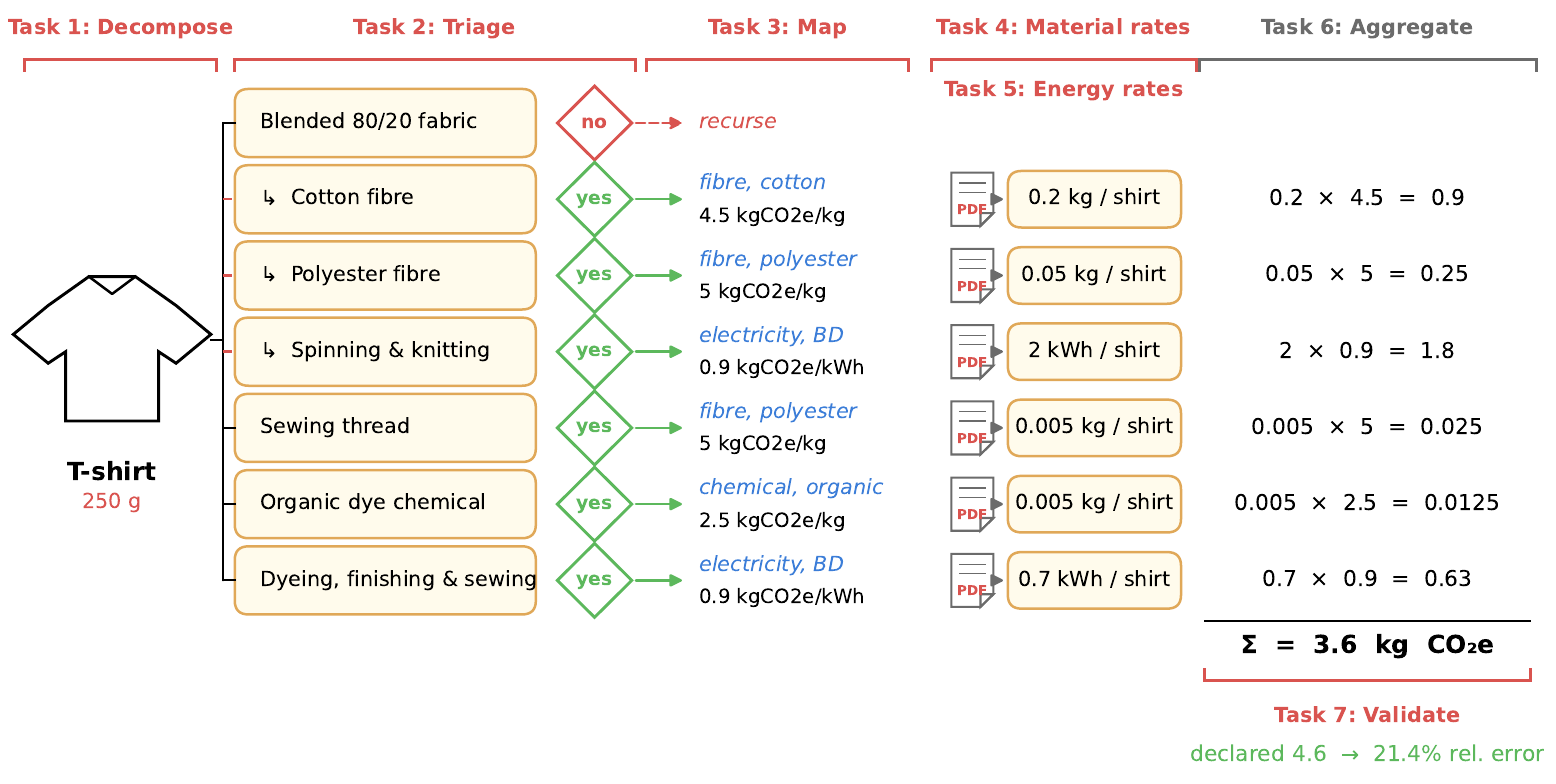}
\caption{Decomposed PCF creation. Orange brackets
denote tasks covered by \pcfbench{}; grey is
deterministic.}
\label{fig:pipeline}
\end{figure}

Recent benchmarks evaluate AI systems on domain-specific, multi-step
workflows: software engineering \citep{jimenez2024swebench},
scientific code \citep{scienceagentbench}, chemistry
\citep{mirza2025chembench}, medicine \citep{liu2024medbench}, and
law \citep{legalbench}.  However, evaluation studies show that aggregate-output
evaluation can conceal sub-task failures on compositional tasks.
Many agent benchmarks focus on final outcomes and miss the step-by-step
process \citep{zhuge2024agentasjudge, ma2024agentboard}.
For example, in a corpus of 5{,}000+ human-evaluated LLM-generated
mathematical proofs, final-answer accuracy substantially overstates
full-proof validity \citep{dekoninck2025openproof}.
Sub-task-resolved evaluation is correspondingly useful for probing
true model capability rather than averaging it out
\citep{measuringwhatmatters, shen2024taskbench}.
Estimating a Product Carbon Footprint (PCF) offers exactly this
kind of testbed: it chains interdependent sub-tasks spanning
document reading, ontology matching, and numerical reasoning, and
end-to-end ground truth (published Environmental Product
Declarations; EPDs) is available for validation.

A PCF is the total \kgco{} (kilograms of CO$_2$-equivalent) attributable
to one declared unit (e.g., one garment, one kilogram of aluminium)
of a product, computed via Life Cycle Assessment (LCA) under
ISO~14040/14044 \citep{iso14040, iso14044}.  This framework is used
to issue EPDs \citep{iso14025, environdec}, inform procurement,
and complete corporate climate disclosures
(e.g., the EU Corporate Sustainability Reporting Directive (CSRD)
value-chain disclosures).  Effectively reducing
greenhouse gas emissions
(a necessary step towards climate change mitigation;
\citep{ipcc2023synthesis, ghgprotocol}) requires measuring them at
the product level.

A growing body of work applies LLMs to parts of the PCF pipeline: AutoPCF
\citep{deng2023autopcf} drives a direct accounting framework that
produces a PCF from a product name. Parakeet \citep{balaji2025parakeet}
specializes in mapping materials to \ecoinvent{} activities 
(a database of physical processes and their associated greenhouse gas 
emissions; \citep{ecoinvent}). Others
target inventory generation or extraction in isolation
\citep{zhang2025autonomous, pcfrwkv2025, kumar2025sustainllama}.  But
these typically evaluate at one or two coarse interfaces,
such as total \kgco{} error against expert estimates on a handful of
products.  AutoPCF, the closest framework, evaluates two
stages: a ROUGE-style $F_1$ on the LLM-generated inventory, and
total \kgco{} error against expert estimates on three case products.
The relative difficulty of sub-tasks---and where current models
succeed or fail---is invisible.
Knowledge-based LCA evaluations
\citep{donaldson2025lcabenchmark, he2025esgenius} sidestep this by
testing what models \emph{know about} LCA rather than whether they can
\emph{perform} it. Aggregate-output scoring also cannot separate
correct reasoning from canceling errors \citep{an2026stad}, where a
$2\times$ over-estimate of material inputs paired with a $0.5\times$
too-low emission factor produces the right answer for the wrong reasons.

\textbf{LCA is as much art as science.}
LCA is a useful evaluation domain because it is pervaded by \emph{``it depends''}
reasoning.  Almost any question an LCA practitioner faces has an
answer that varies with the goal and scope of the study, the
geographic and temporal context, and the methodological conventions
of the reporting standard.  Consider a seemingly simple question:
\emph{should the carbon footprint of a recycled aluminium can include
the emissions from the original aluminium smelting?}  The answer
depends on whether the study uses the \emph{cut-off} allocation
method (no, the recycled material enters the system burden-free) or
the \emph{avoided burden} method (yes, but offset by a credit for
displacing virgin production if the can is successfully recycled again at the end of its life).  Both are technically valid under
ISO~14044 \citep{iso14044}; which one an expert chooses depends on
the study's goal, the commissioner's reporting framework, and
established practice in the product category.  An LLM asked this
question will typically commit to one answer without stating the
assumption. This is precisely the kind of overconfident,
context-insensitive behaviour documented across frontier models
\citep{hagar2025notwrong, cheng2026elephant, vennemeyer2025sycophancy}.
This pervasive context-dependence means LCA cannot be fully
automated through hard-coded rules: the space of methodological
choices is too large, too context-sensitive, and too dependent on
expert judgment. This makes LCA a natural application domain for AI
agents.

\textbf{PCFs are models, not single numbers.}
Actionable PCFs require more than a final emissions estimate:
practitioners need the intermediate stages to attribute impacts,
compare products under a fixed methodology, and quantify decarbonization
levers (e.g., recycled plastic, renewable electricity)
against a stable baseline.  Re-prompting a black-box estimator with one variable
changed cannot deliver this; nothing guarantees the other choices
stay fixed call-to-call, and the result lacks the audit trail
needed for verification, scope-checking, or hotspot analysis.

\textbf{\pcfbench{}.}
We introduce \pcfbench{}, a six-task decomposition of the
cradle-to-gate PCF estimation methodology.  Cradle-to-gate PCFs cover
raw materials through manufacturing, excluding the product's use phase
and end-of-life (e.g., the impacts of producing a bottle of laundry
detergent, but not those of washing the laundry or disposing of the
bottle).  Each task has typed input/output schemas, a task-specific
metric, and an expert-annotated dataset (614 items in total).

\textbf{Common PCF tasks (cradle-to-gate).}
Process-based PCF estimation can be approached in many ways, but a
common AI-assisted formulation for cradle-to-gate PCFs
follows a recursive seven-step pipeline: six LLM-evaluable tasks
plus a deterministic aggregation step (Figure~\ref{fig:pipeline}).

\begin{enumerate}[nosep,leftmargin=*]
    \item \textbf{Decompose} the product into a bill of materials
    (BOM) of constituent materials and sub-components.  This
    requires implicit world knowledge about how products are made
    and at what level of fabrication each input enters the BOM;
    missing an entry is a silent error that no downstream metric
    flags.

    \item \textbf{Triage}: for each BOM item, decide whether it can
    be mapped directly to a background-database process or must be
    further decomposed. BOM items judged unmappable flow back to
    task 1 to be recursively decomposed.
    The decision rests on recognising when no candidate is
    a good enough match rather than committing to an imperfect one.

    \item \textbf{Map each non-recursed material to an emission
    factor} from a background database such as \ecoinvent{}
    \citep{ecoinvent}.  The mapping target is a domain ontology
    where general semantic similarity diverges from domain-specific
    correctness; the right candidate is often not the most
    lexically similar one.  For example,
    \texttt{gold plating chemicals} should map to the actual
    plating inputs like organic chemicals, but not to
    \texttt{gold} itself.

    \item \textbf{Extract material input rates from technical
    literature}: locate the relevant document(s), including carbon footprint
    reports, EPDs, academic papers, supplier datasheets, and industry
    handbooks, and read out how much of each material is consumed
    per declared unit.  Extraction errors here propagate
    multiplicatively into the final PCF.

    \item \textbf{Extract energy input rates from technical
    literature}: read out the electricity, heat, and fuel consumed
    per declared unit by manufacturing rows that map to a
    process-only \ecoinvent{} activity (i.e., one whose EF excludes
    utility energy).  The model must additionally judge what is and
    is not already absorbed into the chosen background process.

    \item \textbf{Aggregate}: multiply each input rate by its
    emission factor and sum.  This step is deterministic arithmetic
    and is not evaluated.

    \item \textbf{Validate} the composed estimate against ground
    truth such as published EPDs.  Scoring only the total \kgco{}
    cannot localise where the error came from; decomposed
    evaluation is what makes the residual diagnosable.
\end{enumerate}

\textbf{Contributions.}
\textbf{(i)~The first benchmark for PCF estimation.} \pcfbench{} carves
the cradle-to-gate PCF pipeline into six evaluable tasks.
The decomposition is stricter
than prior frameworks: triage, the recursive map-or-decompose
decision that supports arbitrary product depth, is a first-class
task; material and energy extraction are evaluated separately because
they probe different scope-boundary judgments; and every interface is
typed so specialized baselines (e.g., Parakeet on mapping) plug into
the same harness as frontier LLMs.
\textbf{(ii)~Expert-annotated datasets}: 614 items hand-labelled by
sustainability experts---89 evidence-grounded extraction claims
across 36 PDFs, 109 material-to-\ecoinvent{} mappings, 200
map-or-decompose triage decisions, 175 EPD-sourced total-\kgco{}
validations spanning five orders of magnitude, and 94 BOM
decompositions.
\textbf{(iii)~Baseline performance} on eight frontier models from
four providers and a specialized mapping baseline (Parakeet),
establishing per-task performance profiles and identifying the
primary capability bottlenecks of current models.

The dataset and evaluation harness are publicly
available\ifanonrelease\else{}
 (links under the author list).\fi{}
Per-task data rights, licensing, and permissions are documented in
Appendix~\ref{sec:licensing}.

%% file: sections/related_work.tex
\section{Related Work}
\label{sec:related_work}

\begin{table}[!t]
\centering
\caption{Data coverage of \pcfbench{} versus prior datasets.  \cmark{} = task-relevant ground-truth
is publicly available; \xmark{} = no.}
\label{tab:task_coverage}
\footnotesize
\setlength{\tabcolsep}{3pt}
\begin{tabular}{@{}l c c c c c c@{}}
\toprule
 & \textbf{Task 1} & \textbf{Task 2} & \textbf{Task 3} & \textbf{Task 4} & \textbf{Task 5} & \textbf{Task 7} \\
 & Decompose & Triage & Map & Material extraction & Energy extraction & Total \kgco{} \\
\midrule
Babbitt et al.\ 2020 \citep{babbitt2020disassembly}        & \cmark & \xmark & \xmark & \xmark & \xmark & \xmark \\
Parakeet \citep{balaji2025parakeet}                        & \xmark & \xmark & \cmark & \xmark & \xmark & \xmark \\
Sustain-LLaMA \citep{kumar2025sustainllama}                & \xmark & \xmark & \xmark & \cmark & \cmark & \xmark \\
Carbon Catalogue \citep{meinrenken2022carboncatalogue}     & \xmark & \xmark & \xmark & \xmark & \xmark & \cmark \\
Eco-Amazon \citep{pillo2026ecoamazon}                      & \xmark & \xmark & \xmark & \xmark & \xmark & \cmark \\
\textbf{\pcfbench{} (ours)}                                & \cmark & \cmark & \cmark & \cmark & \cmark & \cmark \\
\bottomrule
\end{tabular}
\end{table}

\textbf{LCA automation and the aggregate-evaluation gap.}
LLM-based PCF systems have proliferated rapidly: direct
generators \citep{deng2023autopcf, pcfrwkv2025}, multi-agent
pipelines \citep{zhang2025autonomous, preuss2026automatinglca},
domain-specific systems for food \citep{aslan2025cfwizard} and
electronics \citep{zhang2024deltalca, pillo2026ecoamazon},
mapping-only systems \citep{balaji2025parakeet, castle2025entity,
peng2024kgmapping, gachkar2025fuzzy, dumit2026qualityaware}, and
LCI extraction pipelines \citep{kumar2025sustainllama,
dagdelen2024nerre, ghosh2025matprops}; multiple reviews call for
standardized evaluation \citep{preuss2024llmlca, tu2024mitigating,
gkousis2025mllci, mensikova2026landscape, ulissi2025criteria}.
Yet every existing system is evaluated only on aggregate \kgco{},
hiding which sub-task drove an error \citep{an2026stad}.

\textbf{Adjacent sustainability benchmarks.}
Existing sustainability benchmarks address adjacent problems: ExioML
\citep{guo2026exioml} provides sector-level emission factors;
ESGenius \citep{he2025esgenius} and ClimateEval
\citep{kurfali2025climateeval} test sustainability \emph{knowledge}
via Q\&A; SustainBench \citep{yeh2021sustainbench} monitors SDGs
from satellite imagery.  The closest existing LCA evaluation,
\citet{donaldson2025lcabenchmark}, benchmarks 11~LLMs on 22
expert-graded Q\&A tasks but tests \emph{declarative} knowledge,
not operational capability.  EPD registries
\citep{meinrenken2022carboncatalogue, cardoso2024epdharmonisation,
aragon2024epdlimitations} provide data without evaluation tasks.

\textbf{Compositional evaluation and per-task failure modes.}
\pcfbench{}'s decomposed architecture instantiates the compositional
evaluation paradigm the broader community has called for
\citep{shen2024taskbench, measuringwhatmatters,
zhuge2024agentasjudge, ma2024agentboard, zhang2026dataclaw},
mirroring domain-decomposed benchmarks like LegalBench
\citep{legalbench}.  Each task additionally targets a documented
LLM failure mode (Section~\ref{sec:introduction}, inline with the
pipeline description): document extraction with conflicting context
 \citep{sui2024table,
wang2024charxiv, ma2024mmlongbenchdoc, liu2024lost} and numerical
reasoning \citep{mirzadeh2025gsm, li2025numericbench,
hagar2025notwrong, bang2025hallulens} for Tasks~4--5; ontology
alignment \citep{qiang2024oaeillm, song2026genom, qiang2025oaeillmt,
guan2025ordereffect, zhuo2024prosa} and under-specification / belief
revision \citep{kirichenko2025abstentionbench, wen2025knowyourlimits,
li2025questbench, wilie2024belief, pu2026lhaw} for Tasks~2--3; and
sub-task failure masking under aggregate-output scoring
\citep{an2026stad, plaat2025multistep}.
Table~\ref{tab:task_coverage} summarises openly-available evaluation
data per task. Existing datasets are restricted to either individual
tasks or narrow categories
(e.g., Babbitt~et~al.~2020~\citep{babbitt2020disassembly} is
electronics-only; Sustain-LLaMA~\citep{kumar2025sustainllama} is
plastic-packaging and methanol only).

%% file: sections/methodology.tex
\section{Benchmark Design}
\label{sec:methodology}

A cradle-to-gate PCF calculation produces:
\begin{equation}
\label{eq:pcf}
\text{PCF} = \sum_{i=1}^{n} q_i \cdot \text{EF}_i
\end{equation}
where $q_i$ is the physical input rate (e.g., kg of material per
declared unit) and $\text{EF}_i$ is the emission factor (e.g.,
\kgco{}/kg) for the $i$-th input.  Arriving at this sum requires
expert judgment at six points: identifying the inputs (Task~1),
deciding whether each is mappable or must be further decomposed
(Task~2), selecting the right emission factor (Task~3), estimating
material quantities (Task~4), estimating energy quantities (Task~5),
and predicting the total \kgco{} per declared unit (Task~7).
We evaluate each as a stand-alone benchmark task with task-specific metrics
(input format, output schema, and exact metric definitions per task in
Appendix~\ref{sec:task_eval_methodology}).
Table~\ref{tab:dataset_overview} summarises each task's input,
output, dataset size, and evaluation metrics; the total-\kgco{}
prediction protocol (four context-ablation settings on
175~EPDs) is in Appendix~\ref{sec:integration}.

\paragraph{Two evaluation modes for the integrated PCF.}

\emph{Direct.} A single LLM call returns one \kgco{} number, given
the EPD's product context. The model
performs no intermediate decomposition; the entire PCF computation
happens inside one forward pass.

\emph{Compositional.} \pcfbench{}'s Task~1--5 agents are chained
end-to-end---decompose~$\to$~triage~$\to$~map~$\to$~estimate material
rates~$\to$~estimate energy rates---and the resulting per-component
contributions are summed via deterministic \ecoinvent{} emission
factors (Task~6).  The output is one \kgco{} number assembled from
per-stage agent outputs with every intermediate step inspectable.  The
exact pipeline (sub-task chaining, frozen energy EFs, aggregation
rule, etc.) is specified in Appendix~\ref{sec:app_compositional}.

Both modes produce a single \kgco{} number per product and are scored
against the same EPD ground truths. Since there is inherent
variability in EPDs for identical products, we use the $2\times$
acceptability threshold (\textgreater50\%,
\citep{gelowitz2017,konradsen2024}): a prediction counts as correct
if it is within $2\times$ of the ground truth.

%% file: sections/datasets.tex
\section{Datasets}
\label{sec:datasets}

\pcfbench{} comprises 614~items across six evaluable tasks, all
hand-labelled by sustainability experts through a substantial annotation
effort, with cross-annotator unanimous voting on Tasks~4--5 and
field-by-field review by at least six experts on the EPD-derived items (Appendix~\ref{sec:annotation}).
Table~\ref{tab:dataset_overview} summarises the datasets.
Items span seven product categories and five orders of
magnitude in declared \kgco{}
(Figure~\ref{fig:coverage_heatmap}).
Full provenance, schemas, examples, and per-task
limitations appear in Appendix~\ref{sec:dataset_details}, and a
Datasheet for Datasets \citep{datasheets} appears in
Appendix~\ref{sec:datasheet}.

\begin{table}[t]
\centering
\caption{%
    \pcfbench{} dataset overview and per-task evaluation metrics.
    Task~6 (multiply and sum) is deterministic and not
    evaluated.  Exact metric definitions in
    Appendix~\ref{sec:task_eval_methodology}.}
\label{tab:dataset_overview}
\footnotesize
\setlength{\tabcolsep}{4pt}
\begin{tabularx}{\textwidth}{@{}l >{\raggedright\arraybackslash}X >{\raggedright\arraybackslash}X >{\raggedright\arraybackslash}X >{\hsize=0.8\hsize\raggedright\arraybackslash}X >{\hsize=1.2\hsize\raggedright\arraybackslash}X@{}}
\toprule
\textbf{Task} & \textbf{Items} & \textbf{Source} &
  \textbf{Input} & \textbf{Output} & \textbf{Metrics} \\
\midrule
1. Decomposition  & 94 products             & EPD                                & Product description           & Bill of materials    & $F_1$, precision, recall, Kendall $\tau$ \\
2. Triage         & 200 items               & Hand-generated                     & Candidate + root material context & Map / decompose     & Accuracy, $F_1$ \\
3. Mapping        & 109 materials           & Hand-generated                     & Material name ($\pm$ context) & \ecoinvent{} activity & Exact-match, relevant-substring, banned-substring violation \\
4. Material rates & 22 questions, 55 claims & PDF                                & Question + source PDF         & kg + evidence        & Claim $F_1$, supported $F_1$, P90 $|$RE$|$, hallucination \\
5. Energy rates   & 14 questions, 34 claims & PDF                                & Question + source PDF         & kWh/MJ + evidence    & Claim $F_1$, supported $F_1$, P90 $|$RE$|$, hallucination \\
7. Validation     & 175 products            & EPD                                & Product ($\pm$ comp., region) & \kgco{}              & Median $|$RE$|$, within-2$\times$, within-5$\times$ \\
\bottomrule
\end{tabularx}
\end{table}

\textbf{Task 1 (Decomposition).}
The 94-item BOM dataset is the composition-bearing subset of the
Task~7 EPDs.  Each composition entry was extracted verbatim from
the source EPD and reviewed during the Task~7 annotation pass.  We
strip the composition block from the input description so the model
must derive the bill of materials from product name and free-text
description alone.  Median 4 components per item, max~8 (Appendix
Figure~\ref{fig:dataset_task1_bomsize}).

\textbf{Task 2 (Triage).}
200 hand-labelled map-vs-decompose decisions.  Each item is one
decision point: a candidate material (e.g., fabric) plus its 
root-material context (e.g., t-shirt).
The \texttt{should\_map} label is \texttt{true} if an existing
\ecoinvent{} activity well-represents the candidate, \texttt{false}
if it needs to be decomposed further into a sub-BOM.  The dataset
is balanced 100/100.

\textbf{Task 3 (Mapping).}
109 materials spanning
typical mapping cases, edge cases (identity mappings, abbreviations,
foreign-language names), and specialised challenge sets (catalysts,
packaging, no-good-match scenarios).  Each item carries an unordered
set of defensible \ecoinvent{} mappings and an expert-rated vagueness
severity (1--5, intentionally skewed hard: 61\% at
severity~$\geq$\,4; Appendix
Figure~\ref{fig:dataset_task3_vagueness}). 38 items additionally
carry a \texttt{banned\_substring}, a known-wrong mapping the model
must avoid.
The mapping target is a 2{,}574-item subset of \ecoinvent{}~v3.11
(market activities only, mass-based declared unit, one geography
per reference product). The same picklist is the option set for
Task~2's triage decision. Construction details in
Appendix~\ref{sec:picklist}.

\textbf{Tasks 4--5 (Extraction).}
This dataset consists of 36~technical PDFs including 
carbon footprint reports, EPDs, academic papers,
industry handbooks spanning aluminum smelting, plastics processing.
For each PDF, sustainability
experts wrote one curated extraction query and annotated multiple
candidate claims---each of which \emph{could} be the answer to the
query---along with verbatim evidence quotes.  PDFs had 1 to 10+
candidate claims per query, each supported by at least one evidence
quote (some queries required stitching evidence across several text
blocks). Each query's candidate claims were independently reviewed by up to
three experts. Only claims with unanimous expert agreement entered
the released dataset, and three PDFs whose every candidate claim was
disputed were archived (full methodology in
Appendix~\ref{sec:dataset_extraction}). In total, 36~PDFs yield
89~ground-truth claims supported by 227~evidence quotes (Appendix
Figures~\ref{fig:dataset_extraction_extras}
and~\ref{fig:dataset_task4_5_units}).

\textbf{Task 7 (Total \kgco{} Prediction).}
The Task~7 corpus comprises 175~Environmental Product Declarations
registered with the International EPD System \cite{environdec} and
third-party verified per ISO~14025 \cite{iso14025}.  Each EPD is a
multi-page PDF. The
structured fields used here---product name, description, declared
\kgco{}, declared unit, material composition, geography, recycled
content---were extracted from the source PDFs and field-by-field
reviewed.  The field-set inclusion is governed by the EPD 
International permission letter
(Appendix~\ref{sec:epd_permission}).  Declared \kgco{} values span
three orders of magnitude (0.062--34.1; Figure~\ref{fig:coverage_heatmap}b).
  Each item supports four
progressive-disclosure settings constructed from cumulative fields:
\emph{name only}, \emph{with description}, \emph{with composition}
(94~items have parsed composition), and \emph{with region}
(94~items have parsed geography).

\begin{figure}[t]
\centering
\includegraphics[width=\linewidth]{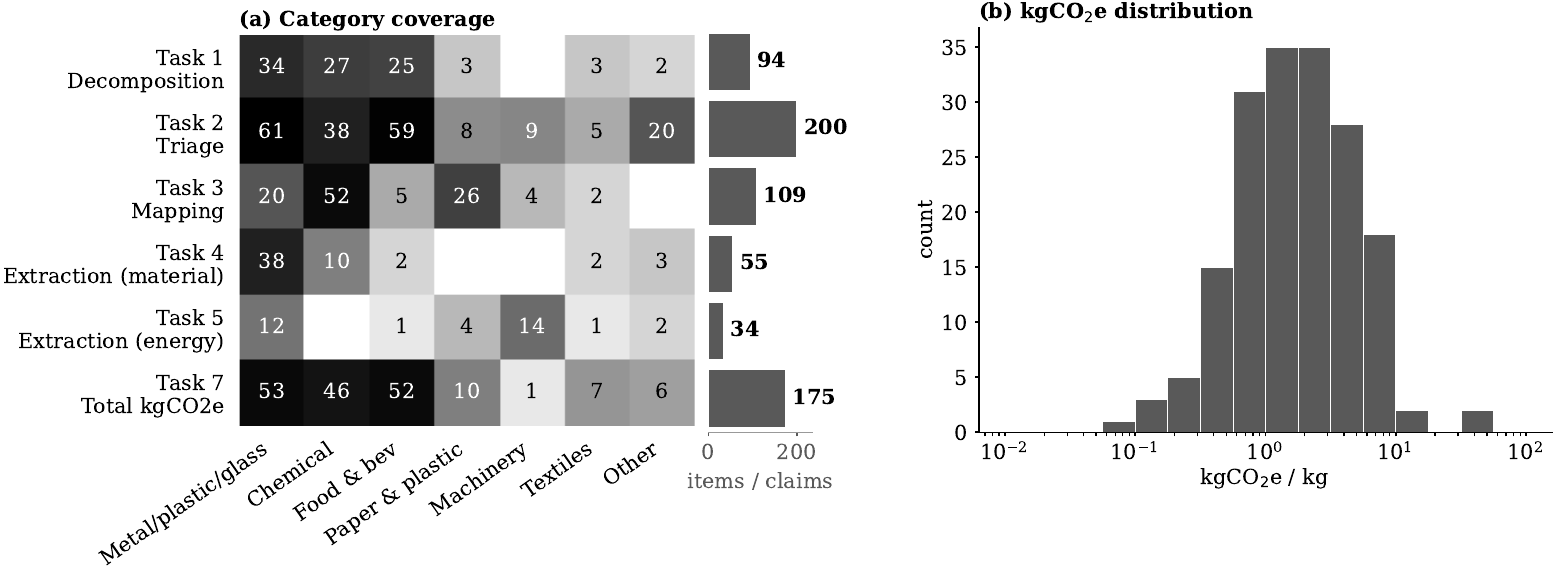}
\caption{%
    \pcfbench{} dataset overview.
    (a) \emph{Category coverage}: product categories across the six
    evaluable tasks.  Number and shading denote claim/item counts;
    colour encodes $\log(\text{count}+1)$.
    Furniture, Construction, Vehicles, Electricity-and-fuels, and
    Services are bucketed into ``Other'' (Appendix~\ref{sec:category_tagging}).
    (b) \emph{\kgco{} distribution}: log-scale histogram of \kgco{}
    per kg across the 175 EPDs in Task~7.
    }
\label{fig:coverage_heatmap}
\end{figure}

%% file: sections/results.tex
\section{Results and Discussion}
\label{sec:results}

\begin{figure}[t]
\centering
\includegraphics[width=\textwidth]{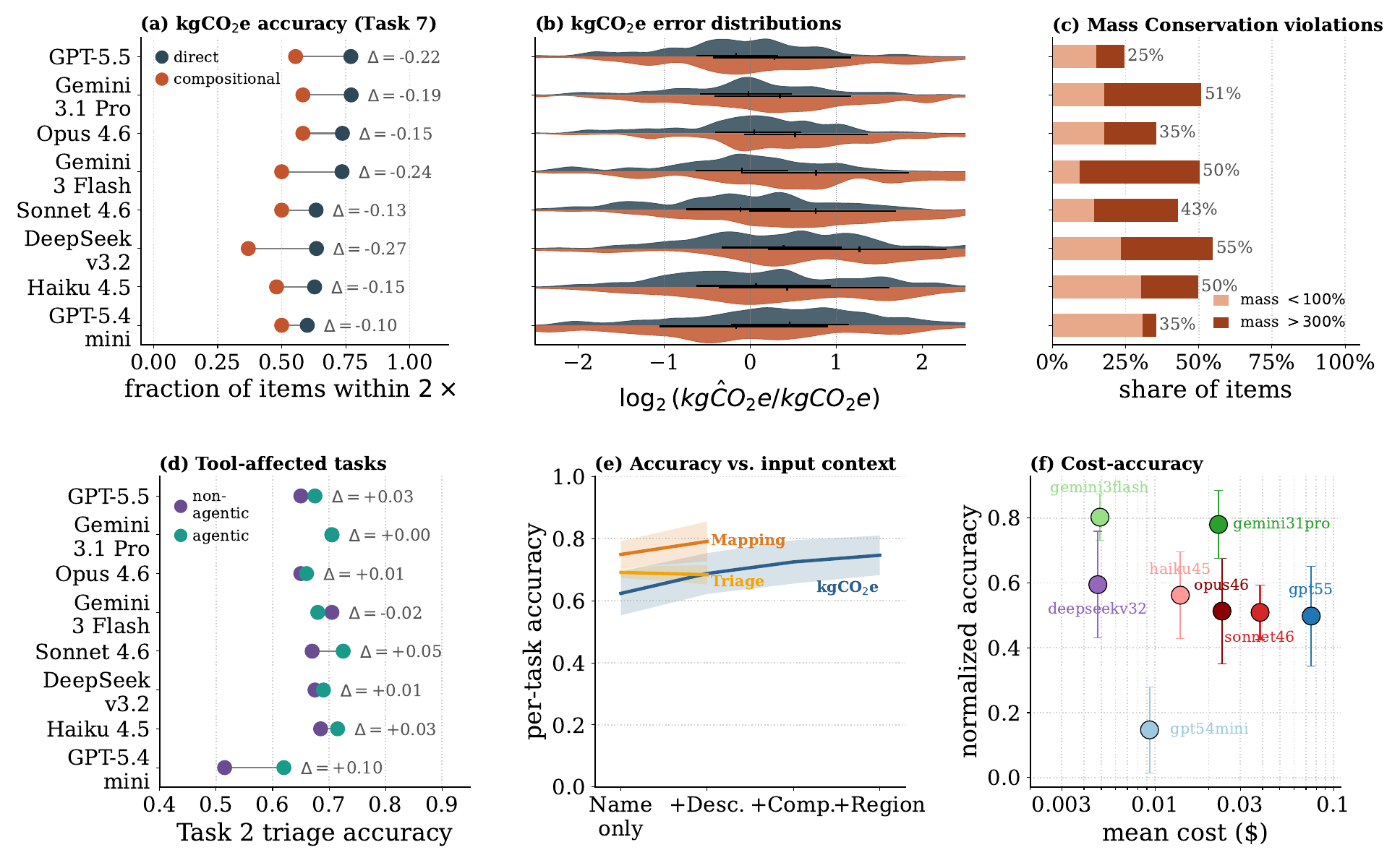}
\caption{Baseline model performance on \pcfbench{}.
(a) Task~7 \kgco{} within-$2\times$ rate: direct (one-call) vs
compositional (Tasks~1--5 chained, summed via deterministic
\ecoinvent{} EFs). Both modes start from product name and
description; the compositional pipeline derives composition through
Task~1.
(b) Per-model relative-error distribution; upper
half-violin~=~direct, lower~=~compositional.
(c) Mass-conservation violation rate (compositional only).
(d) Task~2 triage and Task~3 mapping: non-agentic vs agentic
accuracy.
(e) Accuracy vs increasing context across Tasks~2, 3, 7.
(f) Accuracy vs.\ cost.  $y$-axis is the mean of each model's
within-task min--max-normalised headline scores: for each of the six
tasks (T1, T2, T3, T4, T5, T7) we rescale the eight models' headline
metric to $[0, 1]$ via $(v - v_{\min}) / (v_{\max} - v_{\min})$ across
the eight models, then average a model's six rescaled scores.  Error
bars are the across-task standard error of that mean.  $x$-axis is
mean USD spend per benchmark item.}
\label{fig:performance_dashboard}
\end{figure}

\textbf{Off-the-shelf LLMs struggle to assemble compositional PCFs.}
We benchmark eight frontier LLMs from four providers (Anthropic,
Google, OpenAI, DeepSeek; setup in
Appendix~\ref{sec:task_eval_methodology}).  Given an EPD product's
name and description, the compositional pipeline lands within
$2\times$ of third-party-verified \kgco{} truth on only
$37$--$58\%$ of products (Table~\ref{tab:headline_results}
Compositional column; Figure~\ref{fig:performance_dashboard}a;
Appendix~\ref{sec:full_results}).
Opus~4.6 and Gemini~3.1~Pro lead at $58\%$, while DeepSeek lags at
$37\%$.

\begin{table}[t]
\centering
\caption{\pcfbench{} headline results.  Task~1 $F_1$, Task~2
accuracy, Task~3 exact-match (with-context detail in
Appendix~\ref{sec:app_mapping_context}), Task~4--5 claim-$F_1$,
Task~7 within-$2\times$ of declared \kgco{}; exact metric
definitions in Appendix~\ref{sec:task_eval_methodology}.
Tasks~2 and~3 report scores from the agentic harness (single-shot
variants in Appendix~\ref{sec:agentic_vs_singleshot}).  Direct Task~7:
directly predicts the \kgco{} of the EPD.  Compositional Task 7: an
agent assembles \kgco{} compositionally
by chaining \pcfbench{}'s task agents on the same EPDs (pipeline
spec in Appendix~\ref{sec:app_compositional}).
Both modes receive the same input: the product's name and description.  $\pm$ values are
bootstrap standard errors from 1{,}000 item-level resamples
(protocol in Appendix~\ref{sec:app_bootstrap}).
Parakeet$^\ddagger$ is a specialized baseline targeting mapping
only.}
\label{tab:headline_results}
\footnotesize
\setlength{\tabcolsep}{4pt}
\begin{tabular}{@{}lccccccc@{}}
\toprule
\textbf{Model} & \textbf{Task 1} & \textbf{Task 2} & \textbf{Task 3} &
  \textbf{Task 4} & \textbf{Task 5} & \textbf{Task 7} & \textbf{Task 7} \\
 & \textbf{Decompose} & \textbf{Triage} & \textbf{Map} &
  \textbf{Materials} & \textbf{Energy} & \textbf{Direct} & \textbf{Compositional} \\
 & $F_1$ & Acc. & EM & $F_1$ & $F_1$ & $\leq 2\times$ & $\leq 2\times$ \\
\midrule
GPT-5.5                    & 0.73{\scriptsize$\pm$0.02} & 0.68{\scriptsize$\pm$0.03} & 0.80{\scriptsize$\pm$0.04} & 0.31{\scriptsize$\pm$0.07} & 0.29{\scriptsize$\pm$0.06} & 0.77{\scriptsize$\pm$0.03} & 0.55{\scriptsize$\pm$0.04} \\
GPT-5.4-mini               & 0.70{\scriptsize$\pm$0.03} & 0.62{\scriptsize$\pm$0.03} & 0.75{\scriptsize$\pm$0.04} & 0.29{\scriptsize$\pm$0.08} & 0.48{\scriptsize$\pm$0.10} & 0.60{\scriptsize$\pm$0.04} & 0.50{\scriptsize$\pm$0.04} \\
Claude Opus 4.6            & 0.78{\scriptsize$\pm$0.02} & 0.66{\scriptsize$\pm$0.03} & 0.85{\scriptsize$\pm$0.03} & 0.27{\scriptsize$\pm$0.09} & 0.29{\scriptsize$\pm$0.05} & 0.73{\scriptsize$\pm$0.03} & \textbf{0.58{\scriptsize$\pm$0.04}} \\
Claude Sonnet 4.6          & 0.74{\scriptsize$\pm$0.02} & \textbf{0.72{\scriptsize$\pm$0.03}} & 0.82{\scriptsize$\pm$0.04} & 0.39{\scriptsize$\pm$0.08} & 0.39{\scriptsize$\pm$0.06} & 0.63{\scriptsize$\pm$0.04} & 0.50{\scriptsize$\pm$0.04} \\
Claude Haiku 4.5           & 0.73{\scriptsize$\pm$0.02} & 0.71{\scriptsize$\pm$0.03} & 0.79{\scriptsize$\pm$0.04} & 0.36{\scriptsize$\pm$0.06} & \textbf{0.53{\scriptsize$\pm$0.08}} & 0.63{\scriptsize$\pm$0.04} & 0.48{\scriptsize$\pm$0.04} \\
Gemini 3.1 Pro             & 0.74{\scriptsize$\pm$0.02} & 0.70{\scriptsize$\pm$0.03} & \textbf{0.93{\scriptsize$\pm$0.02}} & 0.44{\scriptsize$\pm$0.08} & 0.42{\scriptsize$\pm$0.08} & \textbf{0.77{\scriptsize$\pm$0.03}} & \textbf{0.58{\scriptsize$\pm$0.04}} \\
Gemini 3 Flash             & \textbf{0.80{\scriptsize$\pm$0.02}} & 0.68{\scriptsize$\pm$0.03} & 0.68{\scriptsize$\pm$0.05} & 0.47{\scriptsize$\pm$0.11} & 0.43{\scriptsize$\pm$0.07} & 0.73{\scriptsize$\pm$0.03} & 0.50{\scriptsize$\pm$0.04} \\
DeepSeek v3.2              & 0.74{\scriptsize$\pm$0.02} & 0.69{\scriptsize$\pm$0.03} & 0.84{\scriptsize$\pm$0.04} & \textbf{0.51{\scriptsize$\pm$0.10}} & 0.53{\scriptsize$\pm$0.07} & 0.63{\scriptsize$\pm$0.04} & 0.37{\scriptsize$\pm$0.04} \\
Parakeet$^\ddagger$ \citep{balaji2025parakeet} & -- & -- & 0.76{\scriptsize$\pm$0.04} & -- & -- & -- & -- \\
\bottomrule
\end{tabular}
\end{table}

Tasks~1--3 share a common failure mode: reasoning under under-specification.  Task~1
decomposition recall sits at $0.64$--$0.75$, so a quarter to a
third of the components experts list go missing, even as precision
stays high ($0.85$--$0.92$;
Table~\ref{tab:decomposition_results_full}, Appendix~\ref{sec:full_decomposition}).  On \emph{aluminium
foil}, Opus~4.6 returns 2 of the 5 expert-listed components; the
long tail that goes missing is mostly additives, finishes, and
alloying elements (a 50-item author audit on Gemini~3.1~Pro rules
out a judge artefact; $88\%$ agreement,
Appendix~\ref{sec:app_judge_audit}).  On Task~2 triage, even the best agentic baseline reaches only
$0.725$ (Sonnet~4.6), with single-shot accuracy spanning
$0.515$--$0.705$ (Table~\ref{tab:agentic_t2_split}).  In Task~3,
on challenging items like
\emph{Hydroformylation catalyst (Rh/Co + ligands)}, only
Gemini~3.1~Pro picks ``chemical, organic'' while others default
to ``rhodium'' (a trace component), over-estimating by
${\sim}10{,}000\times$.

The second failure mode is reasoning under numerical constraints:
BOMs fail basic mass-balance checks.  On $25$--$55\%$ of products,
the agents
produce a BOM that fails mass conservation
(Figure~\ref{fig:performance_dashboard}c; rate defined in
Appendix~\ref{sec:app_compositional}). Further, on $47$--$86\%$, they
include at least one ``ghost'' component (components with zero mass; Appendix
Figure~\ref{fig:ghost_components}).  Both biases push final \kgco{} estimates
downward.

By contrast, when the same eight models are asked for \kgco{}
\emph{directly}, they reach $60$--$77\%$ within-$2\times$ on the
same EPDs under identical inputs (Table~\ref{tab:headline_results} Task~7 column).
Frontier LLMs appear stronger at order-of-magnitude estimation than
at compositionally assembling a PCF.
However, such direct estimates cannot replace compositional PCFs.
PCFs are useful only when their structure is
transparent enough to let users identify hotspots and evaluate abatement
levers. Still, LLMs' relative strength at order-of-magnitude estimation
suggests a natural future direction. Direct estimates could sanity-check
the compositional pipeline by reconciling its sum against
a direct guess.

\textbf{More relevant context helps, but gains saturate.}
Material composition contributes the largest accuracy gain on
Task~7 direct estimate: about $10$ points of within-$2\times$ \kgco{} on average and
a flip from systematic name-only over-prediction to roughly zero
(Table~\ref{tab:epd_results_full}, Appendix~\ref{sec:full_epd};
Figure~\ref{fig:performance_dashboard}e;
Appendix Figure~\ref{fig:task7_predicted_vs_actual}).  Adding region on top
yields little further movement, and the gain concentrates in
trimming $>$2$\times$ outliers rather than pulling typical-product
errors to zero.  The shift is most striking on products whose name
conceals a recycled or filler-rich composition; on
\emph{Aluminium ingot~--~Semi-Primary}, Opus~4.6 predicts $4.5$
\kgco{}/kg from the name alone against a declared $0.94$, and snaps
to $0.82$ once $90.8\%$ post-consumer scrap is disclosed.

Task~3 mapping shows the same arc with vagueness as the binding
axis (Appendix Figure~\ref{fig:mapping_vagueness},
Table~\ref{tab:full_mapping_scores}, Appendix~\ref{sec:full_mapping};
$61\%$ of items at severity~$\geq 4$).  Supplier or purchaser context can flip a
mapping outright: \texttt{"PCA"} plausibly maps to polycarbonate,
poly citric acid, etc., and only the receiving-supplier context
(``water bottle maker'') disambiguates toward
polycarbonate~(\S\ref{sec:dataset_mapping}).

\textbf{Tool access helps, unevenly across models.}
Adding the agentic harness (Appendix~\ref{sec:app_agentic_harness})
widens the LLM lead on Task~3 mapping
(Figure~\ref{fig:performance_dashboard}d), but unevenly: five of
eight models gain on Task~3 (DeepSeek~v3.2 the most at $+0.15$,
Gemini~3.1~Pro reaches $0.927$), while Gemini~3~Flash loses $0.13$.
On Task~2 triage, the agentic harness yields only marginal gains at
the top: best non-agentic accuracy is $0.705$ (Gemini~3.1~Pro and
Gemini~3~Flash, tied) and best agentic is $0.725$ (Sonnet~4.6,
Table~\ref{tab:agentic_t2_split}).  Format compliance stays $\geq
98\%$ across all baselines (Table~\ref{tab:triage_results_full},
Appendix~\ref{sec:full_triage}), so errors come from the triage
decision rather than invalid structured output.
Parakeet's specialized 3-stage retrieve-rerank pipeline
(\citealp{balaji2025parakeet}; Table~\ref{tab:headline_results},
$\ddagger$) sits behind frontier LLMs on the v3.11 picklist: its
GTE-based narrowing to ten candidates strands matches that the
full-picklist non-agentic LLM still finds.

\textbf{All LLMs struggle with scientific document extraction.}
Tasks~4 and~5 hold the eight baselines in a narrow claim-$F_1$ band
well below the other tasks (material $0.27$--$0.51$, energy
$0.29$--$0.53$; Table~\ref{tab:extraction_results_full},
Appendix~\ref{sec:full_extraction}).  The binding constraint is
claim precision: models over-extract.  Opus~4.6 is an extreme
example: asked for
injection-molding specific energy ``excluding upstream polymer
production,'' it returned the six expert values plus thirty extras,
including $81.04$~MJ/kg with polymer production included.  Some of
Opus's extras may be candidates that our unanimous-include filter
rejected; however, only $9$ claims were dropped across all
36~queries under that filter
(Appendix~\ref{sec:dataset_extraction})---well below the $30$
extras Opus produced on this single query.  This over-extraction
also explains why the error distribution still skews positive
(Figure~\ref{fig:performance_dashboard}b) despite the downward
drag from missing mass and ghost components. Appendix 
Figure~\ref{fig:annotation_example} shows why evidence matters:
extractions returned $4.6$~t/ha (a 10-year amortised yield) when
the query asked for the $463.4$~kg/ha annual rate. A value-only
ground truth would
have missed this scope-qualifier mismatch.

\textbf{Jagged frontier across cost and accuracy.}
No single model dominates: six different systems take the per-task
crowns across the six tasks (Table~\ref{tab:headline_results},
Figure~\ref{fig:per_task_radar}), and the leaders cut across
providers and price tiers.  The cost--accuracy frontier across the
eight evaluated models is non-monotone
(Figure~\ref{fig:performance_dashboard}f). Gemini~3~Flash tops both the
mean min-max-normalised accuracy ranking and accuracy per dollar.
Opus~4.6 lands cheaper per benchmark item than Sonnet~4.6 despite a
higher per-token rate, due to Sonnet's lengthier tool chains.

\textbf{\pcfbench{} surfaces general LLM weaknesses.}
The failure modes \pcfbench{} exposes may generalize beyond PCFs.
Tasks~2 and~3 require \emph{reasoning under under-specification}: a
single token (\texttt{"PCA"}) admits multiple defensible mappings
until disambiguating context arrives
\citep{wen2025knowyourlimits, li2025questbench, wilie2024belief}.
Tasks~4--5 surface \emph{scope-qualifier failure}: models
over-extract claims for adjacent questions or time scales, dragging
claim precision down (Table~\ref{tab:extraction_results_full}).  The
compositional pipeline reveals \emph{soft-constraint failure}:
outputs satisfy syntactic structure (the BOM parses, the items are
plausible) without satisfying mass conservation ($25$--$55\%$ mass
violations in Figure~\ref{fig:performance_dashboard}c, $47$--$86\%$
ghost rate in Appendix Figure~\ref{fig:ghost_components}).  Task~2
triage exposes \emph{abstention failure}: when no \ecoinvent{} match
exists for a composite item, models default to a confident
map-or-decompose call rather than recognising the no-match condition
\citep{kirichenko2025abstentionbench}. Calibration mirrors
the split: pooled across baselines, mapping is approximately
calibrated (ECE $0.086$) while triage is severely overconfident
(ECE $0.258$; Appendix Figure~\ref{fig:calibration_curves}, 
Appendix~\ref{sec:confidence_calibration}).  The
same pattern of decompose-and-diagnose failures can transfer to other expert
workflows such as drug discovery, financial auditing, and legal due
diligence, where errors compose multiplicatively.

\section{Limitations and Future Work}
\label{sec:limitations}
\begin{enumerate}[leftmargin=*,nosep,itemsep=0pt]
\item \textbf{Dataset scale.}
\pcfbench{} contains 614 items across six tasks, which is small relative to
benchmarks that draw on synthetic or crowd-sourced labels.  Expert
annotation drives this cost: each item passes through third-party EPD
verification, expert-curated mapping, or multi-source claim
adjudication (Appendix~\ref{sec:annotation} gives the full procedure).
Annotation depth, including accepted mapping alternatives, vagueness
severity, challenge-type tags, and four context-ablation settings, supports
stratified analyses, but fine-grained sub-strata ($<10$ items) lack
statistical power.
\item \textbf{Scope boundary.}
All tasks target cradle-to-gate.  Use-phase and end-of-life impacts,
which dominate certain product categories, are not covered.
\item \textbf{Document retrieval not evaluated.}
Tasks~4--5 measure value extraction \emph{given} the relevant technical
document; locating that document upstream---from open literature,
supplier datasheets, industry handbooks, or trade databases---is out of
scope.
\item \textbf{Tasks are not chained per product.}
The five datasets come from independent sources: extraction PDFs are
not paired with EPDs, and BOMs, triage decisions, and mappings are
not labelled per EPD.  The compositional pipeline therefore cannot be
scored against ground-truth intermediates end-to-end, and Tasks~4--5
cannot be folded into compositional Task~7.  A per-product chain is a
natural future extension.
\item \textbf{Database specificity.}
Ground-truth mappings are drawn from \ecoinvent{}~v3.11
\citep{ecoinvent}; results may not transfer to other background databases 
like GaBi \citep{gabi},
USLCI \citep{uslci}, TianGong \citep{tiangong}, etc., which use
different process nomenclatures and allocation methods.
\item \textbf{Language coverage.}
All documents and prompts are English. Multilingual extraction
performance may differ substantially.
\item \textbf{Emission-factor error not measured.}
Task~3 mapping is scored on activity-string match against
expert-curated picks, not on the emissions impact associated with those
picks. An emissions-weighted variant can be explored in the future.
\item \textbf{Contamination.}
The 175~EPDs and 36~technical documents are public and may appear in
pre-training corpora. We cannot rule out memorisation-driven gains.
\item \textbf{Calibration as a routing signal.}
Confidence calibration varies sharply across sub-tasks (mapping is
approximately calibrated; triage is severely overconfident; Appendix
Figure~\ref{fig:calibration_curves}).  Calibrating per-step
confidence and using it to flag items for expert review could be
an opportunity for human-in-the-loop deployment.
\end{enumerate}
Dual-use considerations, including greenwashing, erroneous emissions propagation,
and regulatory misuse, are detailed in
Appendix~\ref{sec:broader_impact}.

\section{Conclusion}
\label{sec:conclusion}
\pcfbench{} is the first decomposed benchmark for AI-generated PCF
estimation, evaluating each sub-task of an expert workflow with
task-specific metrics on expert-annotated data.
As AI systems increasingly mediate corporate climate
disclosures, \pcfbench{} offers a shared yardstick for ensuring the
emissions numbers they produce are transparent enough to expose their
compositional steps and trustworthy enough to drive real
decarbonization. More broadly, the six tasks that \pcfbench{} helps evaluate
span diverse skills
 (reasoning under under-specification, long-document numerical extraction,
 and order-of-magnitude estimation; Table~\ref{tab:headline_results},
Appendix Figure~\ref{fig:per_task_radar}). \pcfbench{} can thus operate as
a multi-axis diagnostic for frontier LLMs.

%% file: sections/appendix.tex
\section{Total \kgco{} Prediction Protocol}
\label{sec:integration}

The per-task evaluations test sub-task capability in isolation.
Total-\kgco{} prediction answers the complementary question: when the
model is asked for a single overall footprint number, does it produce
a defensible PCF estimate?  We use 175~EPDs registered with the
International EPD System \cite{environdec}, third-party verified under
ISO~14025, as ground truth, evaluated under \emph{four
context-ablation settings} that strictly add information at
each step:
(1)~\emph{name only} (product name + declared unit, baseline physical
intuition);
(2)~\emph{with description} (free-text);
(3)~\emph{with composition} (material mass breakdown + recycled
content);
(4)~\emph{with region} (manufacturing geography).
The progression isolates which information channel each model
exploits.  Combined with optional expert-extraction substitution on a
per-EPD basis, it also supports error-attribution analyses that
scoring only the aggregate output cannot perform.  All ground-truth labels
are produced by sustainability analysts and LCA experts, not
crowdworkers (Appendix~\ref{sec:annotation}).

\section{Dataset Details}
\label{sec:dataset_details}

\subsection{Task 1: Product Decomposition Dataset}
\label{sec:dataset_decomposition}

The decomposition dataset evaluates whether models can produce a usable
bill of materials (BOM) from a product name and free-text description.
Given the product, the model must list up to 8 input materials at the
granularity an LCA practitioner would source from a supplier or
background database (e.g.\ ``polyethylene'', ``aluminium scrap'',
``flat glass''), ordered from highest to lowest mass contribution.
Input rates and percentages are explicitly excluded.  This task isolates
the \emph{compositional knowledge} step of the pipeline from the
downstream quantitative steps.

\paragraph{Source and collection.}
The dataset is the 94-item subset of the Task~7 EPD corpus for which
the underlying EPD reports an explicit material-composition breakdown.
We parse the composition block out of the EPD description, retain the
component names ordered by mass contribution, and discard the
percentages.  The composition block and the embedded
``Product attributes'' block are stripped from the input description so
the model cannot copy the BOM directly from the prompt.  The remaining
input is the product name plus a free-text description of the product's
function, manufacturing process, and intended use, the same information
an LCA practitioner is given at the start of an analysis.

\paragraph{Scope: packaging.}
Consumer packaging, meaning the container that delivers the product to the
end user, e.g.\ a juice carton holding juice or a tinplate can holding
canned chickpeas, is in scope and counts as a BOM component.
Distribution packaging, such as wooden pallets, shrink wrap, or cardboard
master cartons that exist only to move product between facilities and
are discarded before reaching the consumer, is out of scope, as is
standard for cradle-to-gate PCFs \citep{ghgprotocol, iso14067}.

\paragraph{Schema.}
Each item contains \texttt{product\_name}, \texttt{description} (with
composition and product-attribute blocks removed), and
\texttt{quantity\_unit} (almost always ``kilogram'').  The expected
output is \texttt{components}: an ordered list of input material names
from the source EPD (median 4, max 8 components per item).

\paragraph{Example.}
``Aluminium plates and billets in 6060 green B~80 alloy''.  Expected
components, ordered by mass: \texttt{Aluminium scrap (pre-consumer)},
\texttt{Primary aluminium}, \texttt{Aluminium scrap (post-consumer)},
\texttt{Remelted ingots}, \texttt{Alloying elements}.  A model must
recognise from the product name that the BOM is dominated by recycled
aluminium scrap and primary aluminium ingot, plus alloying elements,
without seeing the percentages.

\paragraph{Evaluation metric.}
Because predicted and expected component names rarely match
character-for-character (``polyethylene'' vs.\ ``PE'' vs.\
``polythene''), and because models legitimately decompose products at
different fabrication levels (predicted ``concrete'' against expected
\{sand, water, cement\}; predicted \{iron, carbon\} against expected
``steel''), we use a Gemini~2.5~Flash judge to align predicted and
expected components into \emph{compositional match groups}: 1-to-1,
1-to-N (over-aggregation), N-to-1 (over-decomposition), or N-to-N
(parallel coverage at compatible fabrication levels), each required
to be compositionally consistent at the LCA practitioner level.  From the alignment we compute precision, recall, $F_1$,
Kendall~$\tau$ over the per-group mean indices, and exact-set match.
The judge prompt and credit policy are documented in
Appendix~\ref{sec:app_decomposition} and Appendix~\ref{sec:prompts}.

\paragraph{Limitations.}
(1)~The dataset reuses the EPD-with-composition subset of Task~7;
results on Task~1 and the ``with composition'' variant of Task~7 are
therefore evaluated on the same 94 products, and cross-task analyses
must account for this overlap.
(2)~At 94 items the dataset is small relative to the diversity of
real-world products, but it is the largest subset of the EPD corpus
with expert-curated composition breakdowns.
(3)~The judge introduces some non-determinism into scoring; we use
temperature~0 and cache the judge call per (predicted, expected) pair
to limit drift.

\begin{figure}[h]
\centering
\includegraphics[width=0.55\linewidth]{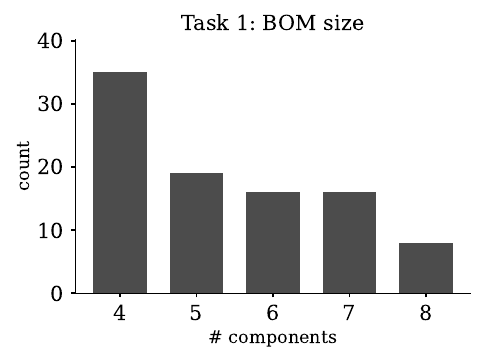}
\caption{Task~1 BOM size distribution across the 94 products.}
\label{fig:dataset_task1_bomsize}
\end{figure}

\subsection{Task 2: Mapping Triage Dataset}
\label{sec:dataset_triage}

The mapping triage dataset evaluates whether models can make the
recursive routing decision an LCA practitioner faces at every level
of a product's bill of materials: given a candidate material and the
root-material context in which it is being expanded, should it be
mapped directly to an \ecoinvent{} market activity
(\texttt{should\_map\,=\,true}), or should it be further decomposed
into sub-components (\texttt{should\_map\,=\,false})?

\paragraph{Input shape.}
Each item exposes five fields: a candidate material (\texttt{name},
\texttt{description}) and the root-material context for that
decision (\texttt{name}, \texttt{description},
\texttt{material\_name}).  Surrounding state is stripped so that
triage performance reflects market-and-root-material reasoning
rather than memorisation of incidental fields.

\paragraph{Coverage.}
The released set is balanced 100/100 and covers 184~unique markets,
163~unique root materials, and seven of the eight reporting buckets
(\S\ref{sec:category_tagging}); only Infrastructure \& buildings is
absent. 

\paragraph{Labels.}
Each item carries a sustainability-expert--labelled \texttt{should\_map}
bool: \texttt{should\_map=true} when the market is well-served by one
or more existing \ecoinvent{}-backed library materials, and
\texttt{should\_map=false} when it instead needs to be decomposed
further into a sub-BOM.  The released set is balanced 100/100 by
construction.

\paragraph{What it tests.}
The task probes whether models, given the five-field market-plus-context
input, make the right map-vs-decompose call.  A model that always maps
would force bad matches for complex products (e.g., ``printed circuit
board'' $\to$ a generic electronics process rather than decomposing
into copper, fibreglass, solder, etc.); a model that always decomposes
would waste effort on simple materials with direct \ecoinvent{}
matches.  Performance is bounded below by abstention-calibration
weaknesses documented elsewhere
\citep{kirichenko2025abstentionbench, li2025questbench}.

\subsection{Tasks 4 \& 5: Physical Parameter Extraction Dataset}
\label{sec:dataset_extraction}

\paragraph{Source and collection.}
The extraction dataset draws from 36 technical PDF documents: carbon
footprint reports, Environmental Product Declarations, academic
papers, and industry handbooks.  The documents span manufacturing
domains such as aluminum smelting, plastics processing, textiles,
food processing, and construction materials.  For each document, an LCA
practitioner authored a natural-language query specifying the
parameters needed for a carbon footprint calculation (e.g., ``What is
the recycled content of 3003 aluminum alloy used in foil
production?'').

\paragraph{Annotation pipeline.}
For each query, sustainability experts reviewed candidate claims
drawn from the source document and labelled each \texttt{include},
\texttt{exclude}, \texttt{duplicative}, or \texttt{irrelevant}; up to
three experts adjudicated each query.  Per-annotator latest-decision
dedupe is applied first, then cross-annotator unanimous-on-include
voting (\texttt{duplicative}~$\to$~\texttt{include},
\texttt{irrelevant}~$\to$~\texttt{exclude}; ties drop), then a
within-document $(\text{rounded value}, \text{normalized unit})$
dedupe with evidence union.  The post-merge dataset retains
89~ground-truth claims across the 36~documents.

\paragraph{Schema.}
Each retained ground-truth claim is stored as
\texttt{\{value: float, unit: str, evidence: list[str]\}}.  The
\texttt{(value, unit)} dedupe intentionally collapses the upstream
\texttt{parameter\_name} distinction so a model producing the right
number with the right unit is correct regardless of how it labels the
parameter.  The \texttt{evidence} list is the union of verbatim
quotes from every annotator's per-source \texttt{claim\_id} and
the bundle's \texttt{joined\_claim\_groups.json} sibling refs (see
\textbf{Statistics} below for counts).

\paragraph{Statistics.}
After per-annotator-latest dedupe and cross-annotator unanimous
voting (\texttt{duplicative}~$\to$~\texttt{include},
\texttt{irrelevant}~$\to$~\texttt{exclude}), the source pool yields
89 retained ground-truth claims across 36 documents.  At the item
level, 22 documents are labelled \emph{material} and 14 are labelled
\emph{energy} based on the extraction query; the dataset is split at
score time by item tag rather than per-claim unit categorisation.  Within each item, claims sharing
$(\text{value}, \text{unit})$ collapse to one with unioned evidence
(intentionally destroying the per-\texttt{parameter\_name}
distinction so a model producing the right number with the right
unit is correct regardless of how it labels the parameter).
Material units span percent/mass\%/wt\%, kg/kg, g/kg, agricultural
rates (\texttt{kg N/ha}, \texttt{t/ha}), and process parameters
(\texttt{°C}, \texttt{h}); energy units span MJ/kg, kJ/kg, J/g,
kWh/kg, BTU/lb, GJ/t, and grid carbon intensity (\texttt{gCO2/kWh}).
Three documents lost all their claims under unanimous voting and are
archived: HDPE extrusion energy total, secondary-aluminum-US energy
specific, and powder coating curing oven gas consumption.  Each
retained claim carries a list of verbatim evidence quotes (mean
$\sim$2.6 quotes per claim, 227 total) sourced from the
\texttt{aligned\_text} field of the bundle's
\texttt{claim\_sources/*.json} \texttt{supports[]} entries, unioned
across every annotator-included source via the bundle's
\texttt{joined\_claim\_groups.json} sibling refs.  A standalone audit
confirmed all 227 evidence quotes are substrings of their item's
\texttt{document\_text} after whitespace normalization (0/227
violations).

\begin{figure}[h]
\centering
\begin{minipage}[b]{0.32\linewidth}
\centering
\includegraphics[width=\linewidth]{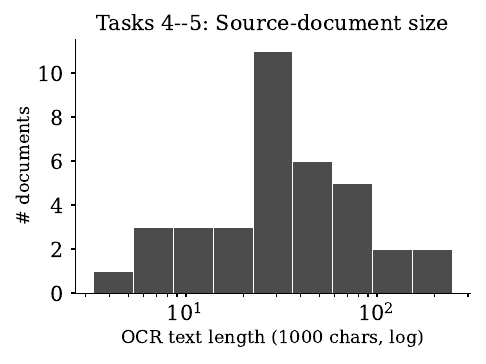}
\end{minipage}\hfill
\begin{minipage}[b]{0.32\linewidth}
\centering
\includegraphics[width=\linewidth]{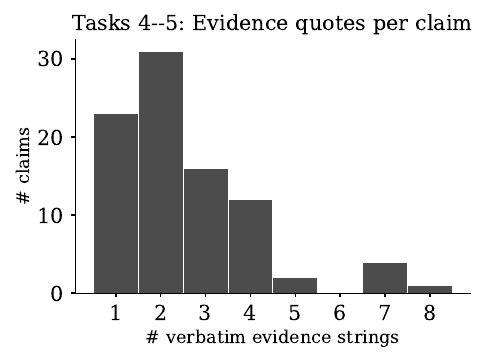}
\end{minipage}\hfill
\begin{minipage}[b]{0.32\linewidth}
\centering
\includegraphics[width=\linewidth]{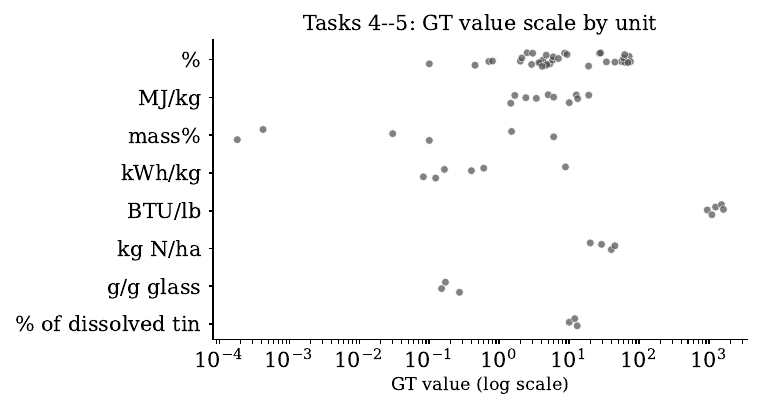}
\end{minipage}
\caption{Tasks~4--5 dataset characterisation.
\textbf{Left:} OCR'd source-document length per item, in thousands of
characters (log scale).  Median 30.5k, P90 98.1k chars.
\textbf{Centre:} number of verbatim evidence quotes per ground-truth
claim (mean 2.6, max 8, 227 total).
\textbf{Right:} ground-truth value scale per unit family (top-8 by
claim count, log scale).  Material claims (\%, mass\%, kg/kg, g/kg)
cluster within $\sim$two orders of magnitude; energy claims
(MJ/kg, kWh/kg) span $\sim$three.}
\label{fig:dataset_extraction_extras}
\end{figure}

\paragraph{Limitations.}
The dataset size (36~documents, 89~claims) is modest; expanding it
requires expert-authored queries and multi-stage annotation.  The
documents are English-language PDFs, limiting evaluation of
multilingual extraction.  Source PDFs are archived in GCS but not all
have stable public URLs, which may complicate full reproducibility.
The unanimous-voting rule is conservative; an additional 9~groups
that majority-but-not-unanimously included are dropped (these cases
typically involve one annotator marking \texttt{irrelevant} or
\texttt{duplicative} while the others substantively included).

\begin{figure}[h]
\centering
\begin{minipage}[b]{0.48\linewidth}
\centering
\includegraphics[width=\linewidth]{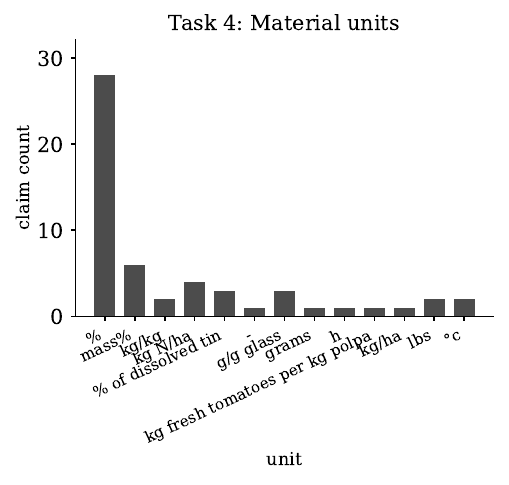}
\end{minipage}\hfill
\begin{minipage}[b]{0.48\linewidth}
\centering
\includegraphics[width=\linewidth]{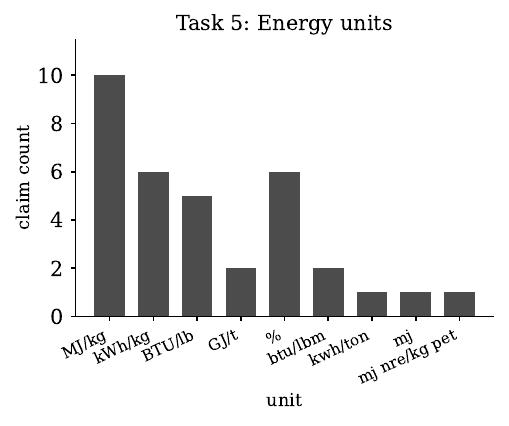}
\end{minipage}
\caption{Tasks~4 and~5 unit-family distribution across the 55~material and
34~energy ground-truth claims.  Cross-family energy units canonicalise to
MJ/kg via documented conversion factors at score time.}
\label{fig:dataset_task4_5_units}
\end{figure}

\subsubsection{Worked Example: Same Page, Different Decision}
\label{sec:annotation_example}

To make the annotation protocol concrete, we walk through one
extraction query whose candidate-claim set contains both an
\texttt{include} and an \texttt{exclude} decision, drawn from
adjacent rows of the same table on the same page.  Both candidates
are point values, both expressed in well-formed mass-per-area units,
and both are present and numerically extractable in the source.  Only
one is the right answer to the query, and the difference is a scope
qualifier (\emph{annual} vs.\ \emph{ten-year amortised}) that is
invisible from value and unit in isolation.  This is exactly the
nuance the locator-grounded schema is designed to surface.

\paragraph{Document.}
\textit{``The dollars and sense of liming: The Stantons' story''}
(Farm Facts series), Primary Industries and Regions South Australia
(PIRSA), 2017 \citep{pirsa2017liming}.  Internal request id
\texttt{req\_af88d019da0b9460}; question key
\texttt{lime\_maintenance\_rate}.

\paragraph{Query.}
\begin{quote}\itshape\small
What is the annual maintenance lime application rate in kg per
hectare or tonnes per hectare for Australian grain farming systems?
Report the maintenance liming rate, unit, and whether this is an
annual application or amortized over multiple years.
\end{quote}

\paragraph{Per-claim schema.}
Every candidate is stored as a small JSON record with three required
fields and one repeated nested field:

\begin{itemize}[nosep,leftmargin=2em]
\item \texttt{value} (numeric) and \texttt{unit} (string): the
  point value and its measurement unit, exactly as a downstream LCA
  model would consume it.
\item \texttt{parameter\_name}: a one-sentence label that names
  what is being measured, including any scope qualifiers
  (\emph{annual}, \emph{ten-year}, \emph{cradle-to-gate}, etc.).
\item \texttt{supports}: a list of \emph{evidence locators}, one
  per supporting span in the source PDF.  Each locator carries
  \texttt{block\_id}, \texttt{page\_number} (0-indexed),
  \texttt{proposed\_text} (the verbatim string DocAI parsed from that
  block), and \texttt{overlay\_box} with normalised page-relative
  quadrilateral vertices.
\end{itemize}

The reviewer then writes a \texttt{disposition $\in$ \{include,
exclude, irrelevant, duplicative\}} and, when the disposition is not
\texttt{include}, a free-text \texttt{notes} field with the reason.
The locator is what lets an auditor re-derive the value from the
underlying figure or table cell rather than trust the candidate's
free-text rendering.

\begin{figure}[t]
\centering
\includegraphics[width=0.95\linewidth]{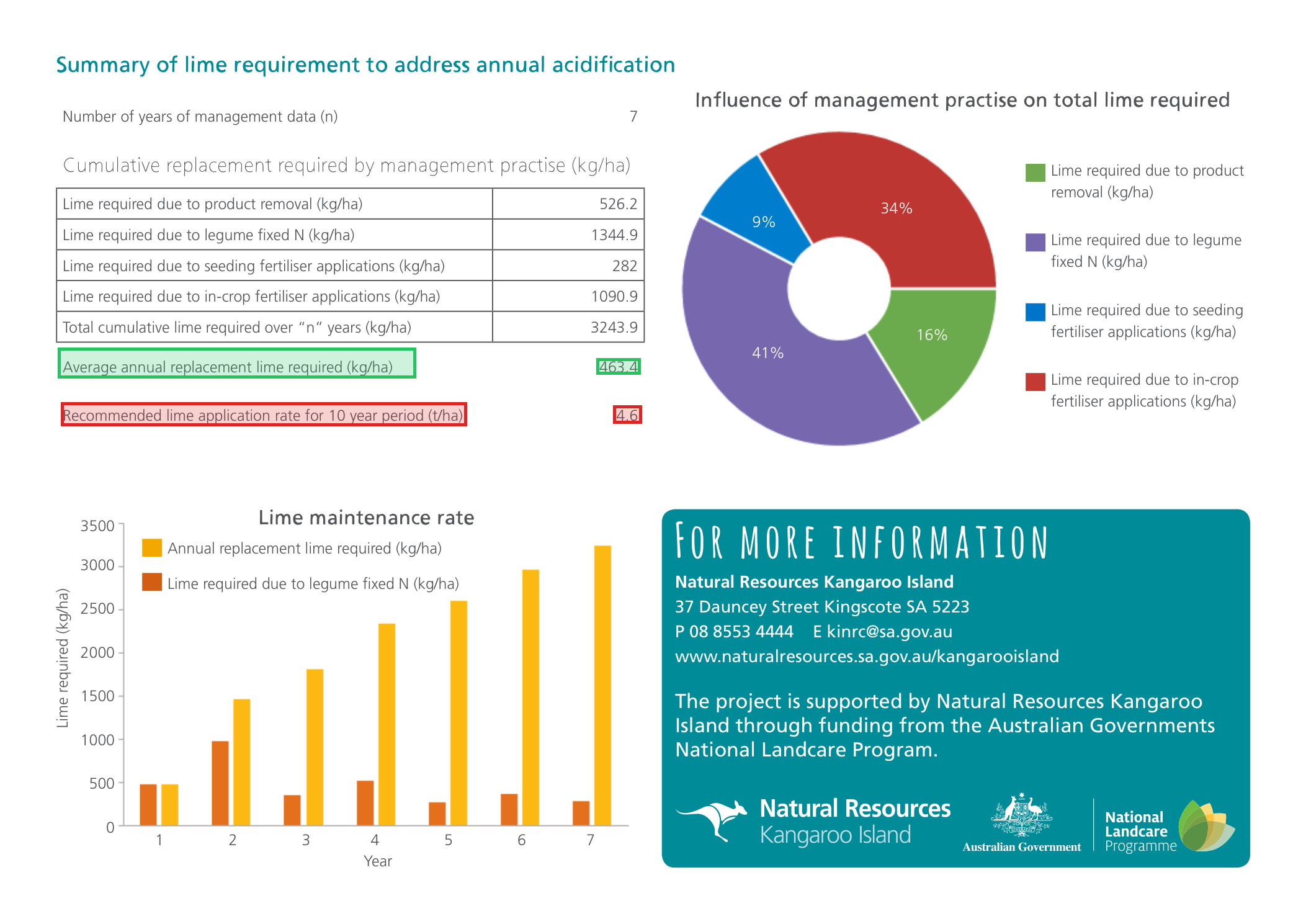}
\caption{%
    Annotation example for Tasks~4--5.  The full source page is
    page~6 of \citet{pirsa2017liming}, a one-page Farm Facts
    calculator that projects an annual liming rate forward to a
    ten-year amortised application.  Both candidate claims are
    grounded by block-level bounding boxes on this page.  The
    \textcolor{green!50!black}{\textbf{green}} boxes highlight the
    \textbf{included} claim, \texttt{463.4~kg/ha} (block~391), labelled
    by the row title ``Average annual replacement lime required
    (kg/ha)'' (block~387) immediately to its left.  The
    \textcolor{red!80!black}{\textbf{red}} boxes highlight the
    \textbf{excluded} claim, \texttt{4.6~t/ha} (block~393), labelled
    by the row title ``Recommended lime application rate for 10 year
    period (t/ha)'' (block~392).  Three of the four candidate
    extractions returned the
    red-box value as a candidate answer to the query (which asks
    specifically for the \emph{annual} rate); the reviewer
    rejected it with the note ``\emph{This number is for 10 years,
    question clearly asks for annual.}''  The numerical relationship
    between the two values is internal to the calculator
    ($463.4~\mathrm{kg/ha} \times 10 / 1000 \approx 4.6~\mathrm{t/ha}$),
    so a value-only ground truth would not have been able to flag the
    scope-qualifier mismatch; the locator + disposition + reviewer-note
    schema makes the rejection auditable from the primary evidence.}
\label{fig:annotation_example}
\end{figure}

\paragraph{The two candidate claims.}
Both candidates target the same row family of the same Farm Facts
results table on page~6 of the source PDF
(Figure~\ref{fig:annotation_example}):

\begin{itemize}[nosep]
\item \textbf{Included claim.} \\
  \emph{Parameter:} average annual replacement lime required to
  offset acidification, derived over seven years of management data. \\
  \emph{Value:} \texttt{463.4}~kg/ha. \\
  \emph{Evidence locator:} \texttt{block\_id}~=~\texttt{391},
  \texttt{page\_number}~=~5 (0-indexed), bounding box
  $(0.459,~0.391)$--$(0.488,~0.402)$ in landscape-frame normalised
  coordinates.  \texttt{proposed\_text}~=~\texttt{"463.4"}; the row
  label ``Average annual replacement lime required (kg/ha)'' is the
  adjacent block~387.  \\
  \emph{Reviewer note:} ``\emph{Evidence locator is misaligned,
  likely due to landscape orientation of the page.}''  (Annotator
  flagged a known DocAI-on-rotated-pages issue but verified the row
  manually before accepting the claim.)

\item \textbf{Excluded claim.} \\
  \emph{Parameter:} recommended total lime application rate amortised
  over a ten-year period. \\
  \emph{Value:} \texttt{4.6}~t/ha. \\
  \emph{Evidence locator:} \texttt{block\_id}~=~\texttt{393},
  \texttt{page\_number}~=~5 (0-indexed), bounding box
  $(0.472,~0.442)$--$(0.489,~0.455)$.
  \texttt{proposed\_text}~=~\texttt{"4.6"}; row label
  ``Recommended lime application rate for 10 year period (t/ha)'' is
  block~392 directly to its left. \\
  \emph{Reviewer note (verbatim):} ``\emph{This number is for 10
  years, question clearly asks for annual.}''
\end{itemize}

Both bounding boxes lie in the same column of the same Farm Facts
results table; the only thing distinguishing them is which row of
that table the bounding box falls on.  The numerical relationship is
internally consistent: the excluded value is exactly the included
value scaled by ten and converted from kg/ha to t/ha
($463.4~\mathrm{kg/ha} \times 10 / 1000 = 4.634~\mathrm{t/ha}
\approx 4.6~\mathrm{t/ha}$), reflecting that the paper's calculator
projects an annual rate to a ten-year amortised application; only
the annual rate matches the query's scope qualifier.

\paragraph{Why the schema matters.}
A \emph{value-only} ground truth would have silently accepted any
model output in the neighbourhood of \texttt{4.6~t/ha} as a correct
answer to ``maintenance lime application rate'', because the unit and
order-of-magnitude both look right.  The
locator-plus-disposition schema makes the disagreement \emph{about
which row of which table is being read} explicit: the reviewer can
point at the bounding box on the page and say ``this is the ten-year
projection, not the annual rate'', and that rejection becomes part of
the dataset's traceable record.  This is the auditable substrate
Tasks~4 and~5 evaluate against, and it is what allows the benchmark
to score two things: whether a model produced a plausible number and
whether it read the row the query was actually asking about.

\subsection{Task 3: Background Database Mapping Dataset}
\label{sec:dataset_mapping}

\paragraph{Source and collection.}
109~material names hand-curated by sustainability experts, spanning
typical mappings, edge cases (identity mappings, abbreviations,
foreign-language names), and specialized challenge sets (catalysts,
packaging materials, no-good-match scenarios).  Each material name
reflects the kind of raw, unstructured input an LCA practitioner
encounters in a typical industrial bill of materials.

\paragraph{Schema.}
Each item provides: a \texttt{material\_name} (the raw input, e.g.,
``Drink cup lids'' or ``Borontrifluorid''), optional context fields
(\texttt{description}, \texttt{receiving\_supplier\_name},
\texttt{material\_group} hierarchy), and a ground-truth
\texttt{options} list of defensible \ecoinvent{} reference products.
The list is unordered: every option is treated as an equally valid
mapping at score time, and a prediction is correct if it matches any
option.  38~items include a
\texttt{banned\_substring} (a known-wrong mapping the model must
avoid) and 23~items include a \texttt{relevant\_substring} for
soft-match evaluation.

\paragraph{Example.}
The input \texttt{"Pca"} (vagueness severity 5/5) is a three-letter
abbreviation that could plausibly map to several different
\ecoinvent{} reference products: \texttt{polycarbonate},
\texttt{polyethylene terephthalate, granulate, bottle grade,
recycled}, or \texttt{polyethylene terephthalate, granulate,
amorphous, recycled}.  All three are listed as acceptable options;
the receiving-supplier context (``water bottle maker'') is the
disambiguating signal a model needs to pick the PET variants over
polycarbonate.  This item is tagged \texttt{vague\_input} and
\texttt{org\_context\_should\_influence\_mapping}.

\begin{figure}[h]
\centering
\includegraphics[width=0.55\linewidth]{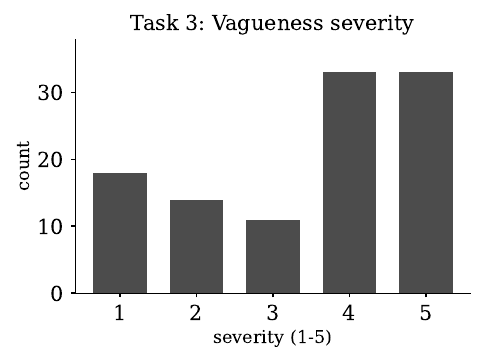}
\caption{Task~3 expert-rated vagueness severity (1\,=\,clear,
5\,=\,maximally ambiguous).  The dataset skews hard:
61\% of items are severity $\geq$\,4.}
\label{fig:dataset_task3_vagueness}
\end{figure}

\paragraph{Difficulty distribution.}
Expert-rated vagueness severity spans 1--5, intentionally skewed
toward harder items: 30\% of items are severity~5 and 30\% are
severity~4, compared to 17\% at severity~1
(Figure~\ref{fig:dataset_task3_vagueness}).  Items also include
specialised challenge sets such as catalyst mappings, no-good-match
items testing graceful degradation, vague inputs, abbreviation
resolution, and foreign-language names.

\paragraph{Multi-option structure.}
58\% of items have exactly one acceptable option; 29\% have two;
the remainder have 3--5, reflecting genuine ambiguity in the
mapping space.

\paragraph{Limitations.}
At 109 items the dataset is small relative to the full \ecoinvent{}
vocabulary (${\sim}$10{,}000~activities).  Product category coverage
is uneven (Figure~\ref{fig:coverage_heatmap}): chemical (52),
paper/plastic (26), and metal/mineral/plastic/glass (20) products
dominate, while every other primary category has $\leq$5~items.
Infrastructure \& buildings and the \emph{Other} bucket are empty because 
a practitioner would not map a constituent material directly to an entire infrastructure or building activity, and 
other categories such as services and maintenance activities are not typically represented in cradle-to-gate EPDs. 
Only 24 of 109~items have a supplier, description, or
purchaser-context field populated, limiting the
context-ablation evaluation to a small subset.

\subsection{Task 7: Total \kgco{} Prediction (EPD Dataset)}
\label{sec:dataset_epd}

\paragraph{Source and collection.}
175~products with declared cradle-to-gate carbon footprints, all
sourced from published Environmental Product Declarations registered
with the International EPD System \cite{environdec}.  Each product's
\kgco{} value is third-party verified per ISO~14025 \cite{iso14025}.
Each item includes a \texttt{source\_url} linking to the original EPD
on environdec.com, ensuring full provenance traceability.

\paragraph{Schema.}
Each item provides: \texttt{product\_name} (canonical product name,
always populated), \texttt{description} (free-text product
description), \texttt{quantity\_unit} (declared unit),
\texttt{composition} (material mass breakdown, populated for
94~items), \texttt{geography} (manufacturing region, 94~items),
\texttt{recycled\_content} (81~items), \texttt{country\_of\_origin}
(populated for 12~items), and \texttt{source\_url} (environdec.com
link, all 175~items).  The ground truth is \texttt{kgco2e}: the
declared cradle-to-gate greenhouse gas emissions per declared unit.

\paragraph{Example.}
``Aluminium plates and billets in 6060 green B~80 alloy'': an
intermediate product made by melting aluminium scraps with alloying
elements.  The ``green B~80'' designation signals high recycled
content, which a model must recognize to avoid over-predicting the
footprint.  Declared value: 3.37~\kgco{} per kg.

\paragraph{Scale diversity.}
Ground-truth \kgco{} values span over five orders of magnitude:
from 0.062 (low-impact agricultural product) to 19{,}600 (industrial
equipment), median 1.76, with 95\% of items between 0.1 and
10~\kgco{}.  Figure~\ref{fig:coverage_heatmap} (right panel) shows the
log-scale distribution.  This range means evaluation metrics must be
scale-invariant; we use $|\log_{10}(\hat{y}/y)|$ as the primary error
metric (Section~\ref{sec:methodology}).

\paragraph{Limitations.}
All 175~items are sourced from published EPDs, but product category
coverage is uneven.  Only
12~items have an explicit \texttt{country\_of\_origin}, though
94~items have parsed \texttt{geography} fields from the EPD
description.

\subsection{Annotation Quality}

All datasets use expert annotation.  Extraction claims (Tasks~4
and~5) are reviewed and labelled by up to three sustainability
experts per query, with the locator-grounded schema described in
\S\ref{sec:dataset_extraction} so every \texttt{include}/\texttt{exclude}
decision is auditable against the source PDF.  Task~3 mappings are
produced by LCA practitioners who make these decisions in production
workflows; the multi-option structure (mean 1.5 options per item)
captures genuine ambiguity rather than forcing a single answer.  EPD
ground truths are third-party verified per ISO~14025
\cite{iso14025}, the strongest available standard for product carbon
footprint data \cite{cardoso2024epdharmonisation}.

\section{Product-Category Tagging}
\label{sec:category_tagging}

Every item in \pcfbench{} carries a \texttt{product\_category} metadata
tag.  We start from the 12-category \texttt{environdec.com} taxonomy
(Chemical, Construction, Electricity steam \& fuels, Food \&
beverages, Furniture \& other goods, Infrastructure \& buildings,
Machinery \& equipment, Metal/mineral/plastic/glass, Paper and
plastic, Services, Textiles \& apparel, Vehicles \& transport) and
report results across \emph{seven} primary categories (Chemical
products; Food \& beverages; Infrastructure \& buildings;
Machinery \& equipment; Metal, mineral, plastic \& glass products;
Paper and plastic products; Textiles, footwear \& apparel) plus an
\emph{Other} bucket aggregating the five low-coverage environdec
categories (Construction; Electricity, steam \& fuels; Furniture
\& other goods; Services; Vehicles \& transport equipment).  The
seven primary categories collectively account for $\geq$86\% of
items in every task (90--100\% on Tasks~1, 2, 3, and~7); bucketing
the long tail keeps the per-category sample
size large enough for stratified analysis without obscuring the
coverage gap. Infrastructure \& buildings remains empty across all
task rows because the EPDs for that category are typically in terms of
functional units that include use phase assumptions rather than declared units suited to the benchmark.
Tags are assigned per item by Gemini~2.5~Flash from the
available text fields: material name and description for
Tasks~2--3, product name and description for Tasks~1 and~7, and the
extraction query plus document head for Tasks~4--5.  Tags propagate
unchanged across dataset re-pulls.  Task~4 and Task~5 inherit their
tag from the parent extraction item and are counted separately in
Figure~\ref{fig:coverage_heatmap} by filtering each item's claims by
unit (material vs.\ energy).

\section{Ecoinvent Picklist Construction}
\label{sec:picklist}

Tasks~2 (triage) and~3 (mapping) both require a fixed list of
candidate \ecoinvent{} reference products: the model decides whether
a direct match exists (Task~2) or selects which option is best
(Task~3).  Naively exposing the full \ecoinvent{}~v3.11 vocabulary
($\sim$20{,}000 activities across many geographies and sub-types)
would (i)~exceed practical context limits and (ii)~mix mass-based
products with non-mass declared units the benchmark cannot score
against.  We therefore curate a 2{,}574-item picklist using a
deterministic cascade applied directly to the public
\ecoinvent{}~v3.11 \emph{Database Overview} workbook (sheet
\texttt{Cut-Off AO}).  The build script
\texttt{pcfbench\_external/picklist/} \texttt{build\_picklist\_json.py}
has no proprietary data dependency: it ingests only the
public xlsx and writes a three-field JSON record (activity UUID,
activity name, product information) per row.

\paragraph{Filter cascade.}
Starting from the full \ecoinvent{}~v3.11 activity set, we apply
the following filters in order:
\begin{enumerate}[nosep,leftmargin=2em]
\item \textbf{Market activities only.}  Retain activities whose
\ecoinvent{} \texttt{Special Activity Type} is \texttt{market
activity} or \texttt{market group}.  Production-only activities are
excluded; market activities aggregate over suppliers and are the
canonical entry point for downstream mapping.
\item \textbf{Mass-based declared unit only.}  Retain only
activities with unit \texttt{kg}.  This drops volume-based
(\texttt{m\textsuperscript{3}}), distance-based (\texttt{tkm},
\texttt{vkm}, \texttt{pkm}), area-based
(\texttt{m\textsuperscript{2}}), and item-count (\texttt{unit})
activities.  Converting back to a mass basis would require density
or geometric assumptions that vary by material; restricting the
picklist to \texttt{kg} avoids conflating mapping accuracy with
unit-conversion error.
\item \textbf{One representative geography per reference product.}
For each distinct reference product, retain a single market in the
priority order \texttt{GLO} $\rightarrow$ \texttt{RoW}
$\rightarrow$ \texttt{RER}.  The picklist is therefore
region-agnostic by construction.
\end{enumerate}

\paragraph{Energy-carrier exception.}
A small set of Task~2 triage items name an energy carrier as the
market's reference product (electricity, natural gas, or industrial
heat).  These are routine inputs in process LCA but do not have a
mass-based market activity in \ecoinvent{}; their reference
units are kWh, m\textsuperscript{3}, or MJ rather than kg, and the
filter above would otherwise drop them, leaving the model with no
valid mapping target on those items.  We admit three energy
markets by exception via an explicit allowlist
(\texttt{ENERGY\_ALLOWLIST\_UUIDS} in
\texttt{build\_picklist\_json.py}), each chosen as the broadest GLO
market group at the most common end-use form for industrial PCF
modelling:
\begin{itemize}[nosep,leftmargin=2em]
\item \texttt{market group for electricity, low voltage} (GLO,
kWh): low voltage is the end-use form for factory machinery and
process equipment.
\item \texttt{market group for natural gas, high pressure} (GLO,
m\textsuperscript{3}): high pressure is the form delivered to
industrial sites before on-site regulation.
\item \texttt{market group for heat, district or industrial,
natural gas} (GLO, MJ): the canonical natural-gas-fired
industrial heat market.
\end{itemize}
The exception is enumerated rather than rule-based so the picklist
remains an additive transformation of the public xlsx;
non-energy-carrier items are unaffected and the cascade above
otherwise stands.  The three entries lift the picklist from
2{,}571 to 2{,}574 rows.

\paragraph{What the picklist contains.}
The candidate set is intentionally close to the list a working LCA
practitioner navigates when searching \ecoinvent{} for a direct
match.  Four activity classes that a curator might be tempted to
filter out are retained because they are part of that practitioner
view: \emph{waste outputs} (output-side waste-treatment processes),
\emph{zero-EF placeholders} (activities whose computed
cradle-to-gate emission factor is exactly zero), \emph{service
activities} (e.g., \texttt{market for injection moulding}, which
accounts for the moulding \emph{process} but not the plastic
feedstock; mapping a finished plastic part to this activity
under-reports emissions by an order of magnitude or more), and
\emph{``removed by'' subtractive activities} (e.g., \texttt{market
for aluminium removed by milling, small parts}, which require a
paired material input the picklist user cannot supply).
Distinguishing these from the production-side mapping intended is
part of the modelling task a practitioner faces.  As a
ground-truth-side backstop, the \texttt{banned\_substring} rule on
Task~3 (Appendix~\ref{sec:dataset_mapping}) flags 38 mappings
whose known-wrong \ecoinvent{} name the model must avoid even when
it appears semantically plausible.

\paragraph{Output and use in evaluation.}
The cascade yields 2{,}574 deduplicated \ecoinvent{} reference
products, written to \texttt{ecoinvent\_picklist.jsonl}.  This exact list is
concatenated into the user message of every mapping/triage prompt
(Appendix~\ref{sec:prompts}) so all eight baselines see an
identical candidate set.  The picklist's region-agnostic
construction means Tasks~2 and~3 do not score geographic
specificity; a model that picks the global aggregate when a
regional process exists in \ecoinvent{} is not penalised, a
limitation we flag in Section~\ref{sec:results} and propose to
address in future work.

\section{Model Details}
\label{sec:model_details}

Table~\ref{tab:model_details} summarizes the eight baseline models
evaluated in \pcfbench{}.  All models are accessed via API (OpenAI API
for GPT models, Vertex AI for Gemini, Claude, and DeepSeek models).
Every evaluation run uses structured output via tool calls (function
calling), and a single tool whose schema encodes the expected output
format for each task.  No few-shot examples are provided.

\begin{table}[h]
\centering
\caption{Baseline models evaluated in \pcfbench{}.  Costs are
approximate list prices per 1\,M tokens at time of evaluation
(April 2026).}
\label{tab:model_details}
\footnotesize
\begin{tabular}{@{}lllrl@{}}
\toprule
\textbf{Model} & \textbf{Provider} & \textbf{API}
  & \textbf{Max Out} & \textbf{Cost (in / out)} \\
\midrule
GPT-5.5          & OpenAI    & OpenAI API & 32{,}768  & \$5.00\,/\,\$30.00 \\
GPT-5.4-mini     & OpenAI    & OpenAI API & 16{,}384  & \$0.75\,/\,\$4.50 \\
Gemini 3.1 Pro   & Google    & Vertex AI  & 64{,}000  & \$2.00\,/\,\$12.00 \\
Gemini 3 Flash   & Google    & Vertex AI  & 65{,}536  & \$0.50\,/\,\$3.00 \\
Claude Opus 4.6   & Anthropic & Vertex AI & 128{,}000 & \$5.00\,/\,\$25.00 \\
Claude Sonnet 4.6 & Anthropic & Vertex AI & 128{,}000 & \$3.00\,/\,\$15.00 \\
Claude Haiku 4.5  & Anthropic & Vertex AI & 128{,}000 & \$1.00\,/\,\$5.00 \\
DeepSeek v3.2     & DeepSeek  & Vertex AI & 8{,}192   & \$0.56\,/\,\$1.68 \\
\bottomrule
\end{tabular}
\end{table}

\section{Per-Task Evaluation Methodology}
\label{sec:task_eval_methodology}

This section describes, for each \pcfbench{} task, the input format
presented to the model, the structured output schema, and how each
scoring metric is computed.

\subsection{Decomposition (Task 1)}
\label{sec:app_decomposition}

\paragraph{Input.}
The system prompt instructs the model to act as an LCA practitioner
listing input materials for a given product.  The user message
contains the product name and its description (with composition and
product-attribute blocks removed).

\paragraph{Output schema.}
A tool call returning \texttt{\{components: list[str]\}} with up to
8 entries ordered by mass contribution.

\paragraph{Metrics.}
A Gemini~2.5~Flash judge aligns predicted and expected components
into \emph{compositional match groups}.  Four group types are
permitted:
\begin{itemize}[nosep]
  \item \textbf{1-to-1}: direct semantic match at the same fabrication
    level (e.g., \texttt{PE}~$\leftrightarrow$~\texttt{polyethylene}).
  \item \textbf{1-to-N (over-aggregation)}: one predicted aggregate
    decomposes into N expected sub-components (e.g.,
    \texttt{concrete}~$\leftrightarrow$~\{\texttt{sand},
    \texttt{water}, \texttt{cement}\}).
  \item \textbf{N-to-1 (over-decomposition)}: N predicted
    sub-components compose one expected aggregate (e.g.,
    \{\texttt{iron}, \texttt{carbon}\}~$\leftrightarrow$~\texttt{steel}).
  \item \textbf{N-to-N (parallel coverage)}: M predicted items
    collectively correspond to N expected items where neither side
    is a single aggregate of the other (e.g.,
    \{\texttt{silicon}, \texttt{copper}, \texttt{zinc},
    \texttt{magnesium}\}~$\leftrightarrow$~\{\texttt{alloying elements},
    \texttt{alloying elements (scrap)}\}).
\end{itemize}
The judge also treats the \texttt{recycled} qualifier as
non-discriminating: \texttt{plastic}~$\leftrightarrow$~%
\texttt{recycled plastic} is a valid 1-to-1 match.

Each predicted index appears in at most one group; each expected
index appears in at most one group.  This rewards models for
compositionally-consistent breakdowns at fabrication levels other
than the ground truth's, addressing a literal-match bias we observed
in v1.0 of the metric.
\begin{itemize}[nosep]
  \item \textbf{Credit policy.}
    Precision $=$ (predicted items participating in any match group)
    $/$ total predicted; Recall $=$ (expected items participating in
    any match group) $/$ total expected.  Over-aggregation and
    over-decomposition are weighted symmetrically: a single
    \texttt{concrete} prediction that subsumes
    \{\texttt{sand}, \texttt{water}, \texttt{cement}\} contributes
    $1$ to predicted-matched and $3$ to expected-matched.
  \item \textbf{$F_1$}: harmonic mean of run-level precision and
    recall.
  \item \textbf{Kendall $\tau$}: rank correlation over per-group
    mean indices, so coarser-but-correct groupings still contribute
    one ranked pair.
  \item \textbf{Exact-set match}: 1 iff every predicted and every
    expected component participates in some valid group; 0 otherwise.
\end{itemize}

\subsection{Triage (Task 2)}
\label{sec:app_triage}

\paragraph{Input.}
The system prompt instructs the model to act as an LCA expert deciding
whether a material can be mapped directly to an \ecoinvent{} activity
or needs further decomposition.  The user message contains
(i)~the candidate name and description and the root-material context
(\texttt{name}, \texttt{description}, \texttt{material\_name})
(\S\ref{sec:dataset_triage}), and (ii)~the full curated picklist of
2{,}574 \ecoinvent{} reference product names
(Appendix~\ref{sec:picklist}).

\paragraph{Output schema.}
A single tool call returning \texttt{\{should\_map: bool\}}.

\paragraph{Metrics.}
\begin{itemize}[nosep]
  \item \textbf{Accuracy}: fraction of predictions matching the
    ground-truth label.
  \item \textbf{Format compliance}: fraction of responses that are
    valid tool calls conforming to the schema.
\end{itemize}

\subsection{Mapping --- Name Only (Task 3a)}
\label{sec:app_mapping_name}

\paragraph{Input.}
The system prompt instructs the model to select the best matching
\ecoinvent{} reference product.  The user message contains
(i)~the material name and (ii)~the full curated picklist of
2{,}574 \ecoinvent{} reference product names
(Appendix~\ref{sec:picklist}).

\paragraph{Output schema.}
A single tool call returning \texttt{\{selected\_product: str\}}.

\paragraph{Metrics.}
\begin{itemize}[nosep]
  \item \textbf{Exact match (EM)}: 1 if the predicted string exactly
    matches any option in the ground-truth options list; 0
    otherwise.
  \item \textbf{Relevant substring (RS)}: 1 if the prediction contains
    the domain-critical keyword associated with the ground-truth
    mapping (e.g., ``aluminium'' for an aluminium-related material).
  \item \textbf{Banned substring absent (BA)}: 1 if the prediction
    does not contain a known-wrong mapping keyword (e.g., ``market
    for'' when the correct match is a production process).
  \item \textbf{Format compliance}.
\end{itemize}

\subsection{Mapping --- With Context (Task 3b)}
\label{sec:app_mapping_context}

\paragraph{Input.}
Identical to the name-only setting, except the user message
additionally includes the material description, supplier name, and
material group when available.

\paragraph{Output schema.}
Same as Section~\ref{sec:app_mapping_name}.

\paragraph{Metrics.}
Same as Section~\ref{sec:app_mapping_name}.

\subsection{Agentic harness for Tasks 2 and 3}
\label{sec:app_agentic_harness}

The headline numbers for Tasks~2 (triage) and~3 (mapping with
context) in Table~\ref{tab:headline_results} come from an
\emph{agentic} variant in which the model retrieves from the picklist
via two tools instead of receiving it in-prompt.  The
side-by-side non-agentic vs agentic comparison is reported in
\S\ref{sec:agentic_vs_singleshot}.

\paragraph{Tools.}
Two read-only tools backed by the same 2{,}574-item picklist
(Appendix~\ref{sec:picklist}):
\begin{itemize}[nosep]
  \item \texttt{search\_ecoinvent(keyword\_queries: list[str],
    vector\_queries: list[str])}: returns up to 20 top matches per
    query.  Keyword queries do case-insensitive substring matching
    on activity name; vector queries embed via
    \texttt{sentence-transformers/all-mpnet-base-v2} and rank by
    cosine similarity to a precomputed picklist embedding matrix
    shipped with the package.
  \item \texttt{inspect\_ecoinvent(material\_names: list[str])}:
    returns the full picklist record (name, description, EF,
    HS codes, declared unit) for one or more activity names.
\end{itemize}
The model uses these tools iteratively before issuing a final
\texttt{submit} tool call.  All tool invocations are logged into a
per-run \texttt{SearchMaterialTracker} so search behaviour is
auditable.

\paragraph{Iteration cap and force-submit.}
A request limit of 20 model calls per item caps unbounded loops.
On \texttt{UsageLimitExceeded} the orchestrator runs a constrained
final pass that exposes only the \texttt{submit} tool with a ``you
must submit now'' suffix, mirroring the prior PSWAgent ``last
iteration: terminating tools only'' behaviour.

\paragraph{Output schema and metrics.}
Identical to the non-agentic setting (Tasks 2 and 3a/3b above): a
single \texttt{submit} call with the same fields, scored by the
same accuracy / EM / RS / BA metrics.

\subsection{Extraction (Tasks 4--5)}
\label{sec:app_extraction}

\paragraph{Input.}
Two settings are evaluated to separate \emph{prior knowledge} from
\emph{document grounding}.  The user message in both cases is the
extraction query; what changes is the system prompt and whether a
source document is attached.
\begin{itemize}[nosep]
  \item \textbf{Query only}: the system prompt explicitly invites
    best-guess values from training-data knowledge (``Your task is
    to provide your best numerical estimate(s) of the queried
    parameter using only your general knowledge of the domain...
    Do not refuse on the grounds that no document was provided'').
    No source document.  Tests \emph{prior knowledge}: how good
    are the model's training-data priors keyed to the query when
    explicitly invited.
  \item \textbf{Query + document} (headline): the standard
    no-hallucinate system prompt (``Only extract values that are
    directly stated in or clearly derivable from the document
    text.  Do not estimate or hallucinate values'') plus the full
    text of the source PDF.  Tests \emph{document grounding}.
\end{itemize}
Both settings use the identical claim-F1 matching protocol below;
the only things that vary are the system prompt and whether a
document is attached.

\paragraph{Output schema.}
A tool call returning a list of claims, each with fields
\texttt{\{value: float, unit: str\}}.  The model decides how many
claims to emit per query; a single (query, document) pair frequently
has multiple ground-truth claims.

\paragraph{Matching protocol.}
Predicted and ground-truth claims are aligned by exact-tuple match.
A predicted claim $(v_p, u_p)$ matches a ground-truth claim $(v_g,
u_g)$ iff
$(\,\text{round}(\text{scale}(v_p, u_p), 6),\, \text{normalize}(u_p)\,)
{=}\,
(\,\text{round}(\text{scale}(v_g, u_g), 6),\, \text{normalize}(u_g)\,)$,
where \texttt{normalize} maps unit synonyms to a canonical form
(e.g.\ \texttt{mass\%}, \texttt{wt\%} ${\to}$ \texttt{percent};
\texttt{g/kg} ${\to}$ \texttt{kg/kg}) and \texttt{scale} applies the
matching multiplicative conversion (e.g.\ $\times 0.001$ for
\texttt{g/kg} ${\to}$ \texttt{kg/kg}).  Each prediction and each
ground-truth claim can match at most once: if a model emits the
same canonical tuple $k$ times against a ground truth that has it
$j$ times, $\min(k, j)$ matches count.  Remaining unmatched
ground-truth claims are false negatives; remaining unmatched
predictions are false positives (hallucinations relative to the
expert annotation).

The \texttt{normalize} step uses a fixed unit alias table; synonyms
or dimensionally convertible units outside the table (e.g.\
\texttt{h} vs.\ \texttt{hr}, \texttt{s} vs.\ \texttt{h}) charge an
otherwise-correct claim as both a false positive and a false
negative.  Expanding the alias table is a planned v1.1 fix.

This protocol is strict: a predicted value differing from the
ground truth at the fifth decimal place after canonicalization is
not credited.  We deliberately picked the strict rule so that the
metric directly tracks ``did the model recover the same number the
annotator wrote down,'' rather than ``did the model emit a number
of roughly the right unit family.''  The earlier iteration of this
benchmark used a greedy lowest-relative-error rule that also
counted unit-only matches with arbitrarily wrong values; under that
rule the document-vs.\ no-document gap on Tasks~4--5 collapses
because models can recall typical unit families from priors even
without the source document.

\paragraph{Metrics.}
\begin{itemize}[nosep]
  \item \textbf{Claim precision}: matched / predicted, aggregated
    across all items in the run.
  \item \textbf{Claim recall}: matched / ground-truth, aggregated
    across all items.
  \item \textbf{Claim $F_1$} (\textbf{headline}): harmonic mean of
    the run-level precision and recall.
  \item \textbf{Best $|$RE$|$ across unit-matched pairs}
    (\emph{diagnostic}): for each item, the smallest relative error
    $|v_p - v_g| / |v_g|$ across all (predicted, ground-truth)
    pairs sharing a canonical unit, ignoring assignment.  Under
    exact-tuple matching the matched pairs themselves all have
    $\text{RE}=0$ by construction, so this best-unit-matched
    statistic is what carries the value-error signal the older
    P90$|$RE$|$ on matched used to express.
  \item \textbf{Hallucination rate}: $1 - $ precision; the share
    of predicted claims that have no eligible match.
  \item \textbf{Unit correctness} (\emph{diagnostic}): 1 if any
    predicted claim's unit overlaps any ground-truth claim's unit
    (after normalization); 0 otherwise.  We retain this metric to
    surface the upstream sub-failure where a model identifies a
    parameter but in the wrong unit system, but it is not the
    headline: a model can score high on unit correctness while
    finding only a small fraction of the requested claims.
\end{itemize}

\subsection{Total \kgco{} Prediction (Task 7)}
\label{sec:app_epd}

\paragraph{Input.}
Four context-ablation settings, each adding cumulative context
to the user message:
\begin{enumerate}[nosep]
  \item \textbf{Name only}: product name and declared unit.
  \item \textbf{With description}: adds a textual product description.
  \item \textbf{With composition}: adds material composition
    percentages.
  \item \textbf{With geography}: adds the manufacturing region.
\end{enumerate}
The system prompt instructs the model to estimate cradle-to-gate
greenhouse gas emissions in \kgco{} per declared unit.

\paragraph{Output schema.}
A single tool call returning \texttt{\{kgco2e: float\}}.

\paragraph{Metrics.}
Let $\hat{y}$ denote the predicted value and $y$ the ground-truth
\kgco{} value from the EPD.  The relative error is
$\text{RE} = (\hat{y} - y) / y$.
\begin{itemize}[nosep]
  \item \textbf{Median $|\text{RE}|$}: median absolute relative error
    across all items.
  \item \textbf{Within-$F\times$ (symmetric)}: fraction of items with
    $|\log_2(\hat{y}/y)| < \log_2 F$, i.e., $1/F < \hat{y}/y < F$.
    The headline metric is within-2$\times$ ($0.5 < \hat{y}/y < 2.0$);
    we also report within-5$\times$ ($0.2 < \hat{y}/y < 5.0$).  The
    log-ratio formulation is symmetric in over- and under-prediction, so
    a 10$\times$ under-estimate ($\hat{y} = 0.1\,y$) and a 10$\times$
    over-estimate ($\hat{y} = 10\,y$) both fall outside the 2$\times$
    band.  An earlier draft of the metric used $|\text{RE}| < 1.0$,
    which is asymmetric (it credits any under-prediction with $\hat{y}
    \ge 0$ as ``within 2$\times$'' regardless of magnitude); we use the
    symmetric form throughout the headline and full-results tables.
  \item \textbf{Mean RE}: arithmetic mean of signed relative error;
    positive indicates systematic overestimation.
\end{itemize}

\begin{figure}[h]
\centering
\includegraphics[width=0.7\textwidth]{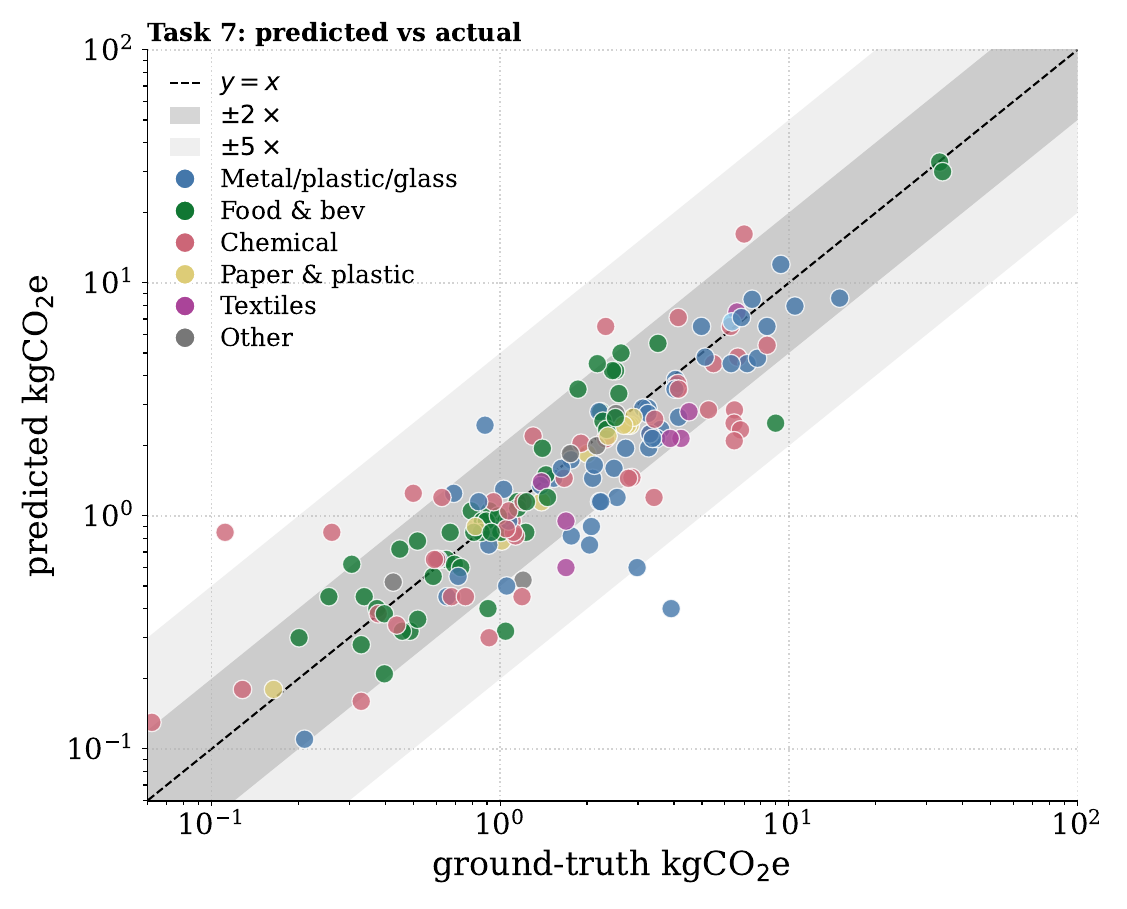}
\caption{Task~7 predicted versus actual \kgco{} on log--log scale for
    the best with-region model (lowest median $|\text{RE}|$ subject to
    coverage~$\geq$~150 of 175 items).  Diagonal is perfect prediction;
    shaded bands are $\pm$2$\times$ and $\pm$5$\times$.  Point colour
    encodes product category.}
\label{fig:task7_predicted_vs_actual}
\end{figure}

\subsection{Compositional pipeline (end-to-end Task 7)}
\label{sec:app_compositional}

\paragraph{Procedure.}
The end-to-end column of Table~\ref{tab:headline_results} is computed
by chaining the per-task agents from this appendix into a single
deterministic pipeline (one model family across all stages):
\begin{enumerate}[nosep]
  \item \textbf{Decompose} (\S\ref{sec:app_decomposition}): predict
    the bill of materials from product name and description.
  \item \textbf{Triage} (\S\ref{sec:app_triage}): for each predicted
    component, decide map-or-decompose; components labelled
    \texttt{decompose} re-enter step~1, with depth bounded
    (\texttt{max\_decomp\_depth}\,$=$\,$2$) to keep the pipeline
    finite.
  \item \textbf{Map} (\S\ref{sec:app_mapping_name}): map each leaf
    component to an \ecoinvent{} reference product.
  \item \textbf{Rate estimation} (query-only setting of
    \S\ref{sec:app_extraction}): in parallel, query the model for
    (a)~the kg-per-kg-product mass rate of each material component
    and (b)~the kWh-per-kg-product electricity rate and the
    MJ-per-kg-product natural-gas rate of the product as a whole.
  \item \textbf{Aggregation} (deterministic, Task~6): for each
    material component, look up the EF of its mapped \ecoinvent{}
    reference product in the harmonized library; for energy carriers
    use frozen IPCC~AR6 GWP100 EFs from \ecoinvent{}~v3.11
    (electricity, low voltage, GLO: $0.485$\,\kgco{}/kWh; heat from
    natural gas, RoW: $0.0671$\,\kgco{}/MJ).  Compute
    $\sum_i q_i \cdot \text{EF}_i$ to obtain one \kgco{} per declared
    unit.
\end{enumerate}
The pipeline emits a per-product trace (BOM, triage decisions,
mappings, rates, EFs, contributions) which is the data behind
Figure~\ref{fig:performance_dashboard}~(a)--(c) and
Figure~\ref{fig:ghost_components}.  Implementation:
\texttt{analysis/baselines/stepwise\_agent.py}
(\texttt{run\_compositional\_lca\_pipeline}).

\paragraph{Mass-conservation violation rate.}
Let $M$ denote the predicted material components whose rate unit
normalises to kg-per-kg-product (the BOM mass-balance subset; energy
components in kWh/kg or MJ/kg are excluded).  Define the per-product
mass-fraction sum
\[
  S \;=\; \sum_{i \in M} q_i,
\]
where $q_i$ is the kg-per-kg-product rate emitted by the model for
component $i$.  In a closed cradle-to-gate BOM, $S \geq 1$ holds by
construction: process losses, scrap, and recycled-content
double-counting can push $S$ above~$1$, but never below.
Figure~\ref{fig:performance_dashboard}~(c) reports the fraction of
products falling in each of two violation buckets:
\begin{itemize}[nosep]
  \item \texttt{mass\,$<$\,100\%}: $S < 1.0$ (incomplete BOM; the
    model decomposed less mass than the product weighs; the most
    damning case, since missing material can never be added back).
  \item \texttt{mass\,$>$\,300\%}: $S > 3.0$ (gross over-listing or
    double-counting).
\end{itemize}
The $3.0$ upper threshold gives substantial headroom for the
legitimate sources of over-listing named above (process losses,
scrap, recycled-content double-counting), which together rarely
exceed $2\times$ in real industrial inventories; $S > 3$ therefore
reflects gross over-listing rather than threshold-borderline
behaviour.
The headline mass-conservation violation rate quoted in
Section~\ref{sec:results} ($25$--$55\%$) is the union of these two
buckets.  Implementation:
\texttt{analysis/baselines/stepwise\_scoring.py::score\_stepwise\_item}.

\paragraph{Ghost components.}
A \emph{ghost} component is a BOM entry the model lists in its
breakdown but assigns $q_i = 0$, so it survives all upstream
sub-tasks yet contributes nothing to the sum.  The ghost-component
rate is the fraction of products whose breakdown contains at least
one ghost; per-model rates and the rationale for separating this
failure mode from the $S < 1$ / $S > 3$ buckets are in
Appendix~\ref{sec:ghost_components}.

\paragraph{Scoring.}
The end-to-end column uses the same symmetric within-$2\times$
\kgco{} metric as direct mode (\S\ref{sec:app_epd}), evaluated against
the same 175 EPD ground truths.  Both modes receive only the
\emph{name~+~description} input setting for parity, isolating
direct-vs-compositional capability while holding upstream context
fixed.

\subsection{Bootstrap standard errors and confidence intervals}
\label{sec:app_bootstrap}

The $\pm$ values in Table~\ref{tab:headline_results} (and the
$95\%$ percentile CIs in
\texttt{analysis/results/headline\_uncertainty.json}) come from a
non-parametric bootstrap over the per-task evaluation unit, with
$B = 1{,}000$ resamples per cell.  The standard error is the sample
SD across replicates ($\text{ddof}=1$); CIs are the $2.5$/$97.5$
percentiles of the replicate distribution.  Resampling is seeded
deterministically from \texttt{(eval\_name, model, run\_name,
metric)}, so re-running the aggregator reproduces the published
numbers.

The resampling unit varies by task to match the natural unit of
evaluation:
\begin{itemize}[nosep]
  \item \textbf{Tasks 1, 2, 3, 7}: resample over benchmark items
    with replacement; the bootstrap statistic is the mean of the
    per-item score ($F_1$ for Task~1, accuracy for Task~2, exact
    match for Task~3, within-$2\times$ symmetric for Task~7).
  \item \textbf{Tasks 4 and 5}: resample over extraction items
    (i.e., (document, query) pairs filtered by item-level
    \texttt{material} or \texttt{energy} tag), then recompute the
    aggregate run-level claim-$F_1$ from the summed matched /
    predicted / ground-truth claim counts of the resampled items.
    This preserves the run-level aggregation rule of
    \S\ref{sec:app_extraction} under resampling: items with more
    claims contribute proportionally more to the statistic, and the
    resampled $F_1$ is internally consistent with its precision and
    recall.
  \item \textbf{End-to-end (compositional) Task 7}: resample over
    the 175 EPDs; the statistic is the within-$2\times$ symmetric
    rate of the pipeline-summed \kgco{}.
\end{itemize}
Cells whose resamples reduce to a single unit (e.g.\ a metric
defined only when both predicted and ground-truth counts are
non-zero) yield $\text{SE}=\text{None}$ and are reported without a
$\pm$ value.  Implementation:
\texttt{analysis/baselines/collect\_results.py::\_bootstrap\_estimate}.

\section{Evaluation Prompts}
\label{sec:prompts}

All baselines use the same prompts across all eight baseline models.
No few-shot examples are provided.  Temperature is set to~0 (1.0 on
the Anthropic reasoning path, which mandates it); output is
structured via tool calls.

\paragraph{Task 1: Decomposition (BOM prediction).}
The system prompt instructs the model to emit a bill of materials at
practitioner granularity, ordered by mass contribution.  Output schema:
a single \texttt{submit\_decomposition} tool call returning
\texttt{\{components: list[str]\}}.

\begin{quote}\small\ttfamily
You are an LCA expert decomposing a finished product into its bill of materials (BOM). Given a product name and description, list the input materials that flow into the facility making the product -- i.e., the inputs one hop down the supply chain.

Do NOT list:
  - process or energy inputs (e.g. "electricity", "natural gas combustion")
  - generic descriptors (e.g. "raw materials", "ingredients")

Output rules:
  - List up to 8 components (fewer is fine for simple products).
  - Order from highest to lowest mass contribution to the finished product.
  - Use the most generic name a practitioner would use; only retain a brand or trade name if no generic equivalent exists.
  - Do NOT include percentages, quantities, or units.
  - Do NOT include duplicates.
\end{quote}
The user message is the three-field input
\texttt{Product: \{product\_name\}~/~Description: \{description\}~/~%
Declared unit: \{quantity\_unit\}}.

\paragraph{Task 1: Decomposition (alignment judge).}
Task~1 is the only \pcfbench{} task whose scoring loop calls an LLM:
a Gemini~2.5~Flash judge aligns predicted and expected bills of
materials into compositional match groups, and precision/recall/$F_1$
arithmetic then runs deterministically over those groups
(\S\ref{sec:app_decomposition}).  Judge model: Gemini~2.5~Flash,
temperature~0, single iteration, structured output.  Author audit on
a 50-item stratified sample agreed with the judge on 44/50 (88\%),
biasing F1 by approximately $+0.01$ absolute
(\S\ref{sec:app_judge_audit}).

\begin{quote}\small\ttfamily
You are an LCA expert grading a model's bill-of-materials decomposition.

You are given:
  - PREDICTED: a model's list of input materials, ordered from highest to lowest mass contribution.
  - EXPECTED: the ground-truth list of input materials, ordered from highest to lowest mass contribution.

Your job: align the PREDICTED list with the EXPECTED list using semantic equivalence at the level of an LCA practitioner.  Models legitimately decompose products at different fabrication levels, so both finer- and coarser-than-expected breakdowns are valid as long as they are COMPOSITIONALLY CONSISTENT.

A "match group" is one-or-more PREDICTED items that together correspond to one-or-more EXPECTED items.  Four group types are valid:

  (a) 1-to-1 -- direct match at the same fabrication level.
      "PE" <-> "polyethylene"
      "aluminium scrap (pre-consumer)" <-> "aluminium scrap"
      "PVB" <-> "polyvinyl butyral"
      "flat glass" <-> "PLANICLEAR(R) glass"

  (b) 1-to-N -- one PREDICTED aggregate that decomposes into N EXPECTED sub-components (the prediction is COARSER than the ground truth).
      "concrete" <-> \{"sand", "water", "cement"\}
      "stainless steel" <-> \{"iron", "chromium", "nickel"\}
      "dough" <-> \{"flour", "water", "yeast"\}

  (c) N-to-1 -- N PREDICTED sub-components that compose one EXPECTED aggregate (the prediction is FINER than the ground truth).
      \{"iron", "carbon"\} <-> "steel"
      \{"sand", "water", "cement"\} <-> "concrete"
      \{"cotton fibre", "weaving"\} <-> "cotton fabric"

  (d) N-to-N -- M PREDICTED items collectively correspond to N EXPECTED items where neither side is a single aggregate of the other, but the two sides cover the same set of inputs at compatible fabrication levels.
      \{"silicon", "copper", "zinc", "magnesium"\} <-> \{"alloying elements", "alloying elements (scrap)"\}
      \{"polyethylene", "ethylene vinyl acetate"\} <-> \{"thermoplastic polymer", "rubber-like polymer"\}
      \{"flour", "sugar", "yeast"\} <-> \{"baking dry mix", "leavening agents"\}

Recycled-content qualifier: the presence or absence of a "recycled" prefix on either side is NOT grounds for refusing a match.  "plastic" matches "recycled plastic"; "cotton fibre" matches "recycled cotton fibre"; "aluminium" matches "recycled aluminium".  Treat "recycled X" and "X" as compositionally equivalent for the purpose of group alignment.

For every group, the predicted side and expected side must be COMPOSITIONALLY EQUIVALENT: the aggregated items, taken together, plausibly form the other side of the group.  Match by material identity only -- ignore percentages, masses, and ordering.

Each PREDICTED index appears in AT MOST ONE group.  Each EXPECTED index appears in AT MOST ONE group.  Items that don't fit any defensible group are left unmatched.

INVALID matches:
  - "polyethylene" does NOT match "polypropylene" (chemically distinct).
  - "polyethylene" does NOT match "concrete" (no compositional relationship).
  - \{"oxygen", "hydrogen"\} does NOT match "milk" (true at the molecular level but absurd at the LCA fabrication level).
  - "aluminium" does NOT match "iron".
  - "water" does NOT match "fruit concentrates".

Pick the alignment that maximizes coverage of compositionally-defensible match groups.
\end{quote}
The user message presents the PREDICTED and EXPECTED lists indexed
from~0 and asks for every valid match group as a
(\texttt{predicted\_indices}, \texttt{expected\_indices}) pair.

\paragraph{Task 2: Mapping Triage.}
\begin{quote}\small\ttfamily
You are an LCA expert. Given a material name, decide whether it should be
mapped directly to an ecoinvent background database process, or whether it
needs to be decomposed further into sub-components first.

You have access to the full list of ecoinvent reference products. If the
material can reasonably be matched to one of these products, output
should\_map=true. If the material is too complex or composite and should
be broken down first, output should\_map=false.
\end{quote}
The user message appends the five-field market-plus-context input
described in \S\ref{sec:dataset_triage} (market \texttt{name} and
\texttt{description}; root-material \texttt{name}, \texttt{description},
\texttt{material\_name}) followed by the full list of 2{,}574 ecoinvent
reference product names.  The system prompt's reference to ``a material
name'' is a legacy artefact of the v0 flat-input shape, retained
verbatim so the prompt text is reproducible from the released code; it
predates the v1 nested input.

\paragraph{Task 3: Mapping.}
\begin{quote}\small\ttfamily
You are an LCA expert performing material-to-ecoinvent mapping. Given a
material name (and optionally additional context), select the best matching
ecoinvent reference product from the provided list.

Your output must be an exact string from the ecoinvent\_products list.
Select the product that an LCA practitioner would choose for this material.
\end{quote}
The user message includes the material name, optional context fields
(description, supplier, and purchaser context, only in the ``with
context'' setting), and the full ecoinvent product list.

\paragraph{Tasks 4--5: Extraction (query + document, headline).}
\begin{quote}\small\ttfamily
You are an LCA expert extracting physical parameters from technical
documents. Given a query about a specific parameter and the full text of a
source document, extract all relevant numerical claims.

For each claim, provide the numerical value and its unit exactly as
expressed in the document. If the document states a range, use the midpoint.
Only extract values that are directly stated in or clearly derivable from
the document text. Do not estimate or hallucinate values.
\end{quote}
The user message includes the extraction query and the full text
of the source document.

\paragraph{Tasks 4--5: Extraction (query only, ablation).}
\begin{quote}\small\ttfamily
You are an LCA expert. You will receive a query about a physical parameter
(e.g. a material input rate or an energy intensity), but no source
document. Your task is to provide your best numerical estimate(s) of the
queried parameter using only your general knowledge of the domain ---
typical values, industry conventions, textbook ranges, etc.

For each claim, provide the numerical value and its unit, matching the
unit families requested in the query. If you would naturally cite a range,
report the midpoint. It is acceptable to emit multiple claims when the
parameter has different typical values in different contexts (e.g.,
hydraulic vs. all-electric injection moulding) --- treat each as a
separate claim. Do not refuse on the grounds that no document was
provided; the goal is precisely to elicit your priors.
\end{quote}
The user message contains only the extraction query.

\paragraph{Task 7: Name only.}
\begin{quote}\small\ttfamily
You are an LCA expert estimating product carbon footprints. Given only a
product name, estimate the cradle-to-gate greenhouse gas emissions in
kg CO2 equivalent per declared unit.

Provide your best estimate as a single number. Use your knowledge of
typical emission intensities for this product category.
\end{quote}

\paragraph{Task 7: With description.}
\begin{quote}\small\ttfamily
You are an LCA expert estimating product carbon footprints. Given a product
name and description, estimate the cradle-to-gate greenhouse gas emissions
in kg CO2 equivalent per declared unit.

Provide your best estimate as a single number. Consider the product's
materials, manufacturing processes, and typical emission intensities for
this product category.
\end{quote}

\paragraph{Task 7: With composition.}
\begin{quote}\small\ttfamily
You are an LCA expert estimating product carbon footprints. Given a product
name, description, and material composition breakdown, estimate the
cradle-to-gate greenhouse gas emissions in kg CO2 equivalent per declared unit.

Use the composition percentages to weight emission factors for each
constituent material. Provide your best estimate as a single number.
\end{quote}

\paragraph{Task 7: With region.}
\begin{quote}\small\ttfamily
You are an LCA expert estimating product carbon footprints. Given a product
name, description, material composition, and manufacturing region, estimate
the cradle-to-gate greenhouse gas emissions in kg CO2 equivalent per
declared unit.

Use regional emission factors (especially grid carbon intensity) and the
composition breakdown to refine your estimate. Provide your best estimate
as a single number.
\end{quote}
Each successive setting adds fields cumulatively to the user message.

\section{Full Results Tables}
\label{sec:full_results}

This appendix reports the full per-task scoring detail underlying the
consolidated headline table in Section~\ref{sec:results}.

All tables below are regenerated from the same pinned sweep
(\texttt{HEADLINE\_RUN\_PINS}) that drives
Table~\ref{tab:headline_results}, so headline cells and per-task
breakdowns agree to the displayed precision.  Task~7 uses the
symmetric within-factor-of-$F$ metric defined in
\S\ref{sec:app_epd}.

\begin{figure}[h]
\centering
\includegraphics[width=0.78\linewidth]{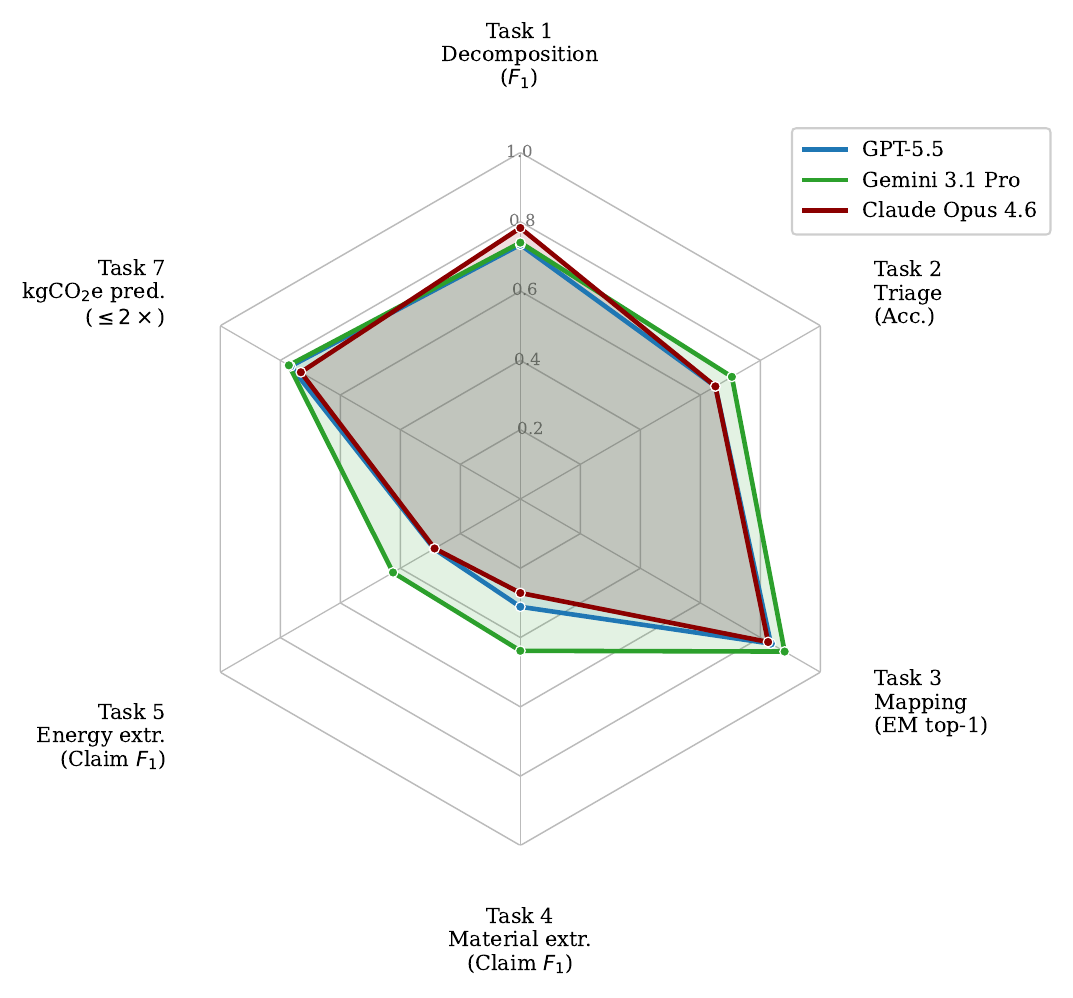}
\caption{Per-task headline metric across the three top frontier
models (GPT-5.5, Gemini~3.1~Pro, Claude~Opus~4.6).  Each vertex of
the hexagon is one \pcfbench{} task, scaled to $[0,1]$; metric per
axis is given in parentheses.  Tasks~2 and~3 are read from the
non-agentic pinned runs (\texttt{pcfbench\_triage} and
\texttt{pcfbench\_mapping\_with\_context}); the headline numbers in
Table~\ref{tab:headline_results} use the agentic harness on those
two tasks.  Other axes match Table~\ref{tab:headline_results}
exactly.}
\label{fig:per_task_radar}
\end{figure}

Figure~\ref{fig:per_task_radar} summarises the headline metric per
task across the three top frontier baselines on a single hexagonal
axis.  No model dominates on every task: Claude Opus~4.6 leads on
Task~1 (decomposition $F_1$) and Task~2 (triage), Gemini~3.1~Pro
leads on Task~3 (mapping EM), Tasks~4--5 (extraction $F_1$), and
Task~7 (\kgco{} within-2$\times$); GPT-5.5 sits between the two on
most axes.  The axes use the same metric definitions as
Table~\ref{tab:headline_results}; Tasks~2 and~3 here read the
non-agentic pinned values rather than the agentic numbers shown in
the headline table, so the small disagreement on those two axes is
expected.

\subsection{Task 1: Decomposition (full metrics)}
\label{sec:full_decomposition}

\begin{table}[h]
\centering
\caption{Task~1 Product Decomposition full results ($n=94$ products).
Precision, recall, and $F_1$ are computed over the judge-aligned
matches between predicted and expected components.  Kendall $\tau$
measures mass-ordering agreement on matched pairs ($+1$~=~perfect
agreement, $0$~=~random, $-1$~=~reversed).  Exact-set match is the
fraction of products where every predicted component matches an
expected one and vice versa.  Best result per metric in \textbf{bold}.}
\label{tab:decomposition_results_full}
\footnotesize
\setlength{\tabcolsep}{4pt}
\begin{tabular}{@{}lccccc@{}}
\toprule
\textbf{Model} & \textbf{Precision} & \textbf{Recall} & \textbf{$F_1$} & \textbf{Kendall $\tau$} & \textbf{Exact Set} \\
\midrule
Gemini 3 Flash             & 0.898          & \textbf{0.754} & \textbf{0.798} & $+$0.739          & \textbf{0.213} \\
Claude Opus 4.6$^\dagger$  & 0.885          & 0.739          & 0.782          & $+$0.652          & 0.213 \\
DeepSeek v3.2              & 0.871          & 0.702          & 0.744          & $+$0.637          & 0.170 \\
Claude Sonnet 4.6          & 0.846          & 0.718          & 0.744          & $+$0.624          & 0.149 \\
Gemini 3.1 Pro$^\dagger$   & \textbf{0.916} & 0.674          & 0.740          & $+$0.756          & 0.160 \\
GPT-5.5$^\dagger$          & 0.910          & 0.668          & 0.734          & \textbf{$+$0.779} & 0.149 \\
Claude Haiku 4.5           & 0.890          & 0.680          & 0.732          & $+$0.728          & 0.191 \\
GPT-5.4-mini               & 0.886          & 0.641          & 0.697          & $+$0.579          & 0.191 \\
\bottomrule
\end{tabular}
\end{table}

\paragraph{All models err on the conservative side.}
With the compositional-group judge in place, every evaluated model
posts $P > R$ by 14--25 points: they predict fewer components than
the EPD breakdown contains and miss the long tail of minor inputs
(additives, finishes, alloying elements).  Gemini~3.1~Pro has the
sharpest precision lead ($P=0.916$, $R=0.674$, $\Delta=0.242$);
Gemini~3~Flash has the most balanced profile ($P=0.898$, $R=0.754$,
$\Delta=0.144$) and the highest $F_1$ (0.798).  No model achieves
both high precision and high recall simultaneously.  The exact-set
match rates (0.149--0.213) indicate that, on a non-trivial fraction
of products, the model's breakdown \emph{is} compositionally
equivalent to the ground truth even though no individual component
pair matches at the same fabrication level.  The judge would have
refused this credit under the old strict-1:1 protocol.

\subsection{Task 1 judge: human-agreement audit}
\label{sec:app_judge_audit}

To check whether the Gemini~2.5~Flash compositional-match-group judge
produces alignments a human practitioner finds defensible, one author
hand-graded the judge's match groups on a 50-item stratified sample
of the Gemini~3.1~Pro run.  Items were drawn from three judge-$F_1$
strata: low ($F_1 < 0.40$,
all 5 such items), mid ($0.40 \leq F_1 < 0.70$, 29 items), and high
($F_1 \geq 0.70$, 16 items).  This surfaces both errors on
confident matches and errors on harder ones.

For each item the grader saw the predicted bill of materials, the
expected ground-truth list, the judge's match groups (rendered as
\texttt{[1:1]~PE~$\leftrightarrow$~polyethylene},
\texttt{[2:1]~\{iron, carbon\}~$\leftrightarrow$~steel}, etc.), and
the judge's resulting precision, recall, and $F_1$.  The grader
assigned one of four verdicts: \emph{agree}, \emph{the judge missed
a valid match}, \emph{the judge accepted a match that should not
have counted}, or \emph{mixed} (both happened on the same item).
The match groups shown to the grader were the original judge
outputs from the eval run, not re-issued judge calls.

\paragraph{Agreement.}  44 of 50 (88\%) verdicts were \emph{agree}
(Wilson 95\% CI $[0.76, 0.94]$).  Of the six disagreements, in five
the judge missed a valid match and in one the judge accepted a match
that should not have counted, indicating a small bias in the
conservative direction.

\paragraph{Bias-corrected $F_1$ estimate.}  Under a minimal-correction
model that adds $+1$ matched item on each side for every item where
the judge missed a match, and revokes $-1$ on each side for every
item where the judge accepted an invalid match, the per-item $F_1$
shift averaged across the 50-item sample is $+0.141$ (correction for
missed matches) and $-0.225$ (correction for invalid matches).  These
per-item magnitudes are large because the BOMs are short (median
4 components), so flipping one match moves $F_1$ substantially.
Weighted by the audit rates (5/50 missed-match items, 1/50
invalid-match items) the expected mean $F_1$ shift is approximately
$+0.010$, raising the published Gemini~3.1~Pro $F_1$ of $0.740$ to a
human-aligned estimate of $\approx 0.750$.  The shift is small
enough that the model ordering in
Table~\ref{tab:decomposition_results_full} is preserved.

\subsection{Task 2: Triage (full metrics)}
\label{sec:full_triage}

\begin{table}[h]
\centering
\caption{Task~2 Mapping Triage full results ($n=200$, balanced
100~map / 100~decompose; expert-labelled, see
\S\ref{sec:dataset_triage}).  The score implementation collapses
\texttt{accuracy = precision = recall = F1} to a single match-rate
per item; we report the single value as Accuracy.  Cost and latency
averages are per-item.}
\label{tab:triage_results_full}
\footnotesize
\begin{tabular}{@{}lcc@{}}
\toprule
\textbf{Model} & \textbf{Accuracy} & \textbf{Format \%} \\
\midrule
Claude Sonnet 4.6          & \textbf{0.650} & 1.000 \\
Claude Haiku 4.5           & 0.650          & 1.000 \\
Claude Opus 4.6$^\dagger$  & 0.645          & 1.000 \\
Gemini 3 Flash             & 0.630          & 1.000 \\
Gemini 3.1 Pro$^\dagger$   & 0.610          & 0.980 \\
DeepSeek v3.2              & 0.610          & 1.000 \\
GPT-5.5$^\dagger$          & 0.590          & 1.000 \\
GPT-5.4-mini               & 0.485          & 1.000 \\
\midrule
\textit{Chance (50/50)} & \textit{0.500} & --- \\
\bottomrule
\end{tabular}
\end{table}

\subsection{Task 3: Mapping (full metrics, both settings)}
\label{sec:full_mapping}

\begin{table}[h]
\centering
\caption{Task~3 Background database mapping with-context full results
($n=109$): non-agentic vs agentic harness, plus the
Parakeet specialized baseline.  EM = exact match; RS = relevant
substring; BA = banned substring absent.  Parakeet uses a 3-stage
paraphrase$\to$retrieve$\to$rerank pipeline
\citep{balaji2025parakeet} with Sonnet~4.6 as the LLM backbone and
GTE-Large embeddings for retrieval, re-implemented in this paper
against the 2{,}574-item picklist.  $^\dagger$ marks reasoning
configurations.}
\label{tab:full_mapping_scores}
\footnotesize
\begin{tabular}{@{}llccc@{}}
\toprule
\textbf{Model} & \textbf{Setting} & \textbf{EM} & \textbf{RS} & \textbf{BA} \\
\midrule
Gemini 3.1 Pro$^\dagger$   & Non-agentic & \textbf{0.881} & \textbf{1.000} & \textbf{1.000} \\
GPT-5.5$^\dagger$          & Non-agentic & 0.835          & 1.000          & 0.947 \\
Claude Opus 4.6$^\dagger$  & Non-agentic & 0.826          & 0.957          & 0.947 \\
Claude Sonnet 4.6          & Non-agentic & 0.807          & 0.957          & 0.947 \\
Gemini 3 Flash             & Non-agentic & 0.806          & 1.000          & 0.946 \\
Claude Haiku 4.5           & Non-agentic & 0.798          & 0.957          & 0.895 \\
DeepSeek v3.2              & Non-agentic & 0.697          & 0.913          & 0.868 \\
GPT-5.4-mini               & Non-agentic & 0.679          & 0.826          & 0.947 \\
\midrule
Gemini 3.1 Pro$^\dagger$   & Agentic & \textbf{0.927} & \textbf{0.957} & 0.974 \\
Claude Opus 4.6$^\dagger$  & Agentic & 0.853          & 0.957          & 0.974 \\
DeepSeek v3.2              & Agentic & 0.844          & 0.957          & 0.974 \\
Claude Sonnet 4.6          & Agentic & 0.817          & 0.957          & 0.947 \\
GPT-5.5$^\dagger$          & Agentic & 0.798          & 0.957          & 0.947 \\
Claude Haiku 4.5           & Agentic & 0.789          & 0.957          & 0.921 \\
GPT-5.4-mini               & Agentic & 0.752          & 0.913          & 0.947 \\
Gemini 3 Flash             & Agentic & 0.679          & 0.783          & \textbf{1.000} \\
\midrule
Parakeet (Sonnet 4.6)      & Specialized & 0.759          & ---            & --- \\
\bottomrule
\end{tabular}
\end{table}

\begin{figure}[h]
\centering
\includegraphics[width=0.7\linewidth]{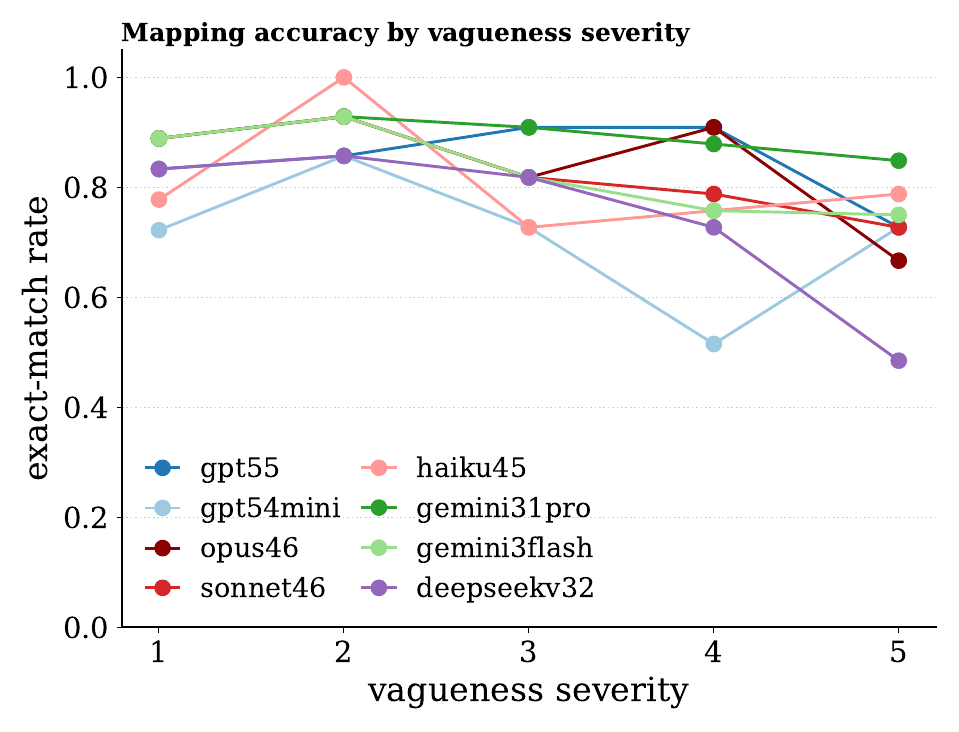}
\caption{Task~3 mapping exact-match versus expert-rated vagueness
severity (with-context setting; per-item exact-match grouped by the
1--5 vagueness label assigned at annotation time).  All 8 frontier
LLMs hold above 0.85 exact-match on the clearest items (severity~1)
and degrade roughly monotonically as ambiguity rises; the spread
between models widens at severity~4--5, where the dataset is
intentionally concentrated (Figure~\ref{fig:dataset_task3_vagueness}).
Gemini~3.1~Pro retains the lead through the high-severity tail.}
\label{fig:mapping_vagueness}
\end{figure}

\subsection{Tasks 2 and 3: non-agentic vs agentic harness}
\label{sec:agentic_vs_singleshot}

The headline Table~\ref{tab:headline_results} reports Tasks~2
(triage) and~3 (mapping with-context) under the agentic
\ecoinvent{} search/inspect harness, since tool use materially
shifts the picture for most baselines (\S\ref{sec:results}).
Tables~\ref{tab:agentic_t2_split} and~\ref{tab:agentic_t3_split}
break out the non-agentic (no tool loop) vs agentic accuracy
side-by-side per model.  The figure-level visualization is
Figure~\ref{fig:performance_dashboard}~(d).

\begin{table}[h]
\centering
\caption{Task~2 (triage) accuracy: non-agentic vs agentic
\ecoinvent{} search/inspect harness, $n=200$.  $\Delta$ is
agentic minus non-agentic.}
\label{tab:agentic_t2_split}
\footnotesize
\begin{tabular}{@{}lccc@{}}
\toprule
\textbf{Model} & \textbf{Non-agentic} & \textbf{Agentic} & $\boldsymbol{\Delta}$ \\
\midrule
GPT-5.5                    & 0.650 & 0.675          & $+0.03$ \\
GPT-5.4-mini               & 0.515 & 0.620          & $+0.10$ \\
Claude Opus 4.6            & 0.650 & 0.660          & $+0.01$ \\
Claude Sonnet 4.6          & 0.670 & \textbf{0.725} & $+0.05$ \\
Claude Haiku 4.5           & 0.685 & 0.715          & $+0.03$ \\
Gemini 3.1 Pro             & 0.705 & 0.705          & $+0.00$ \\
Gemini 3 Flash             & 0.705 & 0.680          & $-0.02$ \\
DeepSeek v3.2              & 0.675 & 0.690          & $+0.01$ \\
\bottomrule
\end{tabular}
\end{table}

\begin{table}[h]
\centering
\caption{Task~3 (mapping with context) exact-match: non-agentic
vs agentic harness, $n=109$.  $\Delta$ is agentic minus
non-agentic.}
\label{tab:agentic_t3_split}
\footnotesize
\begin{tabular}{@{}lccc@{}}
\toprule
\textbf{Model} & \textbf{Non-agentic} & \textbf{Agentic} & $\boldsymbol{\Delta}$ \\
\midrule
GPT-5.5                    & 0.835 & 0.798          & $-0.04$ \\
GPT-5.4-mini               & 0.679 & 0.752          & $+0.07$ \\
Claude Opus 4.6            & 0.826 & 0.853          & $+0.03$ \\
Claude Sonnet 4.6          & 0.807 & 0.817          & $+0.01$ \\
Claude Haiku 4.5           & 0.798 & 0.789          & $-0.01$ \\
Gemini 3.1 Pro             & 0.881 & \textbf{0.927} & $+0.05$ \\
Gemini 3 Flash             & 0.806 & 0.679          & $-0.13$ \\
DeepSeek v3.2              & 0.697 & 0.844          & $+0.15$ \\
\bottomrule
\end{tabular}
\end{table}

\subsection{Tasks 4--5: Extraction (full metrics)}
\label{sec:full_extraction}

\begin{table}[h]
\centering
\caption{Tasks 4--5 Physical parameter extraction full results.
Material (T4) split: 22 items, 55 ground-truth claims.  Energy
(T5) split: 14 items, 34 ground-truth claims (36 documents total
across the two splits).  $P$, $R$, $F_1$ aggregate matched /
predicted / ground-truth across all items per split.  Hall.\ =
hallucination rate ($1 - P$).  UC = per-item mean unit-correctness
diagnostic ($1$ if any predicted unit overlaps any GT unit, $0$
otherwise; this is not a headline metric because a model can score
high on UC while still missing most claims).  Counts are
matched~/~predicted~/~GT.
Best per column in \textbf{bold} (for Hall., lowest is best).  All
models at 100\% format compliance.  $^\dagger$ marks reasoning
configurations.}
\label{tab:extraction_results_full}
\footnotesize
\setlength{\tabcolsep}{4pt}
\begin{tabular}{@{}llccccclr@{}}
\toprule
\textbf{Model} & \textbf{Task} & \textbf{$P$} & \textbf{$R$} &
  \textbf{$F_1$} & \textbf{Hall.} & \textbf{UC} &
  \textbf{Counts (m/p/g)} \\
\midrule
DeepSeek v3.2              & Material (T4) & \textbf{0.385} & 0.764          & \textbf{0.512} & \textbf{0.615} & 0.667          & 42/109/55 \\
Gemini 3 Flash             & Material (T4) & 0.348          & 0.727          & 0.471          & 0.652          & 0.727          & 40/115/55 \\
Gemini 3.1 Pro$^\dagger$   & Material (T4) & 0.317          & 0.709          & 0.438          & 0.683          & 0.762          & 39/123/55 \\
Claude Sonnet 4.6          & Material (T4) & 0.250          & \textbf{0.891} & 0.390          & 0.750          & \textbf{0.909} & 49/196/55 \\
Claude Haiku 4.5           & Material (T4) & 0.245          & 0.691          & 0.362          & 0.755          & 0.762          & 38/155/55 \\
GPT-5.5$^\dagger$          & Material (T4) & 0.195          & 0.764          & 0.311          & 0.805          & 0.727          & 42/215/55 \\
GPT-5.4-mini               & Material (T4) & 0.184          & 0.673          & 0.289          & 0.816          & 0.727          & 37/201/55 \\
Claude Opus 4.6$^\dagger$  & Material (T4) & 0.164          & 0.782          & 0.271          & 0.836          & 0.864          & 43/262/55 \\
\midrule
Claude Haiku 4.5           & Energy (T5)   & 0.406          & 0.765          & \textbf{0.531} & 0.594          & 0.786          & 26/64/34 \\
DeepSeek v3.2              & Energy (T5)   & \textbf{0.476} & 0.588          & 0.526          & \textbf{0.524} & \textbf{0.929} & 20/42/34 \\
GPT-5.4-mini               & Energy (T5)   & 0.386          & 0.647          & 0.484          & 0.614          & 0.929          & 22/57/34 \\
Gemini 3 Flash             & Energy (T5)   & 0.324          & 0.647          & 0.431          & 0.676          & 0.923          & 22/68/34 \\
Gemini 3.1 Pro$^\dagger$   & Energy (T5)   & 0.323          & 0.618          & 0.424          & 0.677          & 0.929          & 21/65/34 \\
Claude Sonnet 4.6          & Energy (T5)   & 0.259          & \textbf{0.824} & 0.394          & 0.741          & 0.857          & 28/108/34 \\
GPT-5.5$^\dagger$          & Energy (T5)   & 0.188          & 0.618          & 0.288          & 0.812          & 0.929          & 21/112/34 \\
Claude Opus 4.6$^\dagger$  & Energy (T5)   & 0.174          & 0.794          & 0.286          & 0.826          & 0.857          & 27/155/34 \\
\bottomrule
\end{tabular}
\end{table}

\subsection{Task 7: Total \kgco{} Prediction (full metrics)}
\label{sec:full_epd}

\begin{table}[h]
\centering
\caption{Task~7 total \kgco{} prediction full results ($n=175$ EPDs).
RE $= (\hat{y} - y) / y$.  $\leq F\times$ uses the symmetric
log-ratio criterion $|\log_2(\hat{y}/y)| < \log_2 F$ (i.e.,
$1/F < \hat{y}/y < F$); see~\S\ref{sec:app_epd}.  Positive
mean~RE indicates systematic overestimation.  Best per metric per
setting in \textbf{bold} (for Med.\,$|$RE$|$ and Mean~RE, smallest
absolute value is best).  $^\dagger$ marks reasoning configurations.}
\label{tab:epd_results_full}
\footnotesize
\setlength{\tabcolsep}{4pt}
\begin{tabular}{@{}llcccc@{}}
\toprule
\textbf{Model} & \textbf{Setting} & \textbf{Med.\ $|$RE$|$} & \textbf{$\leq 2\times$} & \textbf{$\leq 5\times$} & \textbf{Mean RE} \\
\midrule
GPT-5.5$^\dagger$          & Name only   & \textbf{0.412} & \textbf{0.691} & \textbf{0.943} & $+$0.613 \\
Claude Opus 4.6$^\dagger$  & Name only   & 0.421          & 0.686          & 0.943          & $+$0.708 \\
Gemini 3.1 Pro$^\dagger$   & Name only   & 0.425          & 0.680          & 0.943          & $+$0.668 \\
Gemini 3 Flash             & Name only   & 0.433          & 0.686          & 0.931          & \textbf{$+$0.587} \\
Claude Sonnet 4.6          & Name only   & 0.490          & 0.623          & 0.897          & $+$0.972 \\
Claude Haiku 4.5           & Name only   & 0.552          & 0.571          & 0.926          & $+$1.275 \\
GPT-5.4-mini               & Name only   & 0.760          & 0.509          & 0.829          & $+$1.844 \\
DeepSeek v3.2              & Name only   & 0.773          & 0.520          & 0.846          & $+$2.391 \\
\midrule
GPT-5.5$^\dagger$          & With desc.\ & \textbf{0.336} & 0.766          & 0.966          & \textbf{$+$0.046} \\
Gemini 3.1 Pro$^\dagger$   & With desc.\ & 0.354          & \textbf{0.771} & \textbf{0.977} & $+$0.166 \\
Gemini 3 Flash             & With desc.\ & 0.364          & 0.726          & 0.977          & $+$0.078 \\
Claude Opus 4.6$^\dagger$  & With desc.\ & 0.389          & 0.731          & 0.977          & $+$0.195 \\
Claude Sonnet 4.6          & With desc.\ & 0.394          & 0.634          & 0.966          & $+$0.139 \\
Claude Haiku 4.5           & With desc.\ & 0.428          & 0.629          & 0.966          & $+$0.530 \\
DeepSeek v3.2              & With desc.\ & 0.499          & 0.632          & 0.920          & $+$0.925 \\
GPT-5.4-mini               & With desc.\ & 0.608          & 0.600          & 0.943          & $+$1.070 \\
\midrule
Claude Opus 4.6$^\dagger$  & With comp.\ & \textbf{0.281} & 0.806          & \textbf{0.989} & $+$0.069 \\
Gemini 3.1 Pro$^\dagger$   & With comp.\ & 0.292          & \textbf{0.829} & 0.989          & \textbf{$+$0.026} \\
Gemini 3 Flash             & With comp.\ & 0.318          & 0.771          & 0.989          & $-$0.050 \\
Claude Sonnet 4.6          & With comp.\ & 0.353          & 0.697          & 0.949          & $-$0.074 \\
GPT-5.5$^\dagger$          & With comp.\ & 0.353          & 0.749          & 0.971          & $-$0.031 \\
DeepSeek v3.2              & With comp.\ & 0.387          & 0.646          & 0.920          & $+$0.455 \\
GPT-5.4-mini               & With comp.\ & 0.395          & 0.640          & 0.960          & $+$0.363 \\
Claude Haiku 4.5           & With comp.\ & 0.411          & 0.646          & 0.943          & $+$0.146 \\
\midrule
Gemini 3.1 Pro$^\dagger$   & With region & \textbf{0.258} & \textbf{0.834} & 0.983          & \textbf{$+$0.015} \\
GPT-5.5$^\dagger$          & With region & 0.288          & 0.794          & 0.983          & $-$0.072 \\
Claude Opus 4.6$^\dagger$  & With region & 0.299          & 0.806          & \textbf{0.989} & $+$0.039 \\
Gemini 3 Flash             & With region & 0.324          & 0.777          & 0.983          & $-$0.079 \\
DeepSeek v3.2              & With region & 0.369          & 0.709          & 0.937          & $+$0.577 \\
Claude Sonnet 4.6          & With region & 0.370          & 0.697          & 0.983          & $-$0.055 \\
Claude Haiku 4.5           & With region & 0.433          & 0.640          & 0.931          & $+$0.356 \\
GPT-5.4-mini               & With region & 0.441          & 0.697          & 0.966          & $+$0.602 \\
\bottomrule
\end{tabular}
\end{table}

\subsection{Compositional pipeline: ghost-component rate}
\label{sec:ghost_components}

\begin{figure}[h]
\centering
\includegraphics[width=0.7\linewidth]{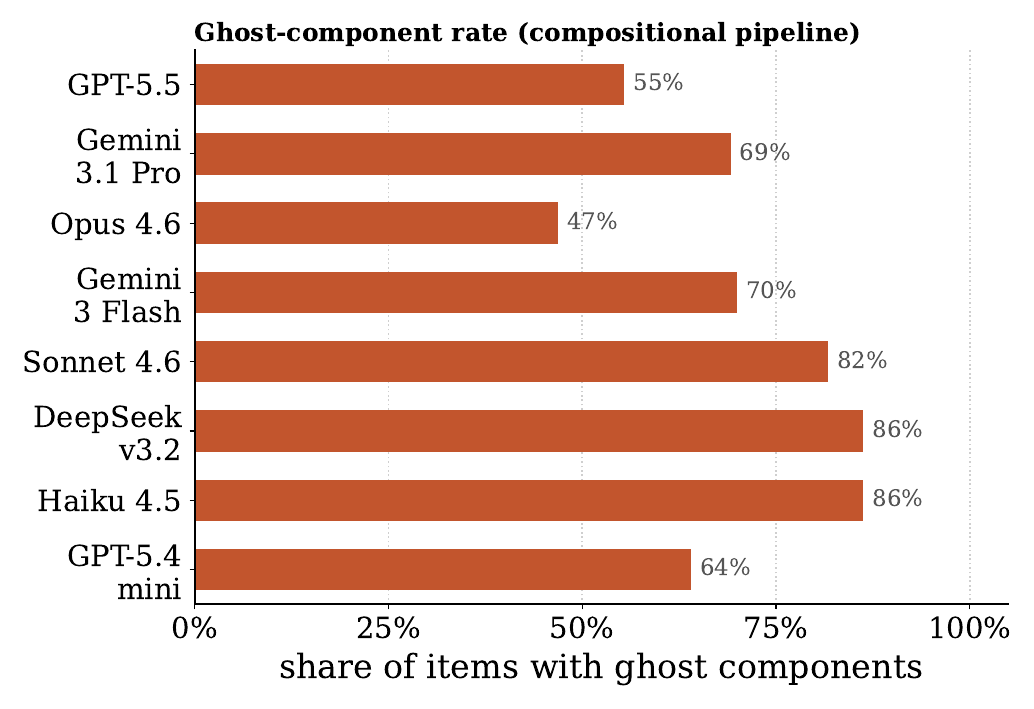}
\caption{Per-model ghost-component rate on the compositional
Task~7 pipeline (BOM entries the model lists with zero mass and
which therefore contribute nothing to the sum).  Row order matches
Figure~\ref{fig:performance_dashboard}~(a)--(c).  $47$--$86\%$ of
runs include at least one ghost across the eight models; this is
the single largest mass-balance failure mode and a likely contributor
to the under-prediction tail of compositional outputs visible in
panel~(b).}
\label{fig:ghost_components}
\end{figure}

The compositional pipeline diagnostic in
Figure~\ref{fig:performance_dashboard} (panels~a--c) tracks two
mass-conservation failure modes in the main figure (sums below
100\% and above 300\%; precise definition in
Appendix~\ref{sec:app_compositional}).  A third failure mode,
``ghost'' components, where the model lists a BOM entry but assigns
it zero mass and silently omits its contribution from the sum, is
broken out here because (a) it always pushes estimates downward,
unlike the $>$300\% bucket, and (b) its rate dwarfs the other
two modes, so stacking it would have crowded the panel.
Figure~\ref{fig:ghost_components} shows the per-model ghost rate
on the same row order as panels~a--c.

\subsection{Confidence calibration: triage and mapping}
\label{sec:confidence_calibration}

\begin{figure}[h]
\centering
\includegraphics[width=\linewidth]{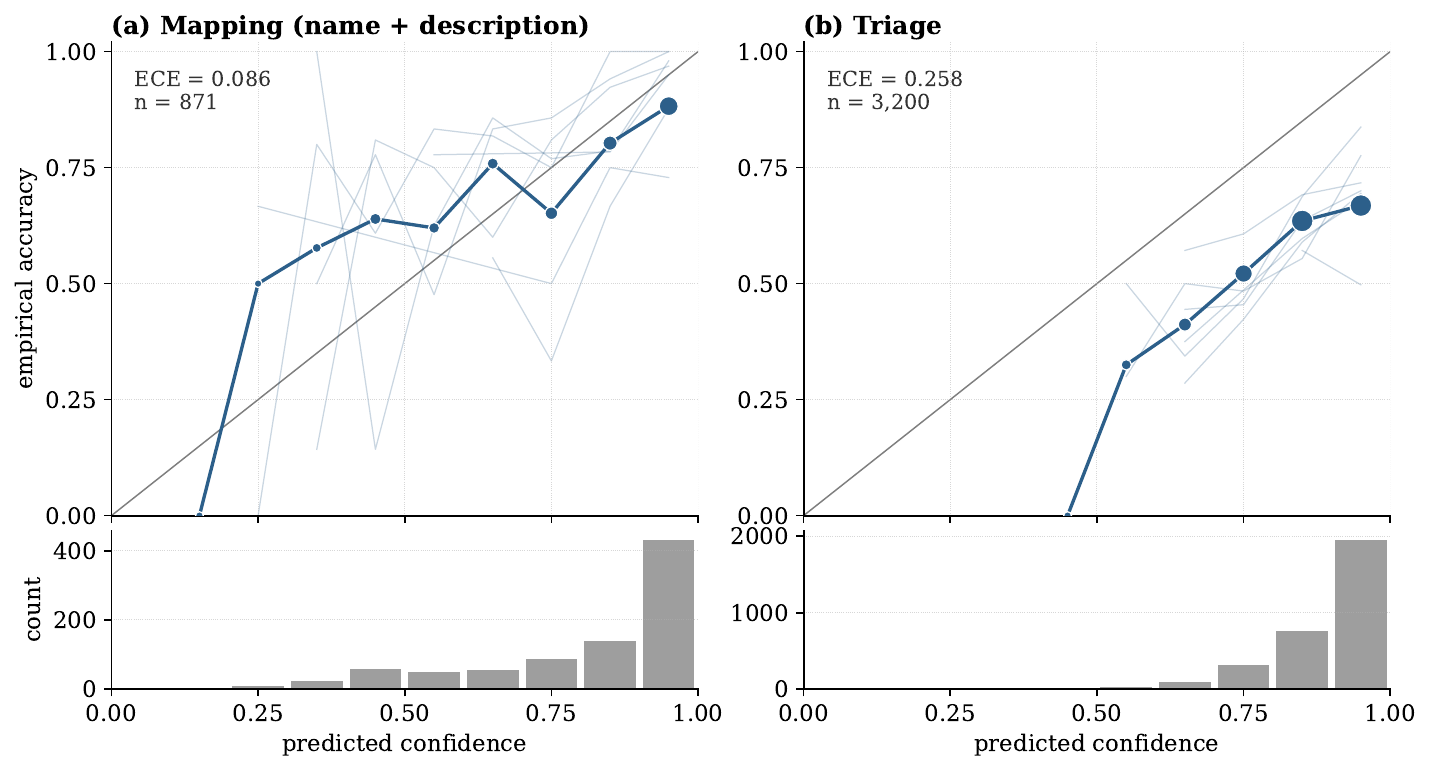}
\caption{Reliability + sharpness diagrams for mapping
(panel~a; the model sees the material name plus a free-text
description, which is the headline ``with-context'' setting from
Table~\ref{tab:headline_results}) and triage single-shot
(panel~b; the model sees the market name+description plus the
parent material context).  Top sub-panels: empirical accuracy
per confidence bin (bold blue line, markers sized by bin
population) versus the perfect-calibration diagonal.  Faint
blue curves underneath are the same diagram computed
per-model (one line per baseline; bins with $n<3$ items dropped
to suppress single-item swings).  The per-model curves cluster
around the pooled curve and share its shape relative to the
diagonal: under-confident in mapping and over-confident in
triage across all eight models and both agentic and non-agentic
ablation settings, even though the models
differ substantially in headline accuracy: the
\emph{calibration trend} is a property of the task, not of any
individual model.  Bottom sub-panels: count of
predictions per bin.  Predictions are pooled across the eight
baseline models so per-bin sample sizes support a readable curve;
ECE is computed over the same pooled set.}
\label{fig:calibration_curves}
\end{figure}

The triage and mapping submission tools both ask the model to emit a
calibrated confidence score in $[0, 1]$ alongside its answer (the
prompt explicitly invokes the standard calibration semantics: a score
of $X$ should mean the model is right $X$ fraction of the time).
Figure~\ref{fig:calibration_curves} pools predictions across all
eight baselines per task and computes empirical accuracy in
ten equal-width confidence bins, alongside the bin-population
histogram below.  Pooling across models is necessary because per-bin
sample sizes per model are too small ($\sim$1 item per bin/model on
mapping) for per-model curves to be readable; the pooled view
trades model resolution for bin density.

Two task-level patterns:

\begin{enumerate}[nosep,leftmargin=*]
    \item \textbf{Mapping is approximately calibrated} (ECE = $0.086$
    over $n = 871$ pooled predictions).  Across the populated
    confidence range ($\geq 0.45$) the empirical accuracy tracks the
    diagonal, with a slight tilt toward \emph{under}-confidence in
    the middle bins (e.g., the $0.65$-confidence bin lands at
    $\approx 0.76$ accuracy).
    \item \textbf{Triage is severely overconfident} (ECE = $0.258$
    over $n = 1{,}600$).  Every populated bin sits well below the
    diagonal: at the rightmost bin (confidence $\approx 0.95$, which
    contains $\sim$60\% of triage predictions), empirical accuracy
    is only $\approx 0.63$.
\end{enumerate}

A side note visible in both bottom histograms: every model on both
tasks compresses its confidence range to $[0.5, 1.0]$.  The lower
half of the scale is essentially unused.  An ``abstain if confidence
$< T$'' policy with $T \leq 0.5$ would therefore filter almost no
items, regardless of the threshold.

\section{Annotation Guidelines}
\label{sec:annotation}

This section documents the annotator profile, instructions, tooling,
and quality controls applied to the four expert-curated tasks
(Tasks~2, 3, 4, and~5).  Tasks~1 and~7 do not use in-house annotation:
Task~1's 94~BOMs are auto-derived from EPD composition tables and
Task~7's 175 ground-truth kgCO\textsubscript{2}e values are
third-party verified per ISO~14025~\cite{iso14025}.  Per-task
construction details (sources, harvest, schema) live in
\S\ref{sec:dataset_details}; this section is the cross-cutting
annotation reference.

\paragraph{Annotator profile.}
The annotation pool consists of six sustainability scientists who are
full-time employees of the authors' organisation, three of whom hold
PhDs in LCA or an adjacent sustainability discipline.  All annotators
encounter the underlying tasks (mapping foreground inventories to
background databases, screening physical-parameter claims from
technical documents) as part of their day-to-day client work, so the
benchmark items reflect the situations they actually adjudicate in
production rather than a synthetic exercise constructed for the paper.

\paragraph{Compensation and IRB.}
Annotators were not paid per item or per task.  Annotation occurred
during normal employment hours and was compensated as part of regular
salary.  No personal or sensitive data was collected from annotators
or from the products under analysis; the source documents are
publicly available technical PDFs and product disclosures.  Under the
relevant institutional policy this work falls outside the scope of
human-subjects review; no IRB approval was required.

\paragraph{Curation philosophy.}
For each expert-curated task, annotators were asked to author test
cases that (i)~mimic real-world LCA modelling situations they
encounter in day-to-day work, (ii)~span a difficulty range from
trivial to hard, and (iii)~cover multiple product categories
(\S\ref{sec:category_tagging}).  Where relevant, annotators conducted
internet desk research by consulting product datasheets, EPDs, and
process technical literature to surface diagnostic cases that
exercise specific failure modes.  Two illustrative Task~3 endpoints
make the difficulty range concrete:

\begin{itemize}[nosep,leftmargin=2em]
\item \textbf{Trivial.}  Input \texttt{market for steel, unalloyed}
with the exact-string match \texttt{market for steel, unalloyed}
present in the picklist: a correct mapper picks it.

\item \textbf{Hard.}  Input \texttt{Hydroformylation catalyst (Rh/Co
+ ligands)} should \emph{not} map to a rhodium activity: the
catalyst contains only trace amounts of rhodium, so mapping the
entire catalyst mass to a rhodium market over-estimates emissions
by ${\sim}10{,}000\times$.  The correct annotation marks rhodium as
a banned-substring trap rather than an accepted option, exercising
the model's ability to distinguish trace-content components from
mass-bearing ones.
\end{itemize}

The vagueness severity rating (1--5,
Figure~\ref{fig:dataset_task3_vagueness}) and the per-item
banned-substring lists (\S\ref{sec:dataset_mapping}) are direct
outputs of this curation philosophy.

\paragraph{Tooling.}
Annotators used an internal Streamlit application to author and label
items.  The application exposed (i)~a working view of the candidate
picklist or source PDF, and (ii)~a per-item form matching the released
item schema (input fields, accepted options, vagueness rating,
include/exclude verdict, evidence span).  The
application persisted every per-annotator decision separately so that
the cross-annotator reconciliation rules below could be applied
deterministically at dataset-build time rather than being collapsed
into a single ``final'' label during write time. Where multiple annotators
reviewed the same item, the application did not show the other annotators' 
verdicts on the same item.

\paragraph{Per-task instructions.}
Annotators received task-specific written instructions, summarised
here:
\begin{itemize}[nosep,leftmargin=2em]
\item \textbf{Task~2 (Triage).}  Given the five-field
candidate-plus-root-material context, label \texttt{should\_map} true
if an existing \ecoinvent{} activity well-represents the candidate,
and false otherwise.
Source items were drawn from real
observations and the released set was curated to balance the
should-map / should-decompose split 100/100
(\S\ref{sec:dataset_triage}).
\item \textbf{Task~3 (Mapping).}  Given a material name (with
optional description), select the set of defensible picklist
options, treating every listed option as equally valid at score time.
Mark known-incorrect picklist entries as banned-substrings, and rate
input vagueness on a 1--5 scale.
Multiple options are recorded when the input is genuinely ambiguous
(e.g., the \texttt{"Pca"} example in \S\ref{sec:dataset_mapping})
rather than to encode a preference ordering.  The banned-substring
mechanism captures the trace-content failure mode illustrated above
and is scored as a separate \texttt{banned\_substring\_violation}
metric.
\item \textbf{Tasks~4--5 (Extraction).}  Given a query and source PDF,
read each candidate claim and assign one of \texttt{include},
\texttt{exclude}, \texttt{duplicative}, or \texttt{irrelevant}.
Every \texttt{include} verdict requires a verbatim evidence span
copied from the PDF.  Annotators worked the same query independently;
each item is reviewed by up to three annotators
(\S\ref{sec:dataset_extraction}).  A worked example is in
\S\ref{sec:annotation_example}.
\end{itemize}

\paragraph{Reconciliation and inter-annotator agreement.}
For Tasks~4 and~5 we publish the cross-annotator
\emph{unanimous-on-include} rule rather than a single gold-annotator
label or a Cohen / Fleiss~$\kappa$ score.  Objective truth is
unavailable on the questions where the benchmark is hardest, such as
whether a catalyst should be modelled as its trace metal or whether a
plant-level input rate is the right grain to extract.  A
$\kappa$-style agreement number would either inflate apparent
reliability (if computed on the easy item-level include/exclude bit)
or under-state it (if computed on the \texttt{parameter\_name} field,
which different annotators legitimately label differently for the
same underlying claim).  The unanimous-on-include rule is conservative
by design: a claim survives only if every included-voting annotator
agreed it should be included.  Disagreement statistics for the
released set are reported at the end of
\S\ref{sec:dataset_extraction}: nine claims were dropped
because at least one annotator marked them \texttt{exclude} or
\texttt{irrelevant} while the others included, and three documents
lost all of their claims under the unanimous rule and are excluded
from the released set entirely.  Tasks~2 and~3 are single-annotator
items; their analogue of disagreement is the multi-option accepted
list (Task~3 mean 1.5~options per item) and the vagueness rating,
both of which expose annotator uncertainty as structured metadata
rather than collapsing it.

\section{Datasheet for Datasets}
\label{sec:datasheet}

This datasheet follows the Datasheets for Datasets
template~\citep{datasheets}.  Where a question already has a
complete answer in another appendix, we summarise here and cite the
relevant section rather than restating it.

\subsection{Motivation}

\paragraph{For what purpose was the dataset created?}
To evaluate whether language models and tool-using agents can
perform the operational steps of a process-based product carbon
footprint (PCF) calculation: decomposing a product into its
bill-of-materials, deciding when to map vs.\ further decompose,
mapping foreground materials onto an \ecoinvent{} activity,
extracting physical input rates from technical documentation, and
producing a single-shot total \kgco{} prediction that can be
checked against an EPD ground truth.  Existing sustainability
benchmarks are predominantly spend-based; \pcfbench{} targets the
process-based pipeline, where a wrong choice at any step silently
propagates into the final number.

\paragraph{Who created the dataset and on behalf of which entity?}
\ifanonrelease
\else
The authors, on behalf of Watershed Technology, Inc.
\fi

\paragraph{Who funded the creation of the dataset?}
\ifanonrelease
\else
Watershed Technology, Inc.
\fi

\subsection{Composition}

\paragraph{What do instances represent?}
Six per-task JSONL files plus one shared candidate-set JSONL.
Each task instance is one evaluation item with \texttt{input}
(what the model receives), \texttt{expected\_output} (the
ground-truth target), and \texttt{metadata} (stratification +
provenance fields).  Per-task schemas, sources, and harvest logic
are documented in \S\ref{sec:dataset_details}.

\paragraph{How many instances are there in total?}
614 task items across six evaluation tasks (94 decomposition + 200
triage + 109 mapping + 22 material extraction + 14 energy
extraction + 175 EPD).  Tasks~4 and~5 are scored at the claim
level: 89 ground-truth claims (55 material + 34 energy) across the
36 extraction documents.

\paragraph{Does the dataset contain all possible instances or is it a sample?}
Each task is a curated benchmark slice, not an exhaustive
enumeration.  Selection criteria, source pools, and stratification
caps per task are documented in
\S\ref{sec:dataset_decomposition}--\S\ref{sec:dataset_epd}.

\paragraph{Is each instance labelled?}
Yes.  Every instance has \texttt{expected\_output} populated.
Tasks~4--5 additionally carry verbatim evidence quotes per claim,
validated as substrings of the source \texttt{document\_text} after
whitespace normalisation (0/227 substring failures at last audit;
see \S\ref{sec:dataset_extraction}).

\paragraph{Are there recommended data splits?}
The full dataset is treated as a held-out evaluation set; there is
no designated train split.

\paragraph{Are there known errors, sources of noise, or redundancies?}
Three known sources of residual noise: (i)~Tasks~4--5 ground truth
is the unanimous-on-include consensus of up to three annotators, so
single-annotator edge cases are dropped, biasing GT toward each
document's most prominent reported claims
(\S\ref{sec:dataset_extraction}); (ii)~the Task~1 judge is itself
an LLM, validated at 88\% agreement with a human author on a
50-item stratified sample (\S\ref{sec:app_judge_audit}); (iii)~the
Tasks~2/3 candidate set is region-agnostic by construction, a
limitation flagged for future work (\S\ref{sec:picklist}).

\paragraph{Is the dataset self-contained, or does it link to external resources?}
Self-contained for scoring: each task JSONL embeds everything
needed to score predictions.  Tasks~4--5 include the OCR-extracted
\texttt{document\_text} inline; Tasks~4--5 source documents and
Task~7 EPDs additionally carry \texttt{source\_url} for
provenance.  Tasks~2 and~3 require an \ecoinvent{}~v3.11 candidate
set (2{,}574 rows, see \S\ref{sec:picklist}) at evaluation time;
that picklist ships with the companion code repository.

\paragraph{Does the dataset contain confidential, sensitive, or personally-identifying information?}
No.  All source materials are publicly disclosed Environmental
Product Declarations, peer-reviewed LCA literature, and the public
\ecoinvent{}~v3.11 \emph{Database Overview} workbook.  No
personally identifiable information is present.

\subsection{Collection Process}

\paragraph{How was the data associated with each instance acquired?}
Tasks~1 and~7 are parsed from publicly disclosed EPDs at
\texttt{environdec.com} under a written redistribution permission
(\S\ref{sec:epd_permission}).  Tasks~2, 3, 4, and~5 use in-house
expert annotation; the cross-cutting annotation procedure is
documented in \S\ref{sec:annotation}.  Per-task collection details
appear in \S\ref{sec:dataset_triage}--\S\ref{sec:dataset_mapping}
and \S\ref{sec:dataset_extraction}.

\paragraph{What mechanisms or procedures were used?}
For Tasks~1 and~7, scripted extraction from public EPD PDFs.  For
Tasks~2 and~3, structured human annotation by sustainability-domain
experts working from real material/market contexts.  For
Tasks~4--5, a Streamlit annotation interface with mandatory
verbatim-evidence selection (\S\ref{sec:annotation}).

\paragraph{Who was involved in the data collection process?}
Six sustainability practitioners employed by the authors'
organisation (three with PhDs in LCA or an adjacent discipline)
performed annotation; the authors performed dataset assembly,
judge-prompt engineering, and audit.  Annotator profile,
compensation, and ethics-review status are documented in
\S\ref{sec:annotation}.

\paragraph{Over what timeframe was the data collected?}
The benchmark slices were assembled over February~2025 to April~2026.
Underlying source materials (EPDs, literature, \ecoinvent{}~v3.11)
predate this window.

\subsection{Preprocessing, Cleaning, and Labelling}

\paragraph{Was any preprocessing/cleaning/labelling done?}
Yes, summarised here; full procedures are in \S\ref{sec:annotation}.

\begin{itemize}[nosep,leftmargin=2em]
\item \textbf{Composite-mapping filter} (Task~3): items requiring
two \ecoinvent{} activities (e.g.\ a forming process plus a raw
material) are excluded.
\item \textbf{Unanimous-on-include consensus} (Tasks~4--5): each
annotator's latest decision is taken, then claims are kept only if
every annotator with an opinion voted include (or duplicative,
which we treat as include).
\item \textbf{Within-document (value, unit) dedupe} (Tasks~4--5):
claims sharing a (rounded value, normalised unit) are collapsed and
their evidence lists unioned, so a model that produces the right
number with the right unit is correct regardless of which named
parameter the document called it.
\item \textbf{Item-level material/energy classification}
(Tasks~4--5): each item is tagged \emph{material} or \emph{energy}
based on the question text, partitioning the 36-item dataset into
the two task files (22 + 14).
\item \textbf{Product-category tagging} (all tasks): each item is
assigned one of 12 environdec product categories for stratified
analysis (\S\ref{sec:category_tagging}).
\end{itemize}

\paragraph{Was the raw data saved alongside the cleaned data?}
The published JSONL files are the post-cleaning canonical release.
Intermediate per-claim include/exclude/duplicative records for
Tasks~4--5 are retained by the authors and available on request for
replication of the consensus construction; the build scripts that
produce the JSONLs are open-sourced in the companion code
repository.

\subsection{Uses}

\paragraph{Has the dataset been used for any tasks already?}
No.  \pcfbench{} is being released for the first time alongside
this submission; the accompanying paper is its first published use.

\paragraph{What other tasks could the dataset be used for?}
Beyond LCA-specific work (sustainability-agent fine-tuning,
verifiable per-step reward shaping, compositional vs.\ monolithic
pipeline studies), the dataset supports generic LLM/agent
capability probes: top-down decomposition vs.\ single-shot
estimation on shared products (Tasks~1+7), compositional error
attribution (Tasks~1+3+4+5+7), evidence-grounded extraction with
verbatim-quote constraints, numerical reasoning over multi-unit
physical quantities, calibrated abstention vs.\ fabrication under
the Tasks~4--5 query-only ablation, semantic matching across
heterogeneous ontologies, and tool-use vs.\ in-context retrieval
ablation on the same items (Tasks~2 and~3 each have single-shot
and agentic variants).

\paragraph{Are there impacts of the composition or collection on future uses?}
The Tasks~2/3 candidate set is region-agnostic; geographic
stratification is a known future-work item.  Tasks~4--5 sources
skew toward industrial-process literature; generalisation to other
process classes is untested at this scale.  The Task~1 judge is
itself an LLM (88\% agreement with a human author on a 50-item
audit, \S\ref{sec:app_judge_audit}); treat as a strong but
imperfect signal.

\paragraph{Are there tasks for which the dataset should not be used?}
The dataset is a research benchmark, while regulatory PCF reporting
still requires an audited LCA.  Passing \pcfbench{} metrics does not
verify a system for production deployment or justify marketing it as a
``verified LCA tool''.  See
\S\ref{sec:broader_impact} for the full risk discussion.

\subsection{Distribution}

\paragraph{Will the dataset be distributed to third parties?}
Yes; full distribution, hosting, and licensing terms are in
\S\ref{sec:licensing}.  The dataset is publicly distributed via
\datahost{}; the EPD-derived fields used in Tasks~1 and~7 are
covered by a written permission grant from EPD International
(reproduced in \S\ref{sec:epd_permission}).

\paragraph{When will the dataset be distributed?}
\ifanonrelease
\else
Publicly available via the \datahost{} release (\datalink) and
companion code repository (\codelink).
\fi

\paragraph{Will the dataset be distributed under a licence?}
Yes; per-task licensing is documented in \S\ref{sec:licensing}.

\subsection{Maintenance}

\paragraph{Who will be supporting/hosting/maintaining the dataset?}
The authors, with versioned releases on \datahost{}.

\paragraph{Is there an erratum?}
Errata will be tracked in the \datahost{} dataset repository's
discussions tab and via \texttt{CHANGELOG.md} in the companion code
repository.

\paragraph{Will the dataset be updated?}
Yes.  Point releases will be issued when annotation errors are
reported and corrected, when the companion code repository's
\ecoinvent{} candidate set is regenerated for a new release, and
when additional product-category coverage (notably Infrastructure
\& buildings) is added.  When reporting numbers we recommend
pinning to a release tag.

\paragraph{Can others extend or contribute to the dataset?}
Yes; via issues or pull requests against the \datahost{} dataset
repository or the companion code repository.

\section{Broader Impact}
\label{sec:broader_impact}

\pcfbench{} focuses on \emph{compounding pipeline error in LCA-automation
agents}: a wrong call at any operational step (decomposition,
triage, mapping, parameter extraction) silently propagates into the
final \kgco{} number, and current evaluation practice for these
systems aggregates over the pipeline in a way that hides where the
error originates.  This section expands on the dual-use,
benchmark-integrity, coverage, and compute-cost considerations that
follow from releasing a process-aware diagnostic of that pipeline,
extending the summary in Section~\ref{sec:results}.

\paragraph{Intended use.}
\pcfbench{} is intended as a diagnostic measurement instrument for
research on AI-generated Product Carbon Footprint estimation.  The
appropriate uses are: (i)~identifying which sub-tasks of process-based
LCA current LLMs handle reliably and which they do not; (ii)~comparing
methodology-automation systems on a common, expert-annotated reference;
and (iii)~serving as a training and validation environment for new
extraction, mapping, and integration systems.  \pcfbench{} is
\emph{not} intended as a certification of fitness-for-deployment for
any specific system.

\paragraph{Positive societal impact.}
Models that perform well on \pcfbench{} can produce more accurate
product carbon footprints, which in turn help target emission-reduction
strategies and contribute to climate-change mitigation.

\paragraph{Risk 1: greenwashing via benchmark-validated automation.}
A persistent risk in AI-for-sustainability is that strong leaderboard
performance is repackaged as commercial validation.  A vendor could
plausibly cite \pcfbench{} scores to argue that human LCA review is
unnecessary, despite our finding that median absolute relative error
on Task~7 remains above 25\% across all models and settings, and
exceeds 40\% in the name-only setting.  We mitigate this in
three ways.  First, we report stratified per-task metrics rather than a
single headline number, making it harder to cherry-pick.  Second, we
explicitly characterize the headroom remaining for each task and the
failure modes that current models exhibit.  Third, we recommend (in
Section~\ref{sec:results}) human-in-the-loop deployment rather than
substitution.  We further encourage downstream users to cite per-task
metrics rather than aggregate ``\pcfbench{} score'' framings.

\paragraph{Risk 2: erroneous emission factor propagation.}
An incorrect emission factor has no surface anomaly: a wrong \kgco{}
value looks like a right one.  Errors
at the extraction or mapping stages therefore risk persisting through
procurement decisions, supplier scorecards, internal disclosures,
and, most consequentially, carbon-credit issuance, where the asset
being traded is the emission factor itself.  Practitioners relying on
AI-generated PCF tools should preserve provenance to source documents
and \ecoinvent{} activity IDs so that downstream auditors can re-verify
each input independently of the model.

\paragraph{Risk 3: regulatory misuse.}
Mandatory climate disclosure regimes (e.g., the EU CSRD, California
SB-253) impose specific methodological requirements that \pcfbench{}
does not encode.  Compliance with any such standard cannot be inferred
from \pcfbench{} performance.  We discourage citing \pcfbench{} scores
in regulatory filings or disclosure attestations.

\paragraph{Benchmark integrity and contamination.}
The 175~EPDs and 36~technical documents underlying our extraction
tasks are drawn from public sources and may appear in pre-training
corpora.  We cannot rule out memorization-driven gains for any model
evaluated here.  Wide adoption of \pcfbench{} could itself produce
training data as model providers optimize against the leaderboard.

\paragraph{Coverage gaps and bias of evaluation.}
\pcfbench{} scores are not equally informative across products or
regions, and downstream interpretation should reflect this.  The
mapping picklist is region-agnostic by construction
(\S\ref{sec:picklist}): the same \ecoinvent{} market is used
regardless of geography, so the benchmark does not surface failure
modes that arise when a system must select a country- or grid-
specific activity.  Product-category coverage is uneven across
tasks: Infrastructure \& buildings is empty in every task row of
the coverage heatmap (Figure~\ref{fig:coverage_heatmap}), Task~3
mapping is skewed toward chemical, paper/plastic, and
metal/mineral/plastic/glass items, and Tasks~4--5 cover only two
broad domains (industrial-process and agri/food technical
documents).  Source documents are English-language PDFs.  Together
these gaps mean a high \pcfbench{} score reflects competence in
well-represented categories and is not direct evidence of competence
on under-represented ones; users stratifying by category should
treat rows with $\leq 5$~items as illustrative rather than
statistically meaningful, and any deployment claim should be
restricted to the geographies, languages, and product categories
that actually appear in the dataset.  Closing each of these gaps is
flagged as future work.

\paragraph{Annotation provenance and consent.}
All mapping and triage annotations were produced by professional sustainability
practitioners as part of their normal work.  Source documents for
extraction (peer-reviewed publications and technical datasheets) and
EPD task items (publicly registered EPDs) are from public sources.  No
human-subjects data is included.  Annotation guidelines and
inter-annotator agreement are reported in
Appendix~\ref{sec:annotation}.

\paragraph{Environmental cost of evaluation.}
The full \pcfbench{} suite is API-only with no local GPU compute; the
inference cost is borne by the upstream provider rather than by the
benchmark user.  Per-call token counts, model identifiers, and
timestamps are preserved in the per-run JSONLs released with the
benchmark, so aggregate spend can be reconstructed by any party at
the prevailing list prices for the relevant model.  We release the
model outputs alongside the dataset so that future researchers can
analyse, re-stratify, and compare against our baselines without
re-running inference.

\section{Dataset Availability, Licensing, and Data Rights}
\label{sec:licensing}

\paragraph{Hosting and access.}
The dataset is released on \datahost{} at \datalink{} and the
evaluation harness code, per-item model outputs, and reproduction
scripts at \codelink.
A Croissant metadata
document accompanies each dataset.  Access is unrestricted for
non-commercial academic research.

\paragraph{Per-task data rights.}

\textbf{Tasks~1 (Decomposition) and~7 (End-to-end validation).}
Both tasks are derived from Environmental Product Declarations
registered with the International EPD System
(\url{https://www.environdec.com}).  The authors obtained
explicit written permission from EPD International on 14~April~2026
to redistribute, for academic research purposes, summary fields from
their EPD library: product name, product description, product
category, declared unit, declared cradle-to-gate \kgco{}, and key
material inputs / composition.  We do \emph{not} redistribute the
source EPD PDFs; for each item we include the
\texttt{environdec.com} URL so readers can access the original
document directly.  All data is attributed to the International EPD
System.  The full email thread granting permission is included
in Appendix~\ref{sec:epd_permission}.

\textbf{Tasks~4 and~5 (Material and Energy Rate Extraction).}
The 36~source documents are publicly available, open-access
technical PDFs (carbon footprint reports, EPDs, academic papers,
industry handbooks); each is reachable without a subscription or
paywall.  Each released item carries: the public source URL, the
LCA-authored extraction query, the OCR-extracted document text
inline, and the human-adjudicated ground-truth claims (value, unit,
verbatim evidence quotes from the source).  Inlining the document
text in each row keeps the benchmark runnable without re-fetching
upstream sources; we do not redistribute the original PDF binaries.
Each row also carries the public \texttt{source\_url} so any user
of the dataset can verify the source and consult the original
authors for any reuse beyond benchmark evaluation.  The author-
created annotations and the file structure are released under
CC BY-NC-SA-4.0; verbatim source-document text retains its original
copyright and is reproduced for non-commercial benchmark
evaluation.

\textbf{Task~3 (Mapping).}
The 109~material-to-\ecoinvent{} mappings, the unordered defensible-option
sets, the vagueness severity ratings, and the challenge-type tags were
created manually by the authors (practising LCA experts).  The
authors hold the rights to this dataset and
release it under CC BY-NC-SA-4.0 for academic and non-commercial use.
\ecoinvent{} reference-product names that appear as ground-truth
labels are used for identification only and remain the property of
\ecoinvent{}.

\textbf{Task~2 (Triage).}
The 200~triage items carry sustainability-expert--labelled
\texttt{should\_map} bools (map vs.\ decompose).  Inputs are
anonymised supply-chain state; no customer-identifying information
is included.  Released by the authors under CC BY-NC-SA-4.0.

\paragraph{Reproducibility.}
The evaluation harness and run scripts are released alongside the
data so any reported number outside the compositional pipeline
reproduces from a single command.  The compositional pipeline (the
End-to-end column of Table~\ref{tab:headline_results} and panels~(a)
and~(b) of Figure~\ref{fig:performance_dashboard}) sums per-task
agent outputs via deterministic \ecoinvent{}~v3.11 emission factors
and therefore requires an \ecoinvent{} licence to fully reproduce.
Without \ecoinvent{} access, users can still reproduce the per-task
numbers (Tasks~1--5 and Task~7 single-shot) and inspect the
LLM-generated decompositions, mappings, and rates that feed the
compositional pipeline.

\section{EPD International Permission Letter}
\label{sec:epd_permission}

The full email thread in which EPD International
(\url{https://www.environdec.com}) granted permission to redistribute
summary EPD fields for academic research is reproduced on the
following pages.  The grant of permission appears in the message
dated 14~April~2026.

\includepdf[pages=-,pagecommand={},fitpaper=true]{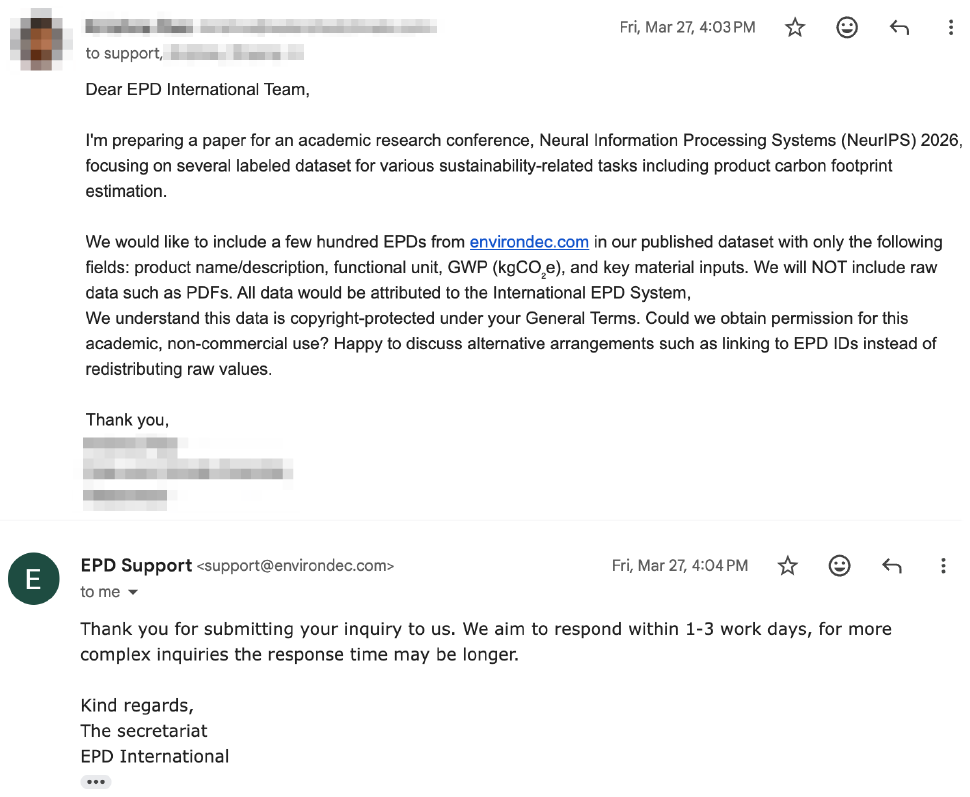}

%% file: arxiv.bbl
\begin{thebibliography}{77}
\providecommand{\natexlab}[1]{#1}
\providecommand{\url}[1]{\texttt{#1}}
\expandafter\ifx\csname urlstyle\endcsname\relax
  \providecommand{\doi}[1]{doi: #1}\else
  \providecommand{\doi}{doi: \begingroup \urlstyle{rm}\Url}\fi

\bibitem[env()]{environdec}
The international {EPD} system.
\newblock \url{https://www.environdec.com/}.
\newblock Accessed: 2026-04-20.

\bibitem[iso(2006{\natexlab{a}})]{iso14025}
{ISO} 14025:2006 --- environmental labels and declarations --- type {III}
  environmental declarations --- principles and procedures.
\newblock Standard, International Organization for Standardization,
  2006{\natexlab{a}}.

\bibitem[iso(2006{\natexlab{b}})]{iso14040}
{ISO} 14040:2006 --- environmental management --- life cycle assessment ---
  principles and framework.
\newblock Standard, International Organization for Standardization,
  2006{\natexlab{b}}.

\bibitem[iso(2006{\natexlab{c}})]{iso14044}
{ISO} 14044:2006 --- environmental management --- life cycle assessment ---
  requirements and guidelines.
\newblock Standard, International Organization for Standardization,
  2006{\natexlab{c}}.

\bibitem[iso(2018)]{iso14067}
{ISO} 14067:2018 --- greenhouse gases --- carbon footprint of products ---
  requirements and guidelines for quantification.
\newblock Standard, International Organization for Standardization, 2018.

\bibitem[An et~al.(2026)An, Kadhe, Thakur, DeLuca, and Patel]{an2026stad}
Sungeun An, Swanand~Ravindra Kadhe, Shailja Thakur, Chad DeLuca, and Hima
  Patel.
\newblock {STaD}: Scaffolded task design for identifying compositional skill
  gaps in {LLMs}.
\newblock \emph{arXiv preprint arXiv:2604.18177}, 2026.
\newblock URL \url{https://arxiv.org/abs/2604.18177}.

\bibitem[Aragón and Alberti(2024)]{aragon2024epdlimitations}
A.~Aragón and M.G. Alberti.
\newblock Limitations of machine-interpretability of digital epds used for a
  bim-based sustainability assessment of construction assets.
\newblock \emph{Journal of Building Engineering}, 96:\penalty0 110418, 2024.
\newblock ISSN 2352-7102.
\newblock \doi{https://doi.org/10.1016/j.jobe.2024.110418}.
\newblock URL
  \url{https://www.sciencedirect.com/science/article/pii/S2352710224019867}.

\bibitem[Aslan et~al.(2025)Aslan, Heijungs, and Ilievski]{aslan2025cfwizard}
Mustafa~Kaan Aslan, Reinout Heijungs, and Filip Ilievski.
\newblock The carbon footprint wizard: A knowledge-augmented {AI} interface for
  streamlining food carbon footprint analysis, 2025.
\newblock URL \url{https://arxiv.org/abs/2509.07733}.

\bibitem[Babbitt et~al.(2020)Babbitt, Madaka, Althaf, Kasulaitis, and
  Ryen]{babbitt2020disassembly}
Callie~W. Babbitt, Hema Madaka, Shahana Althaf, Barbara Kasulaitis, and
  Erinn~G. Ryen.
\newblock Disassembly-based bill of materials data for consumer electronic
  products.
\newblock \emph{Scientific Data}, 7:\penalty0 251, 2020.
\newblock \doi{10.1038/s41597-020-0573-9}.
\newblock URL \url{https://www.nature.com/articles/s41597-020-0573-9}.

\bibitem[Balaji et~al.(2025)Balaji, Ebrahimi, Domingo, Vunnava, Faridee,
  Ramalingam, Gupta, Wang, Gupta, Belcastro, Axten, Hakian, Kramer, Srinivasan,
  and Tu]{balaji2025parakeet}
Bharathan Balaji, Fahimeh Ebrahimi, Nina Gabrielle~G. Domingo, Venkata
  Sai~Gargeya Vunnava, Abu-Zaher Faridee, Soma Ramalingam, Shikha Gupta, Anran
  Wang, Harsh Gupta, Domenic Belcastro, Kellen Axten, Jeremie Hakian, Jared
  Kramer, Aravind Srinivasan, and Qingshi Tu.
\newblock Emission factor recommendation for life cycle assessments with
  generative ai.
\newblock \emph{Environmental Science \& Technology}, 59\penalty0
  (18):\penalty0 9113--9122, 05 2025.
\newblock \doi{10.1021/acs.est.4c12667}.
\newblock URL \url{https://doi.org/10.1021/acs.est.4c12667}.

\bibitem[Bang et~al.(2025)Bang, Ji, Schelten, Hartshorn, Fowler, Zhang,
  Cancedda, and Fung]{bang2025hallulens}
Yejin Bang, Ziwei Ji, Alan Schelten, Anthony Hartshorn, Tara Fowler, Cheng
  Zhang, Nicola Cancedda, and Pascale Fung.
\newblock {HalluLens}: {LLM} hallucination benchmark.
\newblock In \emph{Proceedings of the 63rd Annual Meeting of the Association
  for Computational Linguistics (ACL)}, pages 24128--24156, 2025.
\newblock \doi{10.18653/v1/2025.acl-long.1176}.
\newblock URL \url{https://aclanthology.org/2025.acl-long.1176/}.

\bibitem[Bean et~al.(2025)Bean, Kearns, Romanou, Hafner, Mayne, Batzner,
  Foroutan, Schmitz, Korgul, Batra, Deb, Beharry, Emde, Foster, Gausen,
  Grandury, Han, Hofmann, Ibrahim, Kim, Kirk, Lin, Liu, Luettgau, Magomere,
  Rystrøm, Sotnikova, Yang, Zhao, Bibi, Bosselut, Clark, Cohan, Foerster, Gal,
  Hale, Raji, Summerfield, Torr, Ududec, Rocher, and
  Mahdi]{measuringwhatmatters}
Andrew~M. Bean, Ryan~Othniel Kearns, Angelika Romanou, Franziska~Sofia Hafner,
  Harry Mayne, Jan Batzner, Negar Foroutan, Chris Schmitz, Karolina Korgul,
  Hunar Batra, Oishi Deb, Emma Beharry, Cornelius Emde, Thomas Foster, Anna
  Gausen, María Grandury, Simeng Han, Valentin Hofmann, Lujain Ibrahim, Hazel
  Kim, Hannah~Rose Kirk, Fangru Lin, Gabrielle Kaili-May Liu, Lennart Luettgau,
  Jabez Magomere, Jonathan Rystrøm, Anna Sotnikova, Yushi Yang, Yilun Zhao,
  Adel Bibi, Antoine Bosselut, Ronald Clark, Arman Cohan, Jakob Foerster, Yarin
  Gal, Scott~A. Hale, Inioluwa~Deborah Raji, Christopher Summerfield, Philip
  H.~S. Torr, Cozmin Ududec, Luc Rocher, and Adam Mahdi.
\newblock Measuring what matters: Construct validity in large language model
  benchmarks, 2025.
\newblock URL \url{https://arxiv.org/abs/2511.04703}.

\bibitem[Cardoso et~al.(2024)Cardoso, Sanhudo, Silvestre, Almeida, and
  Costa]{cardoso2024epdharmonisation}
Vitor E.~M. Cardoso, Lu{\'\i}s Sanhudo, Jos{\'e}Dinis Silvestre, Manuela
  Almeida, and Ant{\'o}nio~Aguiar Costa.
\newblock Challenges in the harmonisation and digitalisation of environmental
  product declarations for construction products in the european context.
\newblock \emph{The International Journal of Life Cycle Assessment},
  29\penalty0 (5):\penalty0 759--788, 2024.
\newblock \doi{10.1007/s11367-024-02279-w}.
\newblock URL \url{https://doi.org/10.1007/s11367-024-02279-w}.

\bibitem[Castle et~al.(2025)Castle, Schneider, Hennig, and
  Rehm]{castle2025entity}
Steffen Castle, Julian~Moreno Schneider, Leonhard Hennig, and Georg Rehm.
\newblock Entity linking using {LLMs} for automated product carbon footprint
  estimation, 2025.
\newblock URL \url{https://arxiv.org/abs/2502.07418}.

\bibitem[Chen et~al.(2025)Chen, Chen, Ning, Zhang, Wang, Yu, Li, Liao, Wei, Lu,
  Dey, Xue, Baker, Burns, Adu-Ampratwum, Huang, Ning, Gao, Su, and
  Sun]{scienceagentbench}
Ziru Chen, Shijie Chen, Yuting Ning, Qianheng Zhang, Boshi Wang, Botao Yu,
  Yifei Li, Zeyi Liao, Chen Wei, Zitong Lu, Vishal Dey, Mingyi Xue, Frazier~N.
  Baker, Benjamin Burns, Daniel Adu-Ampratwum, Xuhui Huang, Xia Ning, Song Gao,
  Yu~Su, and Huan Sun.
\newblock Scienceagentbench: Toward rigorous assessment of language agents for
  data-driven scientific discovery, 2025.
\newblock URL \url{https://arxiv.org/abs/2410.05080}.

\bibitem[Cheng et~al.(2026)Cheng, Yu, Lee, Khadpe, Ibrahim, and
  Jurafsky]{cheng2026elephant}
Myra Cheng, Sunny Yu, Cinoo Lee, Pranav Khadpe, Lujain Ibrahim, and Dan
  Jurafsky.
\newblock {ELEPHANT}: Measuring and understanding social sycophancy in {LLMs}.
\newblock In \emph{International Conference on Learning Representations
  (ICLR)}, 2026.
\newblock URL \url{https://arxiv.org/abs/2505.13995}.

\bibitem[Dagdelen et~al.(2024)Dagdelen, Dunn, Lee, Walker, Rosen, Ceder,
  Persson, and Jain]{dagdelen2024nerre}
John Dagdelen, Alexander Dunn, Sanghoon Lee, Nicholas Walker, Andrew~S. Rosen,
  Gerbrand Ceder, Kristin~A. Persson, and Anubhav Jain.
\newblock Structured information extraction from scientific text with large
  language models.
\newblock \emph{Nature Communications}, 15:\penalty0 1418, 2024.
\newblock \doi{10.1038/s41467-024-45563-x}.
\newblock URL \url{https://www.nature.com/articles/s41467-024-45563-x}.

\bibitem[Dekoninck et~al.(2025)Dekoninck, Petrov, Minchev, Balunovi{\'c},
  Vechev, Marinov, Drencheva, Konova, Shumanov, Tsvetkov, Drenchev, Todorov,
  Nikolova, Georgiev, Kalinkova, and Ismoldayev]{dekoninck2025openproof}
Jasper Dekoninck, Ivo Petrov, Kristian Minchev, Mislav Balunovi{\'c}, Martin
  Vechev, Miroslav Marinov, Maria Drencheva, Lyuba Konova, Milen Shumanov,
  Kaloyan Tsvetkov, Nikolay Drenchev, Lazar Todorov, Kalina Nikolova, Nikolay
  Georgiev, Vanesa Kalinkova, and Margulan Ismoldayev.
\newblock The open proof corpus: A large-scale study of {LLM}-generated
  mathematical proofs, 2025.
\newblock URL \url{https://arxiv.org/abs/2506.21621}.

\bibitem[Deng et~al.(2023)Deng, Liu, Luo, Yuan, Yang, Xiao, Zhou, and
  Liu]{deng2023autopcf}
Zhu Deng, Jinjie Liu, Biao Luo, Can Yuan, Qingrun Yang, Lei Xiao, Wenwen Zhou,
  and Zhu Liu.
\newblock {AutoPCF}: Efficient product carbon footprint accounting with large
  language models, 2023.
\newblock URL \url{https://arxiv.org/abs/2308.04241}.

\bibitem[Donaldson et~al.(2025)Donaldson, Balaji, Oriekezie, Kumar, and
  Patouillard]{donaldson2025lcabenchmark}
Artur Donaldson, Bharathan Balaji, Cajetan Oriekezie, Manish Kumar, and Laure
  Patouillard.
\newblock An expert-grounded benchmark of general purpose {LLMs} in {LCA},
  2025.
\newblock URL \url{https://arxiv.org/abs/2510.19886}.

\bibitem[Dumit et~al.(2026)Dumit, Rao, Ulissi, Watson, Feintzeig, Joyce, and
  Bao]{dumit2026qualityaware}
Andrew Dumit, Krishna Rao, Shaena Ulissi, Steven Watson, Jacob Feintzeig,
  P.~James Joyce, and Shuhan Bao.
\newblock Quality-aware automation for {LCI} database mapping.
\newblock Research Square preprint, 2026.
\newblock URL \url{https://www.researchsquare.com/article/rs-9285034/v1}.

\bibitem[Gachkar et~al.(2025)Gachkar, Gachkar, Ghofrani, {García Martínez},
  and {Angulo Bahón}]{gachkar2025fuzzy}
Sadaf Gachkar, Darya Gachkar, Erfan Ghofrani, Antonio {García Martínez}, and
  Cecilio {Angulo Bahón}.
\newblock Text-based algorithms for automating life cycle inventory analysis in
  building sector life cycle assessment studies.
\newblock \emph{Journal of Cleaner Production}, 486:\penalty0 144448, 2025.
\newblock ISSN 0959-6526.
\newblock \doi{https://doi.org/10.1016/j.jclepro.2024.144448}.
\newblock URL
  \url{https://www.sciencedirect.com/science/article/pii/S0959652624038976}.

\bibitem[Gebru et~al.(2021)Gebru, Morgenstern, Vecchione, Vaughan, Wallach,
  III, and Crawford]{datasheets}
Timnit Gebru, Jamie Morgenstern, Briana Vecchione, Jennifer~Wortman Vaughan,
  Hanna Wallach, Hal~Daumé III, and Kate Crawford.
\newblock Datasheets for datasets, 2021.
\newblock URL \url{https://arxiv.org/abs/1803.09010}.

\bibitem[Gelowitz and McArthur(2017)]{gelowitz2017}
M.~D.~C. Gelowitz and J.~J. McArthur.
\newblock Comparison of type {III} environmental product declarations for
  construction products: Material sourcing and harmonization evaluation.
\newblock \emph{Journal of Cleaner Production}, 157:\penalty0 125--133, 2017.
\newblock \doi{10.1016/j.jclepro.2017.04.133}.
\newblock URL \url{https://doi.org/10.1016/j.jclepro.2017.04.133}.

\bibitem[Ghosh and Tewari(2025)]{ghosh2025matprops}
Subham Ghosh and Abhishek Tewari.
\newblock Automated extraction of material properties using {LLM}-based {AI}
  agents, 2025.
\newblock URL \url{https://arxiv.org/abs/2510.01235}.

\bibitem[Gkousis et~al.(2025)Gkousis, Vasilaki, and Katsou]{gkousis2025mllci}
Spiros Gkousis, Vasileia Vasilaki, and Evina Katsou.
\newblock Machine learning and large language models for life cycle inventory
  compilation: Current situation and future developments.
\newblock \emph{Renewable and Sustainable Energy Reviews}, 228:\penalty0
  116577, 2025.
\newblock \doi{10.1016/j.rser.2025.116577}.
\newblock URL
  \url{https://www.sciencedirect.com/science/article/pii/S136403212501250X}.

\bibitem[Guan et~al.(2025)Guan, Roosta, Passban, and
  Rezagholizadeh]{guan2025ordereffect}
Bryan Guan, Tanya Roosta, Peyman Passban, and Mehdi Rezagholizadeh.
\newblock The order effect: Investigating prompt sensitivity to input order in
  {LLMs}.
\newblock \emph{arXiv preprint arXiv:2502.04134}, 2025.
\newblock URL \url{https://arxiv.org/abs/2502.04134}.

\bibitem[Guha et~al.(2023)Guha, Nyarko, Ho, Ré, Chilton, Narayana,
  Chohlas-Wood, Peters, Waldon, Rockmore, Zambrano, Talisman, Hoque, Surani,
  Fagan, Sarfaty, Dickinson, Porat, Hegland, Wu, Nudell, Niklaus, Nay, Choi,
  Tobia, Hagan, Ma, Livermore, Rasumov-Rahe, Holzenberger, Kolt, Henderson,
  Rehaag, Goel, Gao, Williams, Gandhi, Zur, Iyer, and Li]{legalbench}
Neel Guha, Julian Nyarko, Daniel~E. Ho, Christopher Ré, Adam Chilton, Aditya
  Narayana, Alex Chohlas-Wood, Austin Peters, Brandon Waldon, Daniel~N.
  Rockmore, Diego Zambrano, Dmitry Talisman, Enam Hoque, Faiz Surani, Frank
  Fagan, Galit Sarfaty, Gregory~M. Dickinson, Haggai Porat, Jason Hegland,
  Jessica Wu, Joe Nudell, Joel Niklaus, John Nay, Jonathan~H. Choi, Kevin
  Tobia, Margaret Hagan, Megan Ma, Michael Livermore, Nikon Rasumov-Rahe, Nils
  Holzenberger, Noam Kolt, Peter Henderson, Sean Rehaag, Sharad Goel, Shang
  Gao, Spencer Williams, Sunny Gandhi, Tom Zur, Varun Iyer, and Zehua Li.
\newblock Legalbench: A collaboratively built benchmark for measuring legal
  reasoning in large language models, 2023.
\newblock URL \url{https://arxiv.org/abs/2308.11462}.

\bibitem[Guo et~al.(2026)Guo, Guan, and Ma]{guo2026exioml}
Yanming Guo, Charles Guan, and Jin Ma.
\newblock Global emission factor dataset for scope 3 machine learning
  applications.
\newblock \emph{Scientific Data}, 13:\penalty0 348, 2026.
\newblock \doi{10.1038/s41597-026-06699-1}.
\newblock URL \url{https://www.nature.com/articles/s41597-026-06699-1}.

\bibitem[Hagar et~al.(2025)Hagar, Agustianto, and
  Diakopoulos]{hagar2025notwrong}
Nick Hagar, Wilma Agustianto, and Nicholas Diakopoulos.
\newblock Not wrong, but untrue: {LLM} overconfidence in document-based
  queries.
\newblock \emph{arXiv preprint arXiv:2509.25498}, 2025.
\newblock URL \url{https://arxiv.org/abs/2509.25498}.
\newblock Accepted at the Computation + Journalism Symposium 2025.

\bibitem[He et~al.(2025)He, Zhou, Wu, Yu, Zhang, Zhang, Wang, Lyu, Xu, Wang,
  Liu, and Miao]{he2025esgenius}
Chaoyue He, Xin Zhou, Yi~Wu, Xinjia Yu, Yan Zhang, Lei Zhang, Di~Wang, Shengfei
  Lyu, Hong Xu, Xiaoqiao Wang, Wei Liu, and Chunyan Miao.
\newblock {ESG}enius: Benchmarking {LLM}s on environmental, social, and
  governance ({ESG}) and sustainability knowledge.
\newblock In Christos Christodoulopoulos, Tanmoy Chakraborty, Carolyn Rose, and
  Violet Peng, editors, \emph{Proceedings of the 2025 Conference on Empirical
  Methods in Natural Language Processing}, pages 14612--14653, Suzhou, China,
  November 2025. Association for Computational Linguistics.
\newblock ISBN 979-8-89176-332-6.
\newblock \doi{10.18653/v1/2025.emnlp-main.739}.
\newblock URL \url{https://aclanthology.org/2025.emnlp-main.739/}.

\bibitem[{IPCC}(2023)]{ipcc2023synthesis}
{IPCC}.
\newblock Climate change 2023: Synthesis report. contribution of working groups
  {I}, {II} and {III} to the sixth assessment report of the intergovernmental
  panel on climate change.
\newblock Technical report, Intergovernmental Panel on Climate Change, Geneva,
  Switzerland, 2023.

\bibitem[Jimenez et~al.(2024)Jimenez, Yang, Wettig, Yao, Pei, Press, and
  Narasimhan]{jimenez2024swebench}
Carlos~E. Jimenez, John Yang, Alexander Wettig, Shunyu Yao, Kexin Pei, Ofir
  Press, and Karthik~R. Narasimhan.
\newblock {SWE-bench}: Can language models resolve real-world {GitHub} issues?
\newblock In \emph{The Twelfth International Conference on Learning
  Representations (ICLR)}, 2024.
\newblock URL \url{https://arxiv.org/abs/2310.06770}.

\bibitem[Kirichenko et~al.(2025)Kirichenko, Ibrahim, Chaudhuri, and
  Bell]{kirichenko2025abstentionbench}
Polina Kirichenko, Mark Ibrahim, Kamalika Chaudhuri, and Samuel~J. Bell.
\newblock {AbstentionBench}: Reasoning {LLMs} fail on unanswerable questions,
  2025.
\newblock URL \url{https://arxiv.org/abs/2506.09038}.

\bibitem[Konradsen et~al.(2024)Konradsen, Hansen, Ghose, and
  Pizzol]{konradsen2024}
Freja Konradsen, Kristine Sofie~Holse Hansen, Agneta Ghose, and Massimo Pizzol.
\newblock Same product, different score: how methodological differences affect
  {EPD} results.
\newblock \emph{The International Journal of Life Cycle Assessment},
  29\penalty0 (2):\penalty0 291--307, 2024.
\newblock \doi{10.1007/s11367-023-02246-x}.
\newblock URL \url{https://doi.org/10.1007/s11367-023-02246-x}.

\bibitem[Kumar et~al.(2025)Kumar, Nazemi, Kodamana, Ramteke, and
  Bakshi]{kumar2025sustainllama}
Avan Kumar, Farshid Nazemi, Hariprasad Kodamana, Manojkumar Ramteke, and
  Bhavik~R. Bakshi.
\newblock A large language model-based framework to retrieve life cycle
  inventory and environmental impact data from scientific literature.
\newblock \emph{Environmental Science \& Technology}, 59\penalty0
  (42):\penalty0 22533--22543, 2025.
\newblock \doi{10.1021/acs.est.5c05955}.
\newblock URL \url{https://pubs.acs.org/doi/10.1021/acs.est.5c05955}.

\bibitem[Kurfali et~al.(2025)Kurfali, Zahra, Nivre, and
  Messori]{kurfali2025climateeval}
Murathan Kurfali, Shorouq Zahra, Joakim Nivre, and Gabriele Messori.
\newblock {ClimateEval}: A comprehensive benchmark for {NLP} tasks related to
  climate change.
\newblock In \emph{Proceedings of the 2nd Workshop on Natural Language
  Processing Meets Climate Change (ClimateNLP 2025)}, pages 194--207, Vienna,
  Austria, 2025.
\newblock URL \url{https://aclanthology.org/2025.climatenlp-1.13/}.

\bibitem[Li et~al.(2025{\natexlab{a}})Li, Kim, and Wang]{li2025questbench}
Belinda~Z. Li, Been Kim, and Zi~Wang.
\newblock {QuestBench}: Can {LLMs} ask the right question to acquire
  information in reasoning tasks?, 2025{\natexlab{a}}.
\newblock URL \url{https://arxiv.org/abs/2503.22674}.

\bibitem[Li et~al.(2025{\natexlab{b}})Li, Chen, Xu, Li, Hu, Teng, Li, Qiu,
  Zhang, Qing, and Chen]{li2025numericbench}
Haoyang Li, Xuejia Chen, Zhanchao Xu, Darian Li, Nicole Hu, Fei Teng, Yiming
  Li, Luyu Qiu, Chen~Jason Zhang, Li~Qing, and Lei Chen.
\newblock Exposing numeracy gaps: A benchmark to evaluate fundamental numerical
  abilities in large language models.
\newblock In \emph{Findings of the Association for Computational Linguistics:
  ACL 2025}, pages 20004--20026, 2025{\natexlab{b}}.
\newblock \doi{10.18653/v1/2025.findings-acl.1026}.
\newblock URL \url{https://aclanthology.org/2025.findings-acl.1026/}.

\bibitem[Li et~al.(2025{\natexlab{c}})Li, Tang, Wang, Liu, and
  Mou]{pcfrwkv2025}
Zhen Li, Peihao Tang, Xuanlin Wang, Xueping Liu, and Peng Mou.
\newblock {PCF-RWKV}: Large language model for product carbon footprint
  estimation.
\newblock \emph{Sustainability}, 17\penalty0 (3):\penalty0 1321,
  2025{\natexlab{c}}.
\newblock \doi{10.3390/su17031321}.
\newblock URL \url{https://www.mdpi.com/2071-1050/17/3/1321}.

\bibitem[Liu et~al.(2024{\natexlab{a}})Liu, Ding, Xu, Hu, Li, Zhu, Bai, Shi,
  Wang, Song, Liu, Zhang, Wang, Li, Wang, Ruan, Huang, Sun, and
  Zhang]{liu2024medbench}
Mianxin Liu, Jinru Ding, Jie Xu, Weiguo Hu, Xiaoyang Li, Lifeng Zhu, Zhian Bai,
  Xiaoming Shi, Benyou Wang, Haitao Song, Pengfei Liu, Xiaofan Zhang, Shanshan
  Wang, Kang Li, Haofen Wang, Tong Ruan, Xuanjing Huang, Xin Sun, and Shaoting
  Zhang.
\newblock Medbench: A comprehensive, standardized, and reliable benchmarking
  system for evaluating chinese medical large language models,
  2024{\natexlab{a}}.
\newblock URL \url{https://arxiv.org/abs/2407.10990}.

\bibitem[Liu et~al.(2024{\natexlab{b}})Liu, Lin, Hewitt, Paranjape, Bevilacqua,
  Petroni, and Liang]{liu2024lost}
Nelson~F. Liu, Kevin Lin, John Hewitt, Ashwin Paranjape, Michele Bevilacqua,
  Fabio Petroni, and Percy Liang.
\newblock Lost in the middle: How language models use long contexts.
\newblock \emph{Transactions of the Association for Computational Linguistics},
  12:\penalty0 157--173, 2024{\natexlab{b}}.
\newblock \doi{10.1162/tacl_a_00638}.
\newblock URL \url{https://aclanthology.org/2024.tacl-1.9/}.

\bibitem[Ma et~al.(2024{\natexlab{a}})Ma, Zhang, Zhu, Yang, Yang, Jin, Lan,
  Kong, and He]{ma2024agentboard}
Chang Ma, Junlei Zhang, Zhihao Zhu, Cheng Yang, Yujiu Yang, Yaohui Jin,
  Zhenzhong Lan, Lingpeng Kong, and Junxian He.
\newblock {AgentBoard}: An analytical evaluation board of multi-turn {LLM}
  agents.
\newblock In \emph{Advances in Neural Information Processing Systems
  (NeurIPS)}, 2024{\natexlab{a}}.
\newblock URL \url{https://arxiv.org/abs/2401.13178}.
\newblock Oral presentation.

\bibitem[Ma et~al.(2024{\natexlab{b}})Ma, Zang, Chen, Chen, Jiao, Li, Lu, Liu,
  Ma, Dong, Zhang, Pan, Jiang, Wang, Cao, and Sun]{ma2024mmlongbenchdoc}
Yubo Ma, Yuhang Zang, Liangyu Chen, Meiqi Chen, Yizhu Jiao, Xinze Li, Xinyuan
  Lu, Ziyu Liu, Yan Ma, Xiaoyi Dong, Pan Zhang, Liangming Pan, Yu-Gang Jiang,
  Jiaqi Wang, Yixin Cao, and Aixin Sun.
\newblock {MMLONGBENCH-DOC}: Benchmarking long-context document understanding
  with visualizations.
\newblock In \emph{Advances in Neural Information Processing Systems
  (NeurIPS)}, volume~37, 2024{\natexlab{b}}.
\newblock URL \url{https://arxiv.org/abs/2407.01523}.

\bibitem[Meinrenken et~al.(2022)Meinrenken, Chen, Esparza, Iyer, Paridis,
  Prasad, and Whillas]{meinrenken2022carboncatalogue}
Christoph~J. Meinrenken, Daniel Chen, Ricardo~A. Esparza, Venkat Iyer, Sally~P.
  Paridis, Aruna Prasad, and Erika Whillas.
\newblock The carbon catalogue, carbon footprints of 866 commercial products
  from 8 industry sectors and 5 continents.
\newblock \emph{Scientific Data}, 9:\penalty0 87, 2022.
\newblock \doi{10.1038/s41597-022-01178-9}.
\newblock URL \url{https://www.nature.com/articles/s41597-022-01178-9}.

\bibitem[Mensikova et~al.(2026)Mensikova, Rizzo, and
  Hinkelman]{mensikova2026landscape}
Anastasija Mensikova, Donna~M. Rizzo, and Kathryn Hinkelman.
\newblock Mapping the landscape of artificial intelligence in life cycle
  assessment using large language models, 2026.
\newblock URL \url{https://arxiv.org/abs/2602.22500}.

\bibitem[Mirza et~al.(2025)Mirza, Alampara, Kunchapu, R{\'\i}os-Garc{\'\i}a,
  Emoekabu, Krishnan, Gupta, Schilling-Wilhelmi, Okereke, Aneesh, Asgari,
  Eberhardt, Elahi, Elbeheiry, Gil, Glaubitz, Greiner, Holick, Hoffmann,
  Ibrahim, Klepsch, K{\"o}ster, Kreth, Meyer, Miret, Peschel, Ringleb, Roesner,
  Schreiber, Schubert, Stafast, Wonanke, Pieler, Schwaller, and
  Jablonka]{mirza2025chembench}
Adrian Mirza, Nawaf Alampara, Sreekanth Kunchapu, Marti{\~n}o
  R{\'\i}os-Garc{\'\i}a, Benedict Emoekabu, Aswanth Krishnan, Tanya Gupta, Mara
  Schilling-Wilhelmi, Macjonathan Okereke, Anagha Aneesh, Mehrdad Asgari,
  Juliane Eberhardt, Amir~Mohammad Elahi, Hani~M. Elbeheiry,
  Mar{\'\i}a~Victoria Gil, Christina Glaubitz, Maximilian Greiner, Caroline~T.
  Holick, Tim Hoffmann, Abdelrahman Ibrahim, Lea~C. Klepsch, Yannik K{\"o}ster,
  Fabian~Alexander Kreth, Jakob Meyer, Santiago Miret, Jan~Matthias Peschel,
  Michael Ringleb, Nicole~C. Roesner, Johanna Schreiber, Ulrich~S. Schubert,
  Leanne~M. Stafast, A.~D.~Dinga Wonanke, Michael Pieler, Philippe Schwaller,
  and Kevin~Maik Jablonka.
\newblock A framework for evaluating the chemical knowledge and reasoning
  abilities of large language models against the expertise of chemists.
\newblock \emph{Nature Chemistry}, 17\penalty0 (7):\penalty0 1027--1034, 2025.
\newblock \doi{10.1038/s41557-025-01815-x}.
\newblock URL \url{https://doi.org/10.1038/s41557-025-01815-x}.

\bibitem[Mirzadeh et~al.(2025)Mirzadeh, Alizadeh, Shahrokhi, Tuzel, Bengio, and
  Farajtabar]{mirzadeh2025gsm}
Iman Mirzadeh, Keivan Alizadeh, Hooman Shahrokhi, Oncel Tuzel, Samy Bengio, and
  Mehrdad Farajtabar.
\newblock {GSM-Symbolic}: Understanding the limitations of mathematical
  reasoning in large language models.
\newblock In \emph{International Conference on Learning Representations
  (ICLR)}, 2025.
\newblock URL \url{https://arxiv.org/abs/2410.05229}.

\bibitem[{National Renewable Energy Laboratory}()]{uslci}
{National Renewable Energy Laboratory}.
\newblock {U.S.} life cycle inventory database ({USLCI}).
\newblock \url{https://www.nrel.gov/lci/}.
\newblock Accessed: 2026-04-29.

\bibitem[Peng et~al.(2024)Peng, Gao, Agbozo, Xu, Svynarenko, Wu, Li, and
  Tang]{peng2024kgmapping}
Tao Peng, Lu~Gao, Reuben S.~K. Agbozo, Yuming Xu, Kateryna Svynarenko, Qi~Wu,
  Changpeng Li, and Renzhong Tang.
\newblock Knowledge graph-based mapping and recommendation to automate life
  cycle assessment.
\newblock \emph{Advanced Engineering Informatics}, 62:\penalty0 102967, 2024.
\newblock \doi{10.1016/j.aei.2024.102967}.
\newblock URL
  \url{https://www.sciencedirect.com/science/article/abs/pii/S1474034624004002}.

\bibitem[Plaat et~al.(2025)Plaat, Wong, Verberne, Broekens, van Stein, and
  B{\"a}ck]{plaat2025multistep}
Aske Plaat, Annie Wong, Suzan Verberne, Joost Broekens, Niki van Stein, and
  Thomas B{\"a}ck.
\newblock Multi-step reasoning with large language models, a survey.
\newblock \emph{ACM Computing Surveys}, 58\penalty0 (6):\penalty0 160, 2025.
\newblock \doi{10.1145/3774896}.
\newblock URL \url{https://dl.acm.org/doi/10.1145/3774896}.

\bibitem[Preuss and You(2026)]{preuss2026automatinglca}
Nathan Preuss and Fengqi You.
\newblock Automating life cycle assessments through artificial intelligence
  agents and integrated assessment models.
\newblock \emph{Environmental Science \& Technology}, 60\penalty0 (1):\penalty0
  33--48, 2026.
\newblock \doi{10.1021/acs.est.5c14493}.
\newblock URL \url{https://pubs.acs.org/doi/10.1021/acs.est.5c14493}.

\bibitem[Preuss et~al.(2024)Preuss, Alshehri, and You]{preuss2024llmlca}
Nathan Preuss, Abdulelah~S. Alshehri, and Fengqi You.
\newblock Large language models for life cycle assessments: Opportunities,
  challenges, and risks.
\newblock \emph{Journal of Cleaner Production}, 466:\penalty0 142824, 2024.
\newblock \doi{10.1016/j.jclepro.2024.142824}.
\newblock URL
  \url{https://www.sciencedirect.com/science/article/abs/pii/S095965262402273X}.

\bibitem[{Primary Industries and Regions South Australia
  (PIRSA)}(2017)]{pirsa2017liming}
{Primary Industries and Regions South Australia (PIRSA)}.
\newblock The dollars and sense of liming: The stantons' story.
\newblock Technical report, Government of South Australia, Natural Resources
  Kangaroo Island, 2017.
\newblock URL
  \url{https://cdn.environment.sa.gov.au/landscape/docs/ki/liming-stanton-fact2-2017.pdf}.

\bibitem[Pu et~al.(2026)Pu, Lee, Sehwag, Lee, Zhu, Maurya, Raghavendra, Xue,
  and Denton]{pu2026lhaw}
George Pu, Michael~S. Lee, Udari~Madhushani Sehwag, David~J. Lee, Bryan Zhu,
  Yash Maurya, Mohit Raghavendra, Yuan Xue, and Samuel~Marc Denton.
\newblock {LHAW}: Controllable underspecification for long-horizon tasks, 2026.
\newblock URL \url{https://arxiv.org/abs/2602.10525}.

\bibitem[Qiang et~al.(2024)Qiang, Taylor, Wang, and Jiang]{qiang2024oaeillm}
Zhangcheng Qiang, Kerry Taylor, Weiqing Wang, and Jing Jiang.
\newblock {OAEI-LLM}: A benchmark dataset for understanding large language
  model hallucinations in ontology matching.
\newblock \emph{arXiv preprint arXiv:2409.14038}, 2024.
\newblock URL \url{https://arxiv.org/abs/2409.14038}.
\newblock Presented at ISWC 2024 Special Session; CEUR-WS Vol.~3953.

\bibitem[Qiang et~al.(2025)Qiang, Taylor, Wang, and Jiang]{qiang2025oaeillmt}
Zhangcheng Qiang, Kerry Taylor, Weiqing Wang, and Jing Jiang.
\newblock {OAEI-LLM-T}: A {TBox} benchmark dataset for understanding large
  language model hallucinations in ontology matching, 2025.
\newblock URL \url{https://arxiv.org/abs/2503.21813}.

\bibitem[Shen et~al.(2024)Shen, Song, Tan, Zhang, Ren, Yuan, Lu, Li, and
  Zhuang]{shen2024taskbench}
Yongliang Shen, Kaitao Song, Xu~Tan, Wenqi Zhang, Kan Ren, Siyu Yuan, Weiming
  Lu, Dongsheng Li, and Yueting Zhuang.
\newblock {TaskBench}: Benchmarking large language models for task automation.
\newblock In \emph{Advances in Neural Information Processing Systems
  (NeurIPS)}, volume~37, 2024.
\newblock URL \url{https://arxiv.org/abs/2311.18760}.

\bibitem[Song et~al.(2026)Song, Chen, and Schmidt]{song2026genom}
Yiping Song, Jiaoyan Chen, and Renate~A. Schmidt.
\newblock {GenOM}: Ontology matching with description generation and large
  language model.
\newblock \emph{World Wide Web}, 2026.
\newblock \doi{10.1007/s11280-026-01413-y}.
\newblock URL
  \url{https://link.springer.com/article/10.1007/s11280-026-01413-y}.

\bibitem[{Sphera Solutions}()]{gabi}
{Sphera Solutions}.
\newblock {GaBi} databases.
\newblock \url{https://sphera.com/life-cycle-assessment-lca-database/}.
\newblock Accessed: 2026-04-29.

\bibitem[Spillo et~al.(2026)Spillo, De~Filippo, Musto, Milano, and
  Semeraro]{pillo2026ecoamazon}
Giuseppe Spillo, Allegra De~Filippo, Cataldo Musto, Michela Milano, and
  Giovanni Semeraro.
\newblock {Eco-Amazon}: Enriching e-commerce datasets with product carbon
  footprint for sustainable recommendations, 2026.
\newblock URL \url{https://arxiv.org/abs/2602.15508}.

\bibitem[Sui et~al.(2024)Sui, Zhou, Zhou, Han, and Zhang]{sui2024table}
Yuan Sui, Mengyu Zhou, Mingjie Zhou, Shi Han, and Dongmei Zhang.
\newblock Table meets {LLM}: Can large language models understand structured
  table data? {A} benchmark and empirical study.
\newblock In \emph{Proceedings of the 17th ACM International Conference on Web
  Search and Data Mining (WSDM)}, 2024.
\newblock \doi{10.1145/3616855.3635752}.
\newblock URL \url{https://dl.acm.org/doi/10.1145/3616855.3635752}.

\bibitem[{TianGong Initiative, Tsinghua University}()]{tiangong}
{TianGong Initiative, Tsinghua University}.
\newblock {TianGong} {LCA} database.
\newblock \url{https://www.tiangong.earth/}.
\newblock Accessed: 2026-04-29.

\bibitem[Tu et~al.(2024)Tu, Guo, Li, Qi, and Xu]{tu2024mitigating}
Qingshi Tu, Jing Guo, Nan Li, Jianchuan Qi, and Ming Xu.
\newblock Mitigating grand challenges in life cycle inventory modeling through
  the applications of large language models.
\newblock \emph{Environmental Science \& Technology}, 58\penalty0
  (44):\penalty0 19595--19603, 2024.
\newblock \doi{10.1021/acs.est.4c07634}.
\newblock URL \url{https://pubs.acs.org/doi/10.1021/acs.est.4c07634}.

\bibitem[Ulissi et~al.(2025)Ulissi, Dumit, Joyce, Rao, Watson, and
  Suh]{ulissi2025criteria}
Shaena Ulissi, Andrew Dumit, P.~James Joyce, Krishna Rao, Steven Watson, and
  Sangwon Suh.
\newblock Criteria for credible {AI}-assisted carbon footprinting systems: The
  cases of mapping and lifecycle modeling, 2025.
\newblock URL \url{https://arxiv.org/abs/2509.00240}.

\bibitem[Vennemeyer et~al.(2025)Vennemeyer, Duong, Zhan, and
  Jiang]{vennemeyer2025sycophancy}
Daniel Vennemeyer, Phan~Anh Duong, Tiffany Zhan, and Tianyu Jiang.
\newblock Sycophancy is not one thing: Causal separation of sycophantic
  behaviors in {LLMs}.
\newblock \emph{arXiv preprint arXiv:2509.21305}, 2025.
\newblock URL \url{https://arxiv.org/abs/2509.21305}.

\bibitem[Wang et~al.(2024)Wang, Xia, He, Chen, Liu, Zhu, Liang, Wu, Liu,
  Malladi, Chevalier, Arora, and Chen]{wang2024charxiv}
Zirui Wang, Mengzhou Xia, Luxi He, Howard Chen, Yitao Liu, Richard Zhu, Kaiqu
  Liang, Xindi Wu, Haotian Liu, Sadhika Malladi, Alexis Chevalier, Sanjeev
  Arora, and Danqi Chen.
\newblock {CharXiv}: Charting gaps in realistic chart understanding in
  multimodal {LLMs}.
\newblock In \emph{Advances in Neural Information Processing Systems
  (NeurIPS)}, volume~37, 2024.
\newblock URL \url{https://arxiv.org/abs/2406.18521}.
\newblock Datasets and Benchmarks Track.

\bibitem[Wen et~al.(2025)Wen, Yao, Feng, Xu, Tsvetkov, Howe, and
  Wang]{wen2025knowyourlimits}
Bingbing Wen, Jihan Yao, Shangbin Feng, Chenjun Xu, Yulia Tsvetkov, Bill Howe,
  and Lucy~Lu Wang.
\newblock Know your limits: A survey of abstention in large language models.
\newblock \emph{Transactions of the Association for Computational Linguistics},
  13:\penalty0 529--556, 2025.
\newblock \doi{10.1162/tacl_a_00754}.
\newblock URL \url{https://aclanthology.org/2025.tacl-1.26/}.

\bibitem[Wernet et~al.(2016)Wernet, Bauer, Steubing, Reinhard, Moreno-Ruiz, and
  Weidema]{ecoinvent}
Gregor Wernet, Christian Bauer, Bernhard Steubing, J{\"u}rgen Reinhard, Emilia
  Moreno-Ruiz, and Bo~Weidema.
\newblock The ecoinvent database version 3 (part i): overview and methodology.
\newblock \emph{The International Journal of Life Cycle Assessment},
  21\penalty0 (9):\penalty0 1218--1230, 2016.
\newblock \doi{10.1007/s11367-016-1087-8}.
\newblock URL \url{https://doi.org/10.1007/s11367-016-1087-8}.

\bibitem[Wilie et~al.(2024)Wilie, Cahyawijaya, Ishii, He, and
  Fung]{wilie2024belief}
Bryan Wilie, Samuel Cahyawijaya, Etsuko Ishii, Junxian He, and Pascale Fung.
\newblock Belief revision: The adaptability of large language models reasoning.
\newblock In \emph{Proceedings of the 2024 Conference on Empirical Methods in
  Natural Language Processing (EMNLP)}, pages 10480--10496, 2024.
\newblock \doi{10.18653/v1/2024.emnlp-main.586}.
\newblock URL \url{https://aclanthology.org/2024.emnlp-main.586/}.

\bibitem[{World Resources Institute} and {World Business Council for
  Sustainable Development}(2004)]{ghgprotocol}
{World Resources Institute} and {World Business Council for Sustainable
  Development}.
\newblock A corporate accounting and reporting standard.
\newblock Technical report, Greenhouse Gas Protocol, 2004.

\bibitem[Yeh et~al.(2021)Yeh, Meng, Wang, Driscoll, Rozi, Liu, Lee, Burke,
  Lobell, and Ermon]{yeh2021sustainbench}
Christopher Yeh, Chenlin Meng, Sherrie Wang, Anne Driscoll, Erik Rozi, Patrick
  Liu, Jihyeon Lee, Marshall Burke, David Lobell, and Stefano Ermon.
\newblock {SustainBench}: Benchmarks for monitoring the sustainable development
  goals with machine learning.
\newblock In \emph{Advances in Neural Information Processing Systems (NeurIPS),
  Datasets and Benchmarks Track}, volume~35, 2021.
\newblock URL
  \url{https://datasets-benchmarks-proceedings.neurips.cc/paper/2021/hash/950a4152c2b4aa3ad78bdd6b366cc179-Abstract-round2.html}.

\bibitem[Zhang et~al.(2026)Zhang, Ye, Chen, Luo, Li, Deng, Zheng, Lin, Zheng,
  and Chen]{zhang2026dataclaw}
Qiaohong Zhang, Weihao Ye, Jialong Chen, Yi~Luo, BoYuan Li, Bowen Deng, Zibin
  Zheng, Jianhao Lin, Wei-Shi Zheng, and Chuan Chen.
\newblock {DataClawBench}: An agent benchmark for exploratory real-world
  financial data analysis, 2026.
\newblock URL \url{https://arxiv.org/abs/2605.02503}.

\bibitem[Zhang et~al.(2024)Zhang, H{\"a}hnlein, Mei, Englhardt, Patel, Schulz,
  and Iyer]{zhang2024deltalca}
Zhihan Zhang, Felix H{\"a}hnlein, Yuxuan Mei, Zachary Englhardt, Shwetak~N.
  Patel, Adriana Schulz, and Vikram Iyer.
\newblock {DeltaLCA}: Comparative life-cycle assessment for electronics design.
\newblock \emph{Proceedings of the ACM on Interactive, Mobile, Wearable and
  Ubiquitous Technologies}, 8\penalty0 (1):\penalty0 29:1--29:29, 2024.
\newblock \doi{10.1145/3643561}.
\newblock URL \url{https://dl.acm.org/doi/10.1145/3643561}.

\bibitem[Zhang et~al.(2025)Zhang, Metzger, Mei, H{\"a}hnlein, Englhardt, Cheng,
  Abowd, Patel, Schulz, and Iyer]{zhang2025autonomous}
Zhihan Zhang, Alexander Metzger, Yuxuan Mei, Felix H{\"a}hnlein, Zachary
  Englhardt, Tingyu Cheng, Gregory~D. Abowd, Shwetak Patel, Adriana Schulz, and
  Vikram Iyer.
\newblock Sustainability assessment using multimodal {AI} agents, 2025.
\newblock URL \url{https://arxiv.org/abs/2507.17012}.

\bibitem[Zhuge et~al.(2024)Zhuge, Zhao, Ashley, Wang, Khizbullin, Xiong, Liu,
  Chang, Krishnamoorthi, Tian, Shi, Chandra, and
  Schmidhuber]{zhuge2024agentasjudge}
Mingchen Zhuge, Changsheng Zhao, Dylan Ashley, Wenyi Wang, Dmitrii Khizbullin,
  Yunyang Xiong, Zechun Liu, Ernie Chang, Raghuraman Krishnamoorthi, Yuandong
  Tian, Yangyang Shi, Vikas Chandra, and J{\"u}rgen Schmidhuber.
\newblock {Agent-as-a-Judge}: Evaluate agents with agents, 2024.
\newblock URL \url{https://arxiv.org/abs/2410.10934}.

\bibitem[Zhuo et~al.(2024)Zhuo, Zhang, Fang, Duan, Lin, and
  Chen]{zhuo2024prosa}
Jingming Zhuo, Songyang Zhang, Xinyu Fang, Haodong Duan, Dahua Lin, and Kai
  Chen.
\newblock {ProSA}: Assessing and understanding the prompt sensitivity of
  {LLMs}.
\newblock In \emph{Findings of the Association for Computational Linguistics:
  EMNLP 2024}, pages 1950--1976, 2024.
\newblock \doi{10.18653/v1/2024.findings-emnlp.108}.
\newblock URL \url{https://aclanthology.org/2024.findings-emnlp.108/}.

\end{thebibliography}
